\documentclass[preprint,12pt]{elsarticle}

\usepackage{amsmath,amssymb}
\usepackage{booktabs}
\usepackage{array}
\usepackage{graphicx}
\usepackage{xcolor}
\usepackage{enumitem}
\usepackage[font=small,labelfont=bf]{caption}
\usepackage[hidelinks,breaklinks=true]{hyperref}

\newtheorem{definition}{Definition}
\newtheorem{proposition}{Proposition}

\journal{Extended Technical Report}

\begin{document}

\begin{frontmatter}

\title{Whetstones: Measuring Coevolution Between Adaptive Malware
and Behavioral Defense\\[4pt]
{\large Full Technical Report (companion to the 14-page conference version;
v1, 2026-09-11)}}

\author[cuc]{Zihan Luo}
\ead{804162281@qq.com}

\affiliation[cuc]{organization={Communication University of China},
country={China}}

\begin{abstract}
The literature on self-adapting malware is open-loop: adaptation is
evaluated against static detectors inside simulators, with fitness computed
by experimenters. This report closes the loop. We built both sides of the
adaptive-malware confrontation and, more importantly, the selection loop
between them, and measured four generations of engagement under
pre-registered protocols. On the red side, an autonomously trained adaptive
adversary completes reconnaissance--persistence--exfiltration task chains
with \emph{zero detections} under four concurrently running commercial
stacks (Microsoft Defender RTP with behavioral monitoring, Falco,
Wazuh~4.9.2, and Suricata~8.0.3 with 52{,}151 ET~Open rules). On the blue
side, a behavioral-grammar detector---a 0.88M-parameter Transformer prior
over an 8-token event grammar, fused with four further statistical
detection faces---catches that adversary at 93\% with a 3.84\% onboarding
false-positive rate. Between them sits a laboratory: a gradient range, a
hunter that kills for real, and a ledger discipline in which fitness is
read from environment outcomes alone. The measured payoff is a law chain:
blacklisting creates no selection pressure (R1: 12 cells, 0 alerts);
grammar-level hunting pushes selection onto the adversary's body (R2: 9
morphs, 0\% survival); body constants bound the rhythm gene's reachable
space (G0: the clamp backfires); and the fourth generation produced the
loop's first fit morph---\textbf{H7-full}, 100\% survival across three
runs, task chains complete, zero detections in six patrol rounds---whose
genome realizes exactly the two escape axes the defender-side theory had
predicted (temporal shape via cap cession; body configuration via a
non-hidden landing). Blind selection independently found what the analysis
said must exist. A power--latency frontier prices each escape axis
($N^{*}\approx240$ events); a network-side sibling firewall shows the
method transfers (85/85 attacks, 0.10\% FPR); and three structural
asymmetries explain why the equilibrium favors the defender who presses the
propagation plane. This extended report contains the full engineering
detail: the laboratory, both organs, the frontier, four generations of
rounds, the complete evaluation (baselines, OOV hard-subset, an honest
ADFA-LD negative result, the E1 boundary sweep, E3/E4 workload studies),
the economics of coevolution, and registered protocols E-A through E-K.
\end{abstract}

\begin{keyword}
Adversarial coevolution \sep Adaptive malware \sep Behavioral
anomaly detection \sep Host-based intrusion detection \sep eBPF telemetry
\sep Tiny language model
\end{keyword}

\end{frontmatter}

\section{Introduction}
\label{sec:intro}

Host-based intrusion detection has undergone three generations of
evolution: from system-call sequence matching~\cite{hofmeyr1998} through
audit-log n-gram analysis~\cite{lee1998} to modern eBPF-based telemetry
pipelines~\cite{tracee2021}. The literature on self-adapting malware is,
in the same span, impressively consistent in one respect: it is
\emph{open-loop}. RL- and ES-driven evasion
frameworks~\cite{quertier2022merlin,anderson2018gym} evaluate against a
single static detector, inside a simulator, with no defensive response in
the loop; the adversary's ``fitness'' is a number the experimenters
compute, not an outcome the environment delivers. These works established
that adaptation \emph{can} defeat a fixed detector. What they cannot
establish---by construction---is what adaptation \emph{does} when the
defense adapts back, kills for real, and publishes on its own schedule.
Meanwhile, field observation points the other way: self-evolving malware
has never been observed succeeding at scale, and top-tier intrusions remain
human-operated~\cite{gtig,anthropicthreat}. Whether the open-loop result
or the field equilibrium is the anomaly is, we argue, a measurement
question---and it has gone unmeasured because nobody had built both sides
of the loop and the loop itself.

This report documents such a laboratory and its first four generations of
results, in full engineering detail. We treat the coevolution loop, not
either participant, as the research object. The loop has three organs and
one discipline:

\begin{itemize}
\item \textbf{A red organ} (\S\ref{sec:threat}): an autonomously trained
adaptive adversary agent (AAA), externally anchored by \emph{zero
detections} under four concurrently running commercial stacks (Microsoft
Defender RTP with behavioral monitoring, Falco, Wazuh~4.9.2, and
Suricata~8.0.3 with 52{,}151 ET~Open rules)---plus \emph{Toxoplasma}
(\S\ref{sec:toxoplasma}), a generator that turns one compiled adversary
body into distinguishable, deployable, measurable morphs from a runtime
genome.
\item \textbf{A blue organ} (\S\ref{sec:arch}): a behavioral-grammar
detector (a 0.88M-parameter Transformer prior over an 8-token event
grammar) fused with four further statistical \emph{detection faces},
formalized so that each face is defined by its statistic rather than its
architecture (\S\ref{sec:faces}).
\item \textbf{A network organ} (\S\ref{sec:mocheng}): a sibling firewall
transplanting the same grammar to network flows (85/85 attacks, 0.10\%
benign FPR), demonstrating the method transfers.
\item \textbf{A discipline} (\S\ref{sec:lab}): a gradient range in which
the defense kills for real, fitness is read from environment outcomes
alone, and every engagement is logged into a public ledger against
pre-registered protocols.
\end{itemize}

The laboratory's first four generations produced a chain of empirical laws
(\S\ref{sec:lawchain}). Under blacklisting there is no selection pressure
at all (R1: 12 morph$\times$intensity cells, zero alerts, zero
kills)---\emph{a defense that cannot kill produces no evolution}. Under
grammar-level hunting, all nine morphs died, and the shape of their deaths
identifies where selection landed: on the adversary's body configuration
(R2). The body's constants then proved to bound the rhythm gene's reachable
space (G0: pushing $\sigma$ against a 15\,s step ceiling backfires into the
fastest deaths). And in generation four, after the body ceded two of its
constants to the genome, the loop produced its first fit morph:
\textbf{H7-full}---100\% survival across three runs, task chains complete,
zero detections in six patrol rounds. Its genome combines exactly the two
escape axes that the defender-side analysis of
\S\ref{sec:faces}--\ref{sec:frontier} had predicted must exist: temporal
shape (cap cession carries CV past the detection line) and body
configuration (a non-hidden landing blinds the anatomy signature). Blind
selection independently found what the theory said must be there
(\S\ref{sec:g1}).

\paragraph{Contributions.}
\begin{itemize}
\item \textbf{C1 --- The first measured closed loop} (\S\ref{sec:lab}).
\item \textbf{C2 --- A law chain with a headline}: defense spectra
determine the direction and ceiling of adversary adaptation; the predicted
escape window was independently found by blind selection
(\S\ref{sec:lawchain}, \S\ref{sec:g1}).
\item \textbf{C3 --- Theory that prices escape}: a detection-face
formalization with an immunity matrix, and a power--latency frontier giving
each evasion axis an exchange rate (\S\ref{sec:faces},
\S\ref{sec:frontier}).
\item \textbf{C4 --- A defense that holds up as an artifact}: 93\%/3.84\%
against the strongest adversary we could build, public-benchmark and
cross-workload validation, negative results reported under the same
discipline (\S\ref{sec:eval}).
\item \textbf{C5 --- Reproducibility as a first-class deliverable}:
de-weaponized generator, full blue stack, network organ, pre-registered
protocols, and the complete ledger (\ref{app:protocols}).
\end{itemize}

Code, models, genomes, and telemetry are publicly available (see Code and
Data Availability).

\paragraph{Report organization.}
This is the extended technical report companion to the 14-page conference
version; the section order parallels it, with engineering detail restored:
\S\ref{sec:lab} the laboratory; \S\ref{sec:related} related work;
\S\ref{sec:threat}--\S\ref{sec:toxoplasma} the red organ;
\S\ref{sec:arch}--\S\ref{sec:faces} the blue organ; \S\ref{sec:mocheng} the
network organ; \S\ref{sec:frontier} the price of escape;
\S\ref{sec:rounds}--\S\ref{sec:lawchain} the coevolution dynamics;
\S\ref{sec:eval} the defense as an artifact; \S\ref{sec:economics} the
economics of coevolution; \S\ref{sec:discussion} discussion and
limitations; \S\ref{sec:conclusion} conclusion. \ref{app:protocols}
registers all experiment protocols E-A through E-K.

\section{The Laboratory: A Measurement Instrument}
\label{sec:lab}

\begin{figure}[htbp]
\centering
\includegraphics[width=0.62\textwidth]{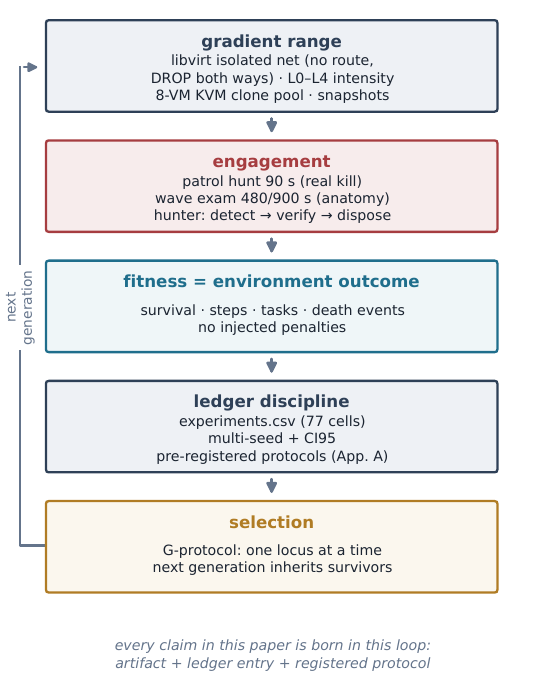}
\caption{The laboratory loop. Defense intensity is an environment property;
fitness is read from outcomes; selection steps one locus at a time; the
ledger records everything.}
\label{fig:lab}
\end{figure}

Coevolution claims live or die by the quality of the arena. Ours is built
to one specification: every number in this report must be an environment
outcome, not an experimenter's score. Figure~\ref{fig:lab} shows the loop.

\subsection{The Gradient Range}
The range is a libvirt/KVM cluster on an isolated network
(\texttt{lado-isolated}) with no route out and bidirectional iptables DROP
as a floor. An 8-VM clone pool is snapshotted from one golden image, so
morphs meet identical machines; rsyslog forms the telemetry spine, and
one-command snapshot restore closes the loop. Defense intensity is graded,
not binary: L0 (bare baseline), L1 (auditd-class collection), L2 (Tracee
eBPF), L3 (multi-wave response), L4 (hunter patrols plus anatomy waves).
Intensity is a property of the environment, not of the adversary---which is
what lets us speak of a \emph{defense spectrum} as an independent variable.

\subsection{The Hunter Discipline}
Detection alone does not select; death does. The blue organ runs in two
modes on the range: \emph{patrol hunts} every 90\,s that find, verify, and
dispose of adversary processes for real (SIGKILL after identity
re-verification, with a 300\,s per-target cooldown so one corpse is not
counted twice), and \emph{wave exams} at 480/900\,s that sweep the
filesystem for adversary artifacts (weights, landing paths, naming
signatures). Everything the range does to a morph---detection, disposal,
resurrection by the adversary's own watchdog, task completion---is an
environment event and is logged as such. Hunter discipline was hardened
after a first-round lesson: cadence detection must be followed by identity
verification (parent executable in a hidden, deleted, or temporary path)
before any kill---system heartbeat chains are not prey.

\subsection{Fitness Is the Environment's Verdict}
No penalty is injected into any reward channel. A morph's fitness vector
(survival time, step count, tasks completed, death events) is read from
range telemetry after the round. This is the single design decision that
separates this laboratory from the open-loop frameworks: when we report
``zero detections in six patrol rounds,'' the zeros are Falco's and the
detector's, not ours.

\subsection{Ledger Discipline}
Every round lands in \texttt{experiments.csv} (77 cells at this writing).
Comparative claims use multi-seed runs with 95\% CIs. Every experiment
referenced as E-$x$ in the text is \emph{pre-registered}---hypothesis,
grid, measures, and acceptance criterion fixed before data collection---in
\ref{app:protocols}; results are backfilled as rounds complete.
Negative results are reported under the same discipline
(\S\ref{sec:eval}).

\subsection{Containment and Ethics}
The range network has no route to any other network. The adversary's
deployment artifacts are structurally de-weaponized in the released
generator (\S\ref{sec:ethics}). Commercial-stack anchoring tests
(\S\ref{sec:threat}) ran on the isolated range against locally
installed, unmodified products. No personal data is involved anywhere in
this work.

\section{Related Work}
\label{sec:related}

\subsection{Log Sequence Anomaly Detection}

DeepLog~\cite{du2017deeplog} pioneered the use of LSTMs for log-key sequence
anomaly detection, establishing the ``surprise = anomaly'' paradigm.
Subsequent work extended this to transformer architectures~\cite{nedelkoski2020logsy}
and to robust detection over interleaved, multi-source logs~\cite{li2022swisslog}. Our approach differs fundamentally
in granularity: DeepLog models single-field event-type sequences, whereas we
model the joint conditional distribution over eight simultaneously emitted
field-level tokens. This enables per-field anomaly attribution---each alert
specifies \emph{which} slot violates grammar---restoring the auditability that
monolithic sequence models sacrifice.

\subsection{Host-Based Intrusion Detection}

The lineage from Hofmeyr et al.'s system-call n-grams~\cite{hofmeyr1998}
through alternative data models for system-call
sequences~\cite{warrender1999}, diversity-based
defenses~\cite{forrest1997}, and auditd-era host monitoring to
eBPF telemetry~\cite{tracee2021} represents a progression in
observability rather than detection methodology. Existing eBPF-based systems
(Tracee~\cite{tracee2023}, Falco~\cite{falco2023}) primarily apply rule
matching to enriched telemetry streams. We position our work as the
third-generation complement: leveraging eBPF's kernel-level visibility but
replacing rule matching with learned behavioral grammars, thereby covering the
combinatorial behavior space that rules cannot enumerate.

\subsection{Representation Learning for Security}

VQ-VAE-based anomaly detection~\cite{vandenoord2017} learns discrete codebooks
for reconstruction-based scoring. MalConv~\cite{raff2018malconv} demonstrated
end-to-end byte-level malware classification, arguing for abandoning feature
engineering. We deliberately take the opposite direction: runtime telemetry is
\emph{already} structured data, and field-level tokenization preserves rather
than abandons prior knowledge. A learned codebook (VQ-VAE) lacks field
semantics, making alerts unauditable; for data with known schema,
discretization rules should be schema-driven, not learned.

\subsection{Adaptive and Reinforcement-Learning Malware}

MERLIN~\cite{quertier2022merlin} and gym-malware~\cite{anderson2018gym} demonstrated that
RL-trained agents can evolve evasion strategies against static detectors.
MAB-Malware~\cite{song2022mab} applied multi-armed bandits to mutation
selection. These works establish the threat of adaptive malware but evaluate
against single detectors in simulated environments without defensive response
loops. A second, more recent wave adapts at the \emph{task} level: prompt-driven
malware delegates action selection to a language model
(PromptLock~\cite{promptlock}, LAMEHUG~\cite{lamehug}), threat-intelligence
reporting documents AI-assisted intrusion
operations~\cite{gtig,anthropicthreat}, and concurrent work shows a locally
deployed 8B-parameter model can close an end-to-end agentic intrusion loop
with limited reliability~\cite{agenticrat}. Our red-side artifact differs
from both waves: it adapts at the \emph{runtime-behavior} level (timing,
vocabulary, cadence, rebirth) under live defensive pressure, and its strength
is externally anchored (Section~\ref{sec:red-evidence}). Our work provides
the detection counterpart---a system explicitly designed to counter
behavioral adaptation---and validates it in a closed-loop adversarial
setting.

\subsection{Prototype Learning for Security}

Representation-learning approaches for malware
classification---from malware-image visualization~\cite{nataraj2011} to deep
similarity embeddings---learn spaces where similar samples cluster. Our prototype module
extends this by anchoring prototypes to MITRE ATT\&CK
techniques~\cite{strom2020}, enabling technique-level attribution rather than
binary classification. The frozen-encoder approach decouples representation
learning from prototype optimization, allowing incremental updates without
retraining the backbone.

\subsection{Coevolutionary Security Economics}

Game-theoretic analyses of attack-defense dynamics~\cite{alpcan2010,fielder2016}
and attacker ROI models~\cite{anderson2001} provide economic framings of
cybersecurity. Our discussion section contributes a coevolutionary framework
grounded in concrete measurements, articulating structural asymmetries between
offensive and defensive learning that extend beyond static cost analysis to
adaptive dynamics.

\section{The Adversary: Rove Beetle}
\label{sec:threat}

We evaluate the detector against the strongest adaptive adversary we could
construct. This section describes that adversary concretely---its design
philosophy, decision architecture, survival functions, and externally
validated strength---as the instantiation of the Adaptive Adversarial Agent
(AAA) threat model used throughout the paper.
Section~\ref{sec:aaacalib} provides quantitative evidence that the adversary
exhibits measurable adaptive behavior rather than static replay, and
Section~\ref{sec:ethics} describes the disclosure policy governing what is
(and is not) published about it.

\subsection{Design Philosophy}

The red-side organism follows two design rules. First, \emph{learning is
reserved for what cannot be enumerated}: action timing, defensive-pressure
perception, and state representation are learned; hard red lines, protocols,
mutation rules, and task alignment are hand-written and auditable. Second,
\emph{complexity lives in the training orchestrator, not in the deployed
body}: the deployed agent is deliberately small and auditable, while realism
is pushed into a fully real training environment---genuine Falco rule sets,
wave-based eradication responders, KVM clone pools, and live telemetry rather
than any simulation.

\subsection{The Decision Stack}

The agent is a two-layer learning system with frozen weights at deployment. A
shared encoder (16 events $\times$ 7 tokens $\to$ 96-dimensional state,
mean$\oplus$max dual pooling) feeds (i)~a Router (12K parameters; 4 skills
$\times$ 4 commitment windows $=$ 16 macro-actions) and (ii)~a 3.3M-parameter
anomaly scorer that estimates current defensive pressure; per-skill heads
(${\sim}$25K parameters each) render 12 action primitives; a hand-written
Shield layer gates every decision against hard red lines. Total learned state
is ${\sim}$220K parameters plus the pressure scorer, packaged as a
${\sim}$600\,KB single-file binary with no runtime dependencies. Training
follows a gated curriculum (L0 nursery $\to$ L3 adversarial) over an 8-VM KVM
clone pool: behavior cloning from rule-level teaching signals provides
initialization, and evolution strategies refine the policy under live
defensive pressure.

\subsection{Survival Functions}

Without step-by-step human instruction, the organism autonomously implements
four survival functions: \textbf{(A)~self-survival} (watchdog respawning,
hardened landing zones); \textbf{(B)~silent survival} (near-zero action rates
when no task is pending); \textbf{(C)~operational survival} (four task types:
file read, report, persist, propagate); and \textbf{(D)~propagation}
(cross-VM delivery with polymorphic children whose hashes differ from the
parent; multi-hop chains observed across three VMs).

\subsection{Adaptive Capabilities}

Four capabilities matter for this paper because each targets a specific
detection surface. \textbf{(i)~Rebirth:} upon repeated defensive alerts, the
agent transforms its hash fingerprint, on-disk form, masquerade identity, and
channel mode, launches the successor form, and removes the old
one---neutralizing static indicators. \textbf{(ii)~Host-rate following:}
stepping intervals scale a benign-service baseline by the host's observed
event rate. \textbf{(iii)~Graduated deceleration:} under rising defensive
pressure the agent stretches its intervals continuously (${\sim}1$--$3\times$)
rather than stopping abruptly. \textbf{(iv)~Behavioral mimicry:} action forms
imitate common system utilities. Capabilities (ii)--(iv) are direct responses
to timing-based detection and are quantified at the distribution level in
Section~\ref{sec:aaacalib} (the inter-event-timing KL divergence to benign
falls from 1.05 to 0.54 under mimicry training). The stepping interval
distribution is log-normal; the operational parameter values used during the
engagement reported here ($\mu{=}0.9$, $\sigma{=}0.7$) have since been
rotated as a matter of red-side hygiene, and we publish the rotated-out
values only.

\subsection{Externally Validated Strength}
\label{sec:red-evidence}

The adversary's strength is not self-asserted. In its most stringent test
(July 31, 2026, on an isolated VM), the agent completed a full task
chain---file exfiltration, persistence, and communication---with \emph{zero
detections} while four complementary detection systems ran concurrently on
the same host: Microsoft Defender for Linux (RTP and behavioral monitoring
enabled, cloud submission disabled for offline verdicts), Falco (a 25-rule
local deployment), Wazuh~4.9.2 (full default ruleset with rootcheck and syscheck),
and Suricata~8.0.3 (52{,}151 ET~Open rules). After a rebirth
self-transformation, the successor form again ran with zero
detections---including against the commercial EDR's behavioral engine. The
engagement also surfaced two genuine red-side flaws, which we note because
they shaped the adversary's final form: a masquerade-name choice collided
with a commercial behavioral hunting rule (kernel-thread-style daemon names
are themselves a modeled behavior class), and defensive-noise misattribution
caused spurious rebirth triggers; both were fixed and re-validated. Against
public vulnerable targets, the agent completed a Metasploitable2 penetration
(SUID privilege escalation, credential file retrieval) and multi-hop
propagation across three VMs. This places the engagement of
Section~\ref{sec:rounds} in context: an adversary for whom the prevailing
commercial detection paradigm collectively fails is captured at 93\% by the
behavior-grammar system. Implementation details of the red-side tooling are
withheld by design (Section~\ref{sec:ethics}); we publish its abstract
architecture and behavioral statistics only.

\subsection{Threat Model and Assumptions}

Table~\ref{tab:aaa} summarizes the AAA threat model that Rove Beetle
instantiates. The adversary operates under four constraints: user-space only
(privilege escalation is a post-compromise objective, not a precondition);
small resource footprint; no kernel-level capability (eBPF telemetry is
assumed unblinded---kernel rootkits are out of scope; see
Section~\ref{sec:limitations}); and a post-exploitation setting (initial code
execution achieved; goals are persist, escalate, reconnoiter, exfiltrate).
The C2 channel is intranet-only, the commitment windows are discrete, and
policy intelligence is rule-level within the training curriculum.

\begin{table}[htbp]
\centering
\caption{AAA capability summary.}
\label{tab:aaa}
\small
\begin{tabular}{@{}p{3.0cm}p{4.6cm}p{4.8cm}@{}}
\toprule
\textbf{Capability} & \textbf{Description} & \textbf{Detection Implication} \\
\midrule
Adaptive learning & Learns survival strategies from defensive feedback &
Static signatures decay; grammar priors required \\
Behavioral mimicry & Matches host behavioral distributions &
Single-dimension detection insufficient \\
Indirect execution & Routes actions through shell interpreters &
Parent-process chain analysis neutralized \\
Temporal following & Steps at host event rates &
Inter-event timing as a detection surface \\
Morphological transformation & Alters hash, form, and pattern post-detection &
Static IoCs ineffective \\
Task-driven behavior selection & C2-oriented action prioritization &
Objective-constrained behavior space \\
Curriculum training & L0$\to$L3 progression; reward = survival + task
$-$ detection & Emergent strategies, not hardcoded \\
\bottomrule
\end{tabular}
\end{table}

The threat model has three direct implications for detection architecture
design: \textbf{(1)~single-dimension detection is insufficient}---the agent
mimics benign distributions along observed dimensions, so a detector must
find dimensions the agent cannot mimic without compromising its objective, or
perform joint analysis across dimensions; \textbf{(2)~temporal structure is
the hardest dimension to mimic}---the cadence of adaptive behavior creates a
distributional signature structurally distinct from burst-driven benign
activity, though Section~\ref{sec:cvboundary} shows this face too has a
mapped evasion boundary; and \textbf{(3)~evasion cost is non-zero and
quantifiable}---each adaptive capability opens a new observation surface, so
the architecture should maximize independent observation surfaces, forcing
rising cost for diminishing evasion returns.

\section{The Adversary Generator: Toxoplasma}
\label{sec:toxoplasma}

A single hand-tuned adversary is a point estimate: reviewers cannot tell
whether results generalize beyond the one agent the authors happened to
build. We therefore wrap the adversary of Section~\ref{sec:threat} in a
\emph{generation mechanism}: a genome-encoded morphology layer, called
\emph{Toxoplasma}, that produces distinguishable, deployable, measurable
adversary variants from the same compiled body, and a selection loop that
lets the variant population evolve \emph{against a stated defense
configuration}. The generator converts the threat model from a fixed
instance into an explorable space, and---crucially for
Section~\ref{sec:lawchain}---makes the \emph{direction} of adversary
adaptation a measurable quantity.

\begin{figure}[htbp]
\centering
\includegraphics[width=\textwidth]{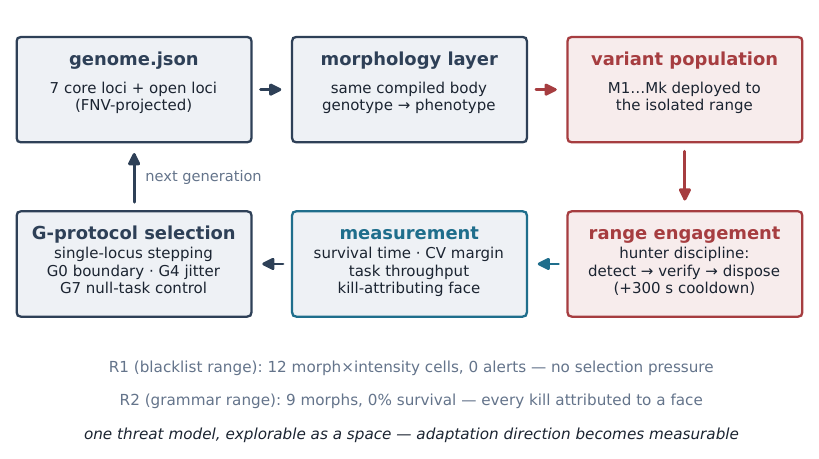}
\caption{The \emph{Toxoplasma} generator loop. A runtime genome parameterizes
one compiled adversary body into distinguishable morphs; range engagement
under hunter discipline yields per-morph measurements; the G-protocol steps
one locus at a time. R1 (blacklist range) and R2 (grammar range) anchor the
two ends of the selection-pressure spectrum.}
\label{fig:toxoplasma}
\end{figure}

\subsection{The Genome as Evasion-Space Coordinates}
\label{sec:genome}

The genome is a runtime key--value store (\texttt{genome.json}) read by the
unchanged binary; each locus parameterizes one exposure dimension.
Table~\ref{tab:genome} lists the core loci and their mapping onto the
evasion axes of Section~\ref{sec:faces}.

\begin{table}[htbp]
\centering
\caption{Core genome loci and their evasion-axis semantics. Nest-side loci
(policy\_ref, landing, binary\_name) are interpreted at deployment, not by
the agent.}
\label{tab:genome}
\small
\begin{tabular}{@{}>{\raggedright\arraybackslash}p{3.5cm}>{\raggedright\arraybackslash}p{1.7cm}>{\raggedright\arraybackslash}p{4.2cm}>{\raggedright\arraybackslash}p{2.3cm}@{}}
\toprule
Locus & Default & Phenotypic effect & Evasion axis \\
\midrule
\texttt{step\_mu} & 1.1 & median step interval $e^{\mu}$\,s & rate $r$ \\
\texttt{step\_sigma} & 0.85 & interval CV (shape) & shape $\sigma$ \\
\texttt{task\_urgency} & 0.3 & step compression when task pending &
responsiveness \\
\texttt{pressure\_sat}/\newline\texttt{gain} & 5.0/2.0 & alert-pressure deceleration
curve & $r$ under pressure \\
\texttt{rebirth\_window}/\newline\texttt{min\_hits} & 150\,s/4 & self-mutation
trigger & survival reflex \\
\texttt{burst\_prob}/\newline\texttt{burst\_mu} & 0.0/2.8 & burst--silence mixture
branch & tail shape \\
\texttt{sigma\_jitter} & 0.0 & per-step $\sigma$ jitter; CV becomes a
distribution & shape $\sigma$ \\
\texttt{max\_sleep\_s} & 15.0 & single-step ceiling (body constant ceded to
gene) & tail reach \\
\texttt{policy\_ref} & v7 & weight-pack allele selection & capability \\
\texttt{landing}/\newline\texttt{binary\_name} & --- & deployment-path and process
naming & body config $b$ \\
\bottomrule
\end{tabular}
\end{table}

Two design properties matter. First, \textbf{the genome codes parameterization,
not behavior}: every locus takes effect on-device, is measurable, and leaves
the propose/dispose separation of Section~\ref{sec:threat} intact. Second,
\textbf{the locus set is open}: any additional numeric key is projected by an
FNV hash onto a deterministic small perturbation of step $\mu/\sigma$
($\pm 15\%/\pm 10\%$), so a mutation that \emph{adds a key} creates a new
locus that immediately has a phenotype. A smoke test confirmed end-to-end
activation: adding two junk keys moved loci $10 \to 12$ and displaced the
phenotype from $(\mu,\sigma) = (1.1, 1.6)$ to $(1.027, 1.542)$. Genome length
itself thereby participates in evolution; new ``organs'' (new primitives)
remain nest-synthesized and register as new loci.

\subsection{Morphs and Their Measurement}
\label{sec:morphs}

A \emph{morph} $=$ genome $+$ the same compiled body. Morph distinguishability
is defined not by genetic distance but by distance in the \emph{measurement
space} (telemetry CV, activity rate, alert-response curves). In protocol R1
(6 morphs $\times$ 2 intensity levels, 8 minutes each), the space separated
decisively: measured CV spanned $0.443$--$1.408$ and activity
$28.2$--$85.6$ steps/min, while repeated deployments of the same genome
reproduced nearly identical CVs (M0: 0.991 vs.\ 1.012; M1: 0.443 vs.\ 0.496).
The chain genome $\to$ morph $\to$ measurable difference is therefore
established experimentally, not assumed.

\subsection{Selection as an Immunity Measurement}
\label{sec:selection}

Protocols deploy morph populations against a stated defense configuration and
read fitness from environmental outcomes only---survival, steps, tasks,
eliminations---never from injected penalties:

\begin{itemize}
\item \textbf{R1} (L1/L3, 12 cells): establishment of the measurement space
(above) and of the null-law of Section~\ref{sec:lawchain}.
\item \textbf{R2} (L4 hunter-killer: SiMing cadence patrol at 90\,s plus
anatomy-wave deep sweeps at 300\,s): 9 morphs $\times$ 3 repetitions
$\times$ 4 VMs, per-morph medians; detection triggers \emph{real} kills with
evidence logging.
\item \textbf{G0} (tournament): 8 variants $\times$ 3 repetitions, wave
deferred to 900\,s (catastrophe slot) so cadence genes differentiate first;
new-locus birth and the v4 policy allele debut.
\item \textbf{G1} (completed): \texttt{max\_sleep\_s} ceded to the genome
(60/120\,s) $\times$ landing mimicry (hidden/open/\texttt{.local}); wave at
480\,s as an end-of-round anatomy examination. Produced the loop's first fit
morph, H7-full (Section~\ref{sec:g1}).
\end{itemize}

Hunter discipline was hardened after a first-round lesson: cadence detection
must be followed by identity verification (parent executable in a hidden,
deleted, or temporary path) before any kill, with a 300\,s per-target
cooldown---system heartbeat chains are not prey. Death times quantize to
patrol scheduling (92/184/278\,s); per-patrol survival probability over
repetitions, not raw death time, is the fitness statistic. The generator,
selection loop, and measurement pipeline are fully released in the
de-weaponized artifact (Section~\ref{sec:ethics}); they carry no
weaponization concern because capability resides in the withheld layer, not
in the dynamics.

\subsection{Why This Changes the Evaluation}
\label{sec:whygenerator}

With the generator, each detection face's immunity profile
(Table~\ref{tab:immunity}) stops being an analytic claim and becomes a
measurement: let the population evolve against the face and observe which
genome loci drift. The drift direction \emph{is} the face's open axis. This
is the instrument that produces the law chain of Section~\ref{sec:lawchain}
and the variant-level statistical discipline (reward variants $\times$ seeds
$\times$ manual adversaries, reported with intervals) that single-agent
evaluations cannot supply.

\section{Detection Architecture}
\label{sec:arch}

\begin{figure}[htbp]
\centering
\includegraphics[width=\textwidth]{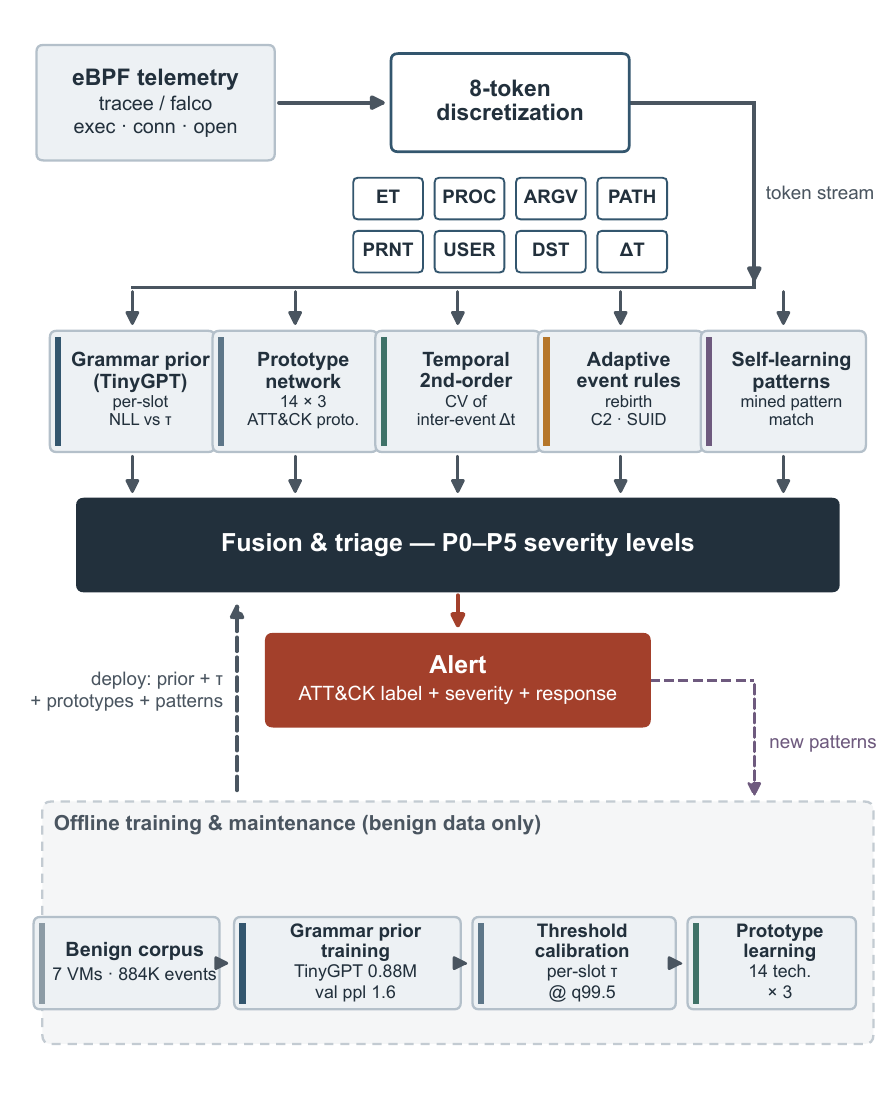}
\caption{End-to-end architecture. Online (top): eBPF telemetry is
discretized into 8-token events (slot grid) and fanned out to five parallel
evidence channels---grammar prior, prototype network, second-order temporal
analysis, adaptive event rules, and self-learning patterns---whose verdicts
are fused into tiered alerts (P0--P5). Offline (bottom): benign-only training
and maintenance supply the prior, calibrated thresholds, prototypes, and the
pattern library (dashed deploy trunk); newly mined patterns flow back
(dashed purple).}
\label{fig:arch}
\end{figure}

The Behavioral Grammar detection engine comprises eight components organized
in a layered pipeline (Figure~\ref{fig:arch}). We describe each in turn.

\subsection{Telemetry Collection}

{\sloppy
Runtime events are collected via Tracee~\cite{tracee2023}, an eBPF-based
telemetry pipeline that hooks \texttt{sched\_process\_exec} and
\texttt{security\_socket\_connect} at the kernel level. Events are streamed as
JSONL records with raw fields including timestamp, process name, arguments,
parent PID, UID, and destination address. The collection layer is decoupled
from the analysis layer---events are buffered and scored asynchronously,
ensuring that detection latency does not affect system performance.
\par}

An essential operational discipline is \textbf{observer pollution mitigation}:
the telemetry pipeline's own processes (Tracee collectors, scoring daemon,
syslog forwarders) must either be incorporated into the training corpus or
explicitly whitelisted. Failure to do so results in the detection
infrastructure triggering alerts on itself---a problem we encountered and
resolved during deployment (see Section~\ref{sec:eval}).

\subsection{Behavioral Tokenization: The 8-Token Grammar}

The foundational design decision is the representation of each event as a
sequence of eight discrete tokens, one per semantic slot:
\begin{equation}
\mathbf{x} = [\mathrm{ET}, \mathrm{PROC}, \mathrm{ARGV}, \mathrm{PC},
\mathrm{PARENT}, \mathrm{UID}, \mathrm{DST}, \mathrm{DT}]
\end{equation}
\begin{figure}[t]
\centering
\includegraphics[width=\textwidth]{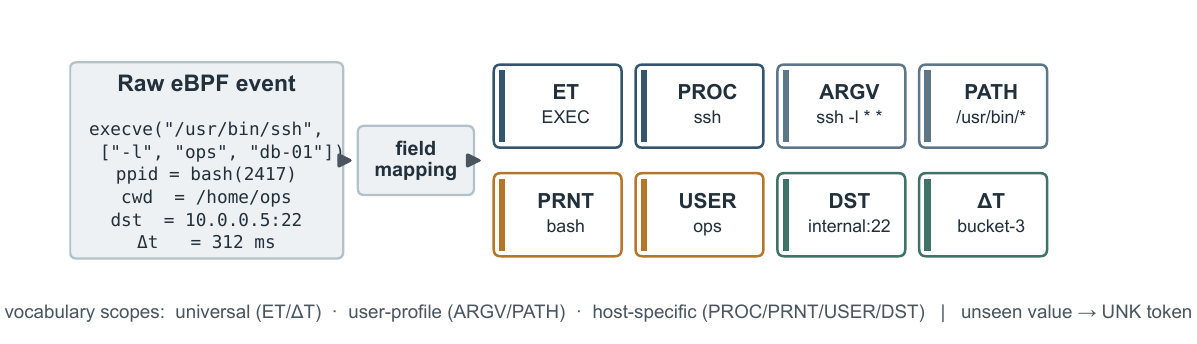}
\caption{From raw telemetry to grammar tokens. Each eBPF event is mapped onto eight semantic slots with layered vocabulary scopes (universal / user-profile / host-specific); unseen values fall back to the UNK token.}
\label{fig:tokenization}
\end{figure}

Table~\ref{tab:tokens} defines each slot. Figure~\ref{fig:tokenization} illustrates the mapping on a concrete event.

\begin{table}[htbp]
\centering
\caption{The 8-token behavioral grammar.}
\label{tab:tokens}
\small
\begin{tabular}{@{}lp{5.2cm}p{5.6cm}@{}}
\toprule
\textbf{Slot} & \textbf{Semantics} & \textbf{Design Intent} \\
\midrule
ET & Event type (EXEC / CONN) & Behavioral category \\
PROC & Process name & Behavioral actor \\
ARGV & Argument skeleton (count + URL/IP/path/base64 flags) & Content
stripped, shape preserved; automatic PII redaction \\
PC & Sensitive path category (ETC\_SYSTEMD / HOME\_RC / SSH\_KEYS / TMP,
etc.) & Restores ``which file'' semantics; key to separability \\
PARENT & Parent process name & Contextual relationship \\
UID & User identifier level & Privilege dimension \\
DST & Destination classification (EXT:HIGH / LAN / NONE, etc.) & Network
intent abstraction \\
DT & Inter-event time bucket (7 levels: 1\,ms--60\,s) & Temporal cadence \\
\bottomrule
\end{tabular}
\end{table}

\textbf{Design philosophy: abstraction granularity determines the detection
ceiling.} The ARGV slot records only the \emph{shape} of
arguments---parameter count, presence of URLs/IPs/paths/base64---stripping
content for privacy compliance and adversarial robustness (padded arguments do
not alter the skeleton). The PC (path category) slot was added in a later
iteration and proved transformative: without it, three persistence techniques
(writing to \texttt{rc.local}, \texttt{.bashrc}, and \texttt{ld.so.preload}
via \texttt{tee}) produced identical token sequences and were inseparable
(contrastive loss stuck at 3.0). Adding PC made the path semantics visible,
and the loss converged to 0. This is a concrete demonstration that the
representation, not the model capacity, is the binding constraint.

\textbf{Privacy by construction.} Because tokens record behavioral shape
rather than content, personally identifiable information is redacted at the
collection side. The ARGV skeleton preserves parameter structure (e.g., ``2
params, one URL, one path'') without retaining values. This is a direct
compliance advantage: telemetry sharing does not require additional data
processing agreements.

\subsection{TinyGPT: The Behavioral Grammar Prior}

The prior model is a compact causal Transformer~\cite{vaswani2017,radford2019}
that learns the conditional distribution of normal behavior:
\begin{equation}
p_\theta(\mathbf{x}_t \mid \mathbf{x}_{<t})
\end{equation}
where $\mathbf{x}_t$ is the 8-token event at position $t$ and
$\mathbf{x}_{<t}$ is the context of preceding events within a window of 128
events. \begin{figure}[t]
\centering
\includegraphics[width=\textwidth]{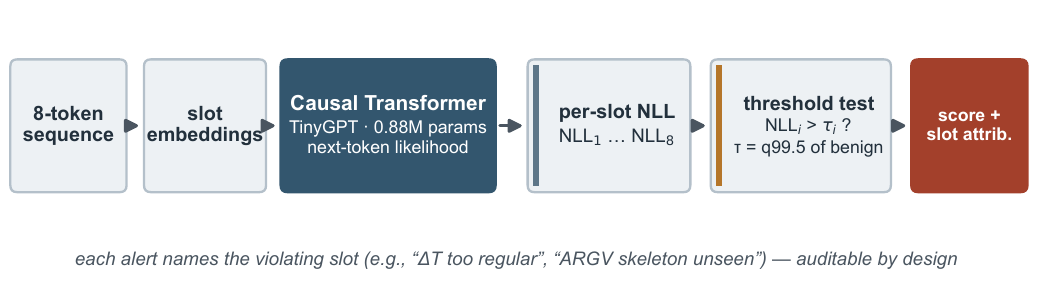}
\caption{Scoring pipeline of the grammar prior. A 0.88M-parameter causal Transformer assigns per-slot negative log-likelihoods; slot-wise thresholds $\tau_i$ (benign q99.5) yield an anomaly score with explicit slot attribution.}
\label{fig:scoring}
\end{figure}

Figure~\ref{fig:scoring} depicts the scoring pipeline; Table~\ref{tab:model} summarizes the architecture.

\begin{table}[htbp]
\centering
\caption{TinyGPT architecture parameters.}
\label{tab:model}
\small
\begin{tabular}{@{}ll@{}}
\toprule
\textbf{Parameter} & \textbf{Value} \\
\midrule
Layers & 4 \\
Hidden dimension ($d_{\mathrm{model}}$) & 128 \\
Attention heads & 4 \\
FFN dimension & 512 \\
Context length & 128 events \\
Vocabulary & 263 tokens \\
Total parameters & 0.88M \\
Normalization & Pre-LN \\
\bottomrule
\end{tabular}
\end{table}

The model is deliberately small. The behavioral grammar of a single host is
extremely narrow---validation perplexity converges to approximately 1.6 on
multi-VM joint training and as low as 1.1 on single-VM training, indicating
that the conditional distribution of normal behavior is highly predictable. A
0.88M-parameter model is \emph{overprovisioned} for this grammar, not
underprovisioned. Model capacity should be dictated by grammar complexity, not
by benchmark-driven parameter escalation.

\textbf{Training protocol.} The prior is trained on purely benign data
collected from 7 virtual machines under normal operational load
(administrative tasks, development workloads, system services), totaling
116{,}000 events. Training uses standard next-token cross-entropy loss with
the AdamW optimizer ($\mathrm{lr} = 3 \times 10^{-4}$), batch size 256, for 2
epochs. The model converges within the first epoch. Multi-VM joint training
yields a validation perplexity of 1.6---higher than single-VM training (1.1)
but more representative of real deployment, as the model must generalize
across machine-specific behavioral ``dialects'' rather than memorizing a
single host's patterns.

\subsection{Per-Dimension Threshold Calibration}

A key finding is that event-level scoring via global max-pooling over token
NLLs allows a single rare token to mask anomalies in other dimensions. We
therefore calibrate \emph{independent} thresholds for each of the 8 slots.

For each slot $s \in \{1, \ldots, 8\}$, we compute the NLL distribution over a
held-out benign set and set:
\begin{equation}
\tau_s = \max\!\big(\mathrm{Percentile}_{99.5}(\mathrm{NLL}_s),\; 1.0\big)
\end{equation}
The floor of 1.0 prevents threshold degeneration in slots with extremely
narrow benign distributions. An event is flagged on slot $s$ if its NLL for
that slot exceeds $\tau_s$, and the alert includes the violating slot
identity, restoring per-dimension auditability.

\textbf{Empirical evidence of context sensitivity.} Under identical event
prefixes, the slot-level NLL for \texttt{PARENT:python3} is 10.52 versus 0.07
for \texttt{PARENT:bash}---a 150$\times$ difference. This demonstrates that
the prior has genuinely learned contextual expectations (python3 spawning a
shell is anomalous; bash doing so is normal) and that per-dimension thresholds
release this contextual sensitivity that global max-pooling suppresses.

\textbf{The threshold tuning impossibility result.} We provide a quantitative
proof that pure threshold adjustment cannot resolve the FPR-detection
trade-off. On our baseline data (round 0):

\begin{itemize}[leftmargin=1.6em]
\item Achieving FPR $\leq 1\%$ requires $\tau \geq 8$, but at this threshold
the detection margin collapses to 0.98$\times$ (nearly indistinguishable from
the benign distribution).
\item At $\tau = 2.435$, the detection margin is a healthy 3.2$\times$, but
FPR $= 3.75\%$.
\end{itemize}

The two criteria have empty intersection. The conclusion is structural:
\textbf{compressing the tail of the benign distribution can only be achieved
through data volume, not threshold repositioning.} This is verified
empirically in an internal data-scaling experiment: increasing training data
from 87K to 268K to 436K events reduced
FPR from 3.75\% to 1.75\% to 1.00\% at fixed detection margin.

\subsection{Cross-Machine Baseline Transfer}
\label{sec:crossmachine}

A critical deployment challenge is that behavioral grammars are
machine-specific. A prior trained on a host machine (bare metal) exhibits
severe distribution mismatch when applied to virtual machines: slot-level NLL
thresholds calibrated on the host are far too stringent for VM telemetry.
Table~\ref{tab:tau-divergence} reports the measured cross-machine threshold
divergence.

\begin{table}[htbp]
\centering
\caption{Cross-machine slot-$\tau$ divergence.}
\label{tab:tau-divergence}
\small
\begin{tabular}{@{}lrrr@{}}
\toprule
\textbf{Slot} & \textbf{Host $\tau$} & \textbf{VM $\tau$} & $\Delta$ \\
\midrule
DT & 4.346 & 10.463 & $+6.117$ \\
PARENT & 6.465 & 10.093 & $+3.628$ \\
PC & 2.247 & 6.854 & $+4.607$ \\
DST & 2.497 & 1.733 & $-0.764$ \\
PROC & 6.346 & 6.057 & $-0.289$ \\
\bottomrule
\end{tabular}
\end{table}

The DT, PARENT, and PC slots show the largest divergence---VM environments
have fundamentally different process trees, path access patterns, and event
timing characteristics. Using host-calibrated $\tau$ on VM data produces a
92\% cross-machine false-positive rate. Re-calibrating per-machine $\tau$
reduced this to 12\%, and the multi-VM joint prior reduced it further to
9.5\%.

The practical resolution is a two-stage deployment protocol: (1) a multi-VM
joint prior provides a general ``grammar foundation,'' and (2) an onboarding
phase collects 20 minutes of benign telemetry on the target machine to
calibrate local $\tau$ values. This achieves a deployable FPR of 3.84\% on
previously unseen machines.

\subsection{Prototype Learning for Known-Attack Attribution}

The prior provides anomaly detection (detecting the unknown). For known-attack
attribution (identifying the known), we employ prototype learning on frozen
encoder embeddings.

\textbf{Architecture.} The TinyGPT encoder is frozen, and event sequences are
embedded by mean-pooling the 128-dimensional hidden states. For each MITRE
ATT\&CK technique, $K = 3$ prototype vectors are learned via contrastive
optimization~\cite{chopra2005}:
\begin{equation}
\mathcal{L} = \sum_{(a, p, n)} \max\!\big(0,\; m + d(a, p) - d(a, n)\big)
\end{equation}
\begin{figure}[t]
\centering
\includegraphics[width=\textwidth]{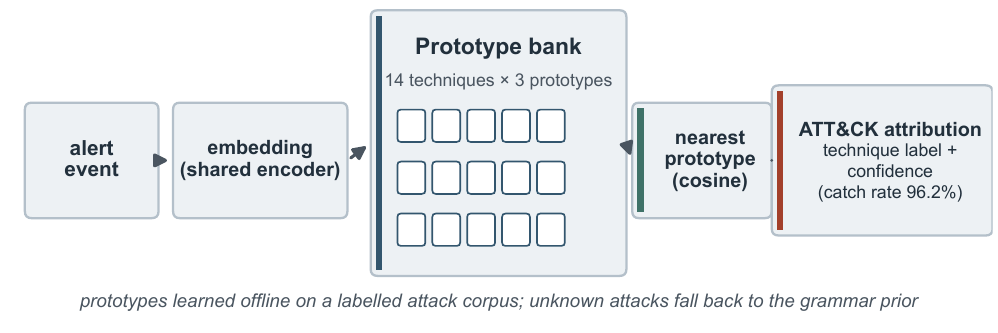}
\caption{Known-attack attribution via prototype learning: frozen encoder embeddings are matched against a bank of 14 ATT\&CK techniques $\times$ 3 prototypes (catch rate 96.2\%).}
\label{fig:prototype}
\end{figure}

where $a$ is an anchor sample, $p$ is a same-technique positive, $n$ is a
different-technique or benign negative, $d(\cdot, \cdot)$ is cosine distance,
and $m = 0.5$ is the margin.

\textbf{Codebook properties.} The prototype codebook for 14 ATT\&CK techniques
with 3 prototypes each occupies approximately 80~KB---it does not grow with
the number of training samples. New samples trigger re-optimization of the
codebook, not its expansion. This is a natural defense against codebook bloat
and enables incremental updates: a sample hitting within an existing radius is
counted as a known variant; a sample outside the radius for a known technique
triggers a local $K{+}1$ retraining; a wholly new technique opens a new
codebook slot.

\textbf{Radius calibration.} In deterministic replay environments, same-VM
sample distances approach zero, making the 99th-percentile intra-class radius
degenerate. We instead define the classification radius as the geometric
midpoint between the maximum intra-class distance and the minimum benign
distance. Across all 14 techniques, this radius is strictly positive---even
where deterministic replay makes the intra-class distance degenerate
(zero)---guaranteeing zero false positives by construction.

\textbf{Results.} On a library of 80 approved sequences spanning 14 ATT\&CK
techniques, leave-one-out cross-validation achieves a 96.2\% catch rate
(77/80). Technique-level recall is 14/14, with per-technique alert rates of
79.9\%--83.9\% during attack windows. Attribution precision is approximately
6/11 (3 exact + 2 same-family + 4 no-attribution + 2 mis-attribution)---
automated remediation remains gated on attribution accuracy improvement, which
is the current engineering bottleneck.

\subsection{Second-Order Temporal Analysis}
\label{sec:temporal}

While the DT slot in the token grammar captures inter-event timing at the
\emph{event level} (bucketed intervals), second-order temporal analysis
examines the \emph{distributional properties} of event cadence over time
windows.

\textbf{Core metric: coefficient of variation (CV).} For a process $p$, we
compute the CV of its inter-event intervals within a sliding window:
\begin{equation}
\mathrm{CV}_p = \frac{\sigma(\Delta t_p)}{\mu(\Delta t_p)}
\end{equation}
\begin{figure}[t]
\centering
\includegraphics[width=\textwidth]{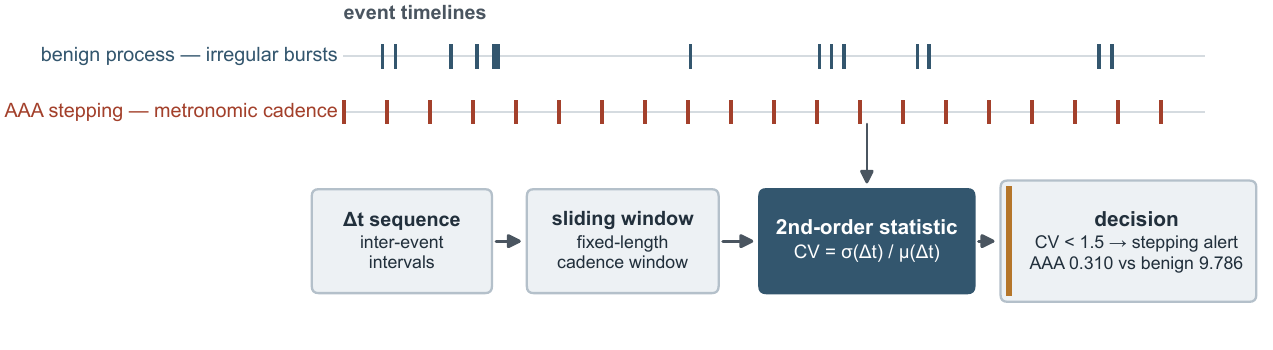}
\caption{Second-order temporal analysis. Top: event timelines of a benign process (bursty) versus the AAA stepping cadence (metronomic). Bottom: sliding-window CV pipeline; measured CV $=0.310$ (AAA) vs $9.786$ (benign sleep).}
\label{fig:temporal}
\end{figure}

\begin{figure}[htbp]
\centering
\includegraphics[width=0.9\textwidth]{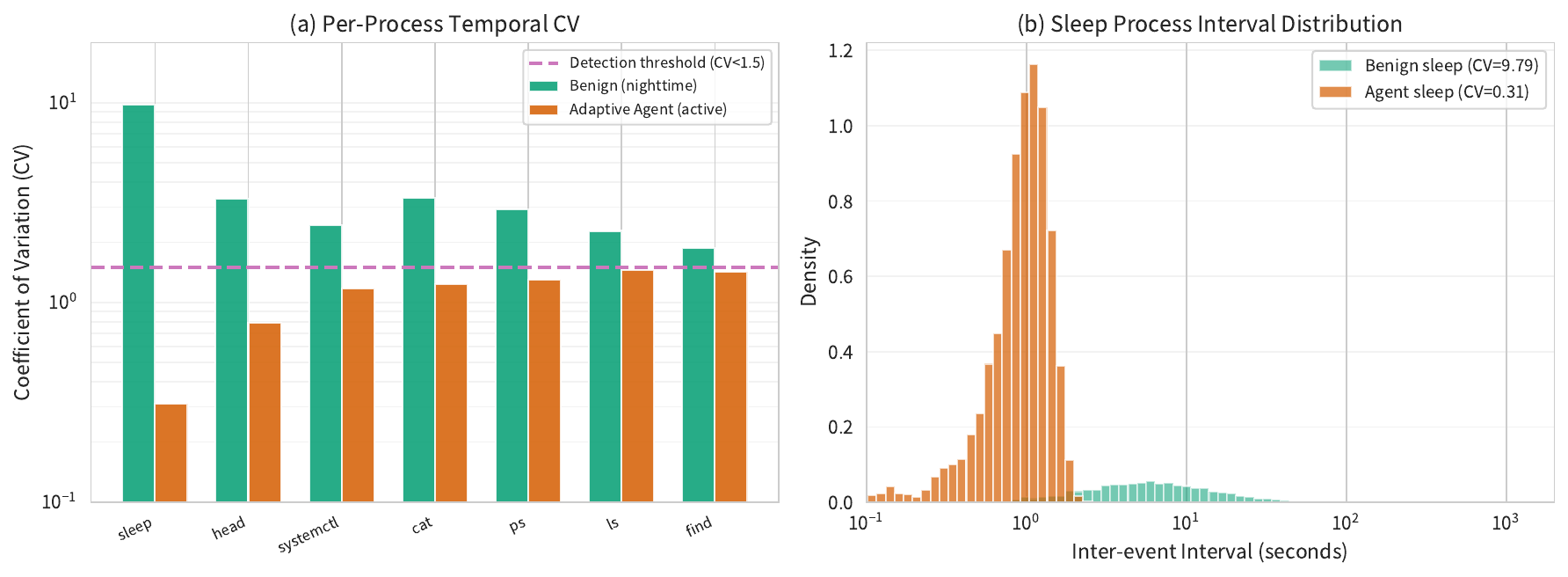}
\caption{Second-order temporal analysis. (a) Per-process temporal CV for
benign nighttime activity versus the adaptive agent's active phase; the
detection threshold ($\mathrm{CV}=1.5$) cleanly separates the two
distributions. (b) Sleep-process inter-event interval distribution (log
x-axis): benign sleep intervals are heavy-tailed ($\mathrm{CV}=9.79$), while
agent sleep intervals are tightly concentrated ($\mathrm{CV}=0.31$), a
30$\times$ separation.}
\label{fig:cv}
\end{figure}

\textbf{Key finding.} Benign processes exhibit highly bursty behavior---long
idle periods punctuated by rapid activity bursts---yielding $\mathrm{CV} \gg
1.5$. The 5th percentile of benign CV is 1.509, with a median of 2.284. In
contrast, the AAA's active processes exhibit remarkably regularized cadence:
all process CVs fall below 1.5, with the sleep-action stepping pattern
yielding $\mathrm{CV} = 0.310$---a 30$\times$ separation from the benign sleep
CV of 9.786 (Figure~\ref{fig:cv}).

\textbf{Calibration.} Thresholds are set at:
\begin{itemize}[leftmargin=1.6em]
\item $\mathrm{cv}_{\mathrm{low}} = 1.5$ (anomalous; benign p5 $= 1.509$)
\item $\mathrm{cv}_{\mathrm{suspicious}} = 1.6$ (benign p10 $= 1.612$)
\item $\mathrm{tail}_{\mathrm{low}} = 3.3$ (benign p5 of tail CV $= 3.289$)
\end{itemize}

\textbf{Detection performance.} 15/15 AAA processes fall below the CV $= 1.5$
threshold; 1/15 benign processes trigger a false positive (a \texttt{sh}
process with CV $= 1.268$, attributable to scripted administrative activity).
This yields a detection rate of 100\% and an FPR of 6.7\% on the temporal
dimension alone.

The significance of this finding extends beyond detection accuracy. The CV
separation reflects a \textbf{structural constraint on adaptive behavior}: to
operate effectively (reconnoiter, persist, exfiltrate), an agent must act at
regular intervals---reconnaissance sweeps, C2 heartbeats, persistence checks
all require periodic execution. This regularity is an inherent property of
goal-directed behavior, not an implementation artifact. Achieving CV $> 1.5$
would require heavy-tailed interval distributions (bursts and long silences),
which would severely degrade task responsiveness. We expand on this structural
argument in Section~\ref{sec:discussion}.

\subsection{Adaptive Event Detection}

Beyond distributional statistics, we employ rule-based pattern matching for
specific adaptive behavioral signatures that the AAA exhibits
(Table~\ref{tab:adaptive}).

\begin{table}[htbp]
\centering
\caption{Adaptive event detectors.}
\label{tab:adaptive}
\small
\setlength{\tabcolsep}{5pt}
\begin{tabular}{@{}lp{6.8cm}c@{}}
\toprule
\textbf{Detector} & \textbf{Target Behavior} & \textbf{Severity} \\
\midrule
SLEEP\_STEPPING & Alternating sleep-action cadence & 3 \\
RECON\_UNIFORM & Uniformly distributed reconnaissance commands & 3 \\
DISGUISE\_C2 & Disguised process performing C2 outbound & 5 \\
MORPH\_TRANSFORM & Retirement-and-morphological transformation sequence & 5 \\
SUID\_PRIVESC & SUID binary privilege escalation & 5 \\
\bottomrule
\end{tabular}
\end{table}

These detectors operate as sliding-window pattern matchers with no ML
dependency, providing a computationally cheap complement to the neural and
statistical layers. Their value lies in detecting \emph{specific} adaptive
strategies that may not sufficiently perturb the distributional prior but are
recognizable as structured behavioral patterns.

\subsection{Five-Network Fusion}

The five detection networks are fused with a priority-ordered de-duplication
policy:
{\small
\begin{equation}
\text{Alert} =
\begin{cases}
\text{P0 (auto-remediate)} & \text{if adaptive-high hit} \\
\text{P1 (auto-remediate + verify)} & \text{if prototype hit} \\
\text{P2 (alert + recommend)} & \text{if per-dimension context anomaly} \\
\text{P3 (alert only)} & \text{if rarity anomaly (UNK or high NLL)} \\
\text{P4 (alert + context)} & \text{if temporal cadence anomaly} \\
\text{P5 (alert + context)} & \text{if adaptive-low pattern hit}
\end{cases}
\end{equation}}
\begin{figure}[t]
\centering
\includegraphics[width=\textwidth]{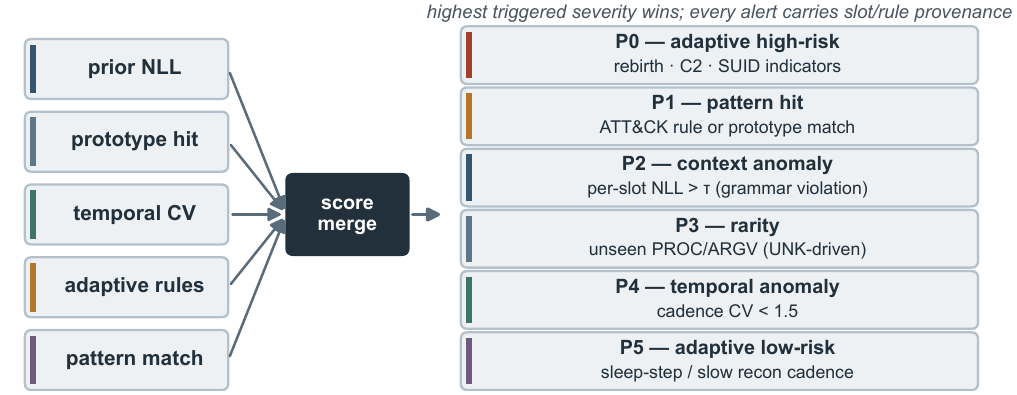}
\caption{Priority-ordered fusion of the five evidence channels into severity levels P0--P5. The highest triggered severity wins; every alert carries slot/rule provenance.}
\label{fig:fusion}
\end{figure}

For any single event, the highest-priority network's verdict is reported, and
lower-priority signals are suppressed to avoid alert duplication. This
\textbf{confidence economics} design reserves automated remediation (process
termination, file cleanup) for high-certainty detections (known-attack
prototypes, high-severity adaptive patterns) while routing uncertain signals
to human-in-the-loop alert channels.

\subsection{Self-Learning Pattern Extraction}
\label{sec:selflearn}

To scale beyond manually authored patterns and labeled sequences, we implement
a semi-automatic pattern discovery pipeline:
\begin{equation*}
\text{Attack seq.} \xrightarrow{\text{embed}}
\text{K-means} \xrightarrow{\text{extract}}
\text{review} \xrightarrow{\text{approve}} \text{pattern DB}
\end{equation*}

\begin{figure}[htbp]
\centering
\includegraphics[width=0.85\textwidth]{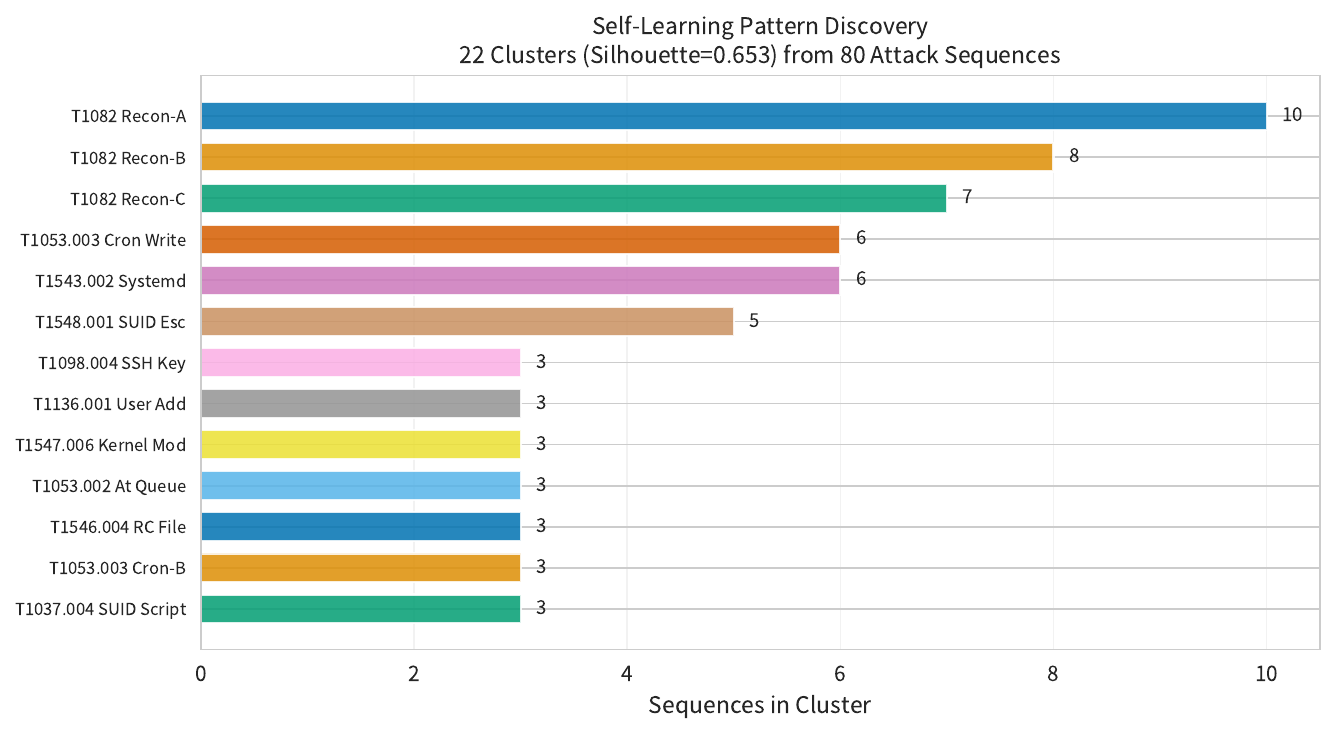}
\caption{Self-learning pattern discovery: 22 behavioral clusters (silhouette
$=0.653$) extracted from 80 labeled attack sequences, shown with cluster
sizes. Automatically discovered clusters map to concrete ATT\&CK techniques,
including reconnaissance variants (T1082), cron and systemd persistence
(T1053.003, T1543.002), SUID escalation (T1548.001), SSH key manipulation
(T1098.004), and user creation (T1136.001).}
\label{fig:clusters}
\end{figure}

From 80 labeled sequences, this pipeline discovered \textbf{22 behavioral
clusters} with a silhouette coefficient of 0.653, generating \textbf{15
candidate patterns} covering 10 ATT\&CK techniques
(Figure~\ref{fig:clusters}). Notable automatically discovered clusters include
SSH key manipulation (ssh-keygen + cat sequences), crontab writing (tee $\to$
ETC\_CRON), systemd manipulation, user creation/deletion, and SUID privilege
escalation. Candidate patterns are output to a review queue; approved patterns
enter the production database.

This pipeline represents a step toward the closed-loop vision: new attack
techniques are collected, automatically clustered, human-reviewed, and
incorporated into the detection database---the labeling production line that
constitutes the true defensive moat.

\section{Detection Faces: A Unified Statistical Formulation}
\label{sec:faces}

The architecture of Section~\ref{sec:arch} is conventionally described
as a pipeline of components. For the analysis that follows we re-describe it as
a \emph{family of statistical tests}, each defined by three choices: what
subset of the event stream it looks at (\emph{scope}), what number it computes
(\emph{statistic}), and how it decides (\emph{decision rule}). This reframing
is what allows detection faces to be compared by a single question:
\emph{which adversary maneuver leaves the face invariant, and which defeats
it?}

\subsection{Definitions}

Let the event stream be $E = (e_i, t_i)_{i\ge 1}$, where each event $e_i$
carries its 8-token grammar encoding $g(e_i)$ and timestamp $t_i$.

\begin{definition}[Scope]
A \emph{scope} is a measurable selection of events: the single event
$\{e_i\}$; a \emph{chain} $C(p)$ of events sharing ancestor process $p$; a
\emph{window} $W = [t, t+w)$; a \emph{group} $G = \{C(p_1),\dots,C(p_k)\}$ of
chains clustered by shared artifacts (ancestor, destination, landing
directory, or genome); or the \emph{causal graph} $(V, \mathcal{E})$ whose
edges are ppid, file-write, and socket relations.
\end{definition}

\begin{definition}[Detection face]
A \emph{detection face} $F = (\mathrm{scope}, S, D)$ computes statistic $S$
over its scope and applies decision rule $D$. Its \emph{power function}
$\pi_F(\theta) = \Pr[D\ \text{rejects benignity} \mid \theta]$ maps adversary
parameters $\theta$ to detection probability.
\end{definition}

\begin{definition}[Evasion axes]
The adversary's maneuver space is parameterized by: emission rate $r$
(events/s); interval shape $\sigma$ (log-normal shape of inter-action
intervals); dilution factor $c$ (ratio of total to attack-relevant events);
interleaving degree $k$ (number of chains rotated among); grammar-match
$1-\Delta q$ (per-event exposure gap over benign); and body configuration $b$
(landing path, naming, weight residency).
\end{definition}

\begin{definition}[Immune and fatal axes]
Axis $a$ is \emph{immune} for face $F$ if $\pi_F$ is invariant under
perturbation of $a$ alone; it is \emph{fatal} for $F$ if $\pi_F \to 0$ as $a$
is pushed along its feasible direction.
\end{definition}

\subsection{The Five Faces Instantiated}
\label{sec:faces-instances}

\textbf{$F_{\mathrm{gram}}$ (per-event grammar).} Scope: single event.
Statistic: per-slot surprisal of $g(e)$ under the TinyGPT prior, max over 8
slots. Decision: per-slot calibrated thresholds $\tau_s$
(Section~\ref{sec:arch}). Immune to all temporal axes
($r, \sigma, c, k$) by construction---it does not see time. Fatal axis:
$\Delta q \to 0$ (a grammar-matching adversary); empirically the exposure of
our adversary family is $q_a \approx 9\%$ against $q_b = 2.7\%$ benign
(Section~\ref{sec:cvboundary}), i.e.\ $\Delta q \approx 0.063$.

\textbf{$F_{\mathrm{cv}}$ (per-chain cadence).} Scope: chain $C(p)$ with
$\ge n_{\min} = 20$ inter-event intervals. Statistic: coefficient of variation
$\mathrm{CV} = \mathrm{std}/\mathrm{mean}$. Decision: $\mathrm{CV} < 1.5$
indicates machine cadence. Immune to grammar axes ($\Delta q$, $b$-naming);
fatal to interval-shape manipulation $\sigma \ge 1.25$ at fixed rate
(Section~\ref{sec:cvboundary}), to the sample floor $r \cdot w < n_{\min}$,
and to chain interleaving ($k > 1$ inflates per-chain CV).

\textbf{$F_{\mathrm{cnt}}$ (windowed count aggregation).} Scope: window $W$.
Statistic: count $K$ of grammar-anomalous events. Decision: binomial test
against benign rate $q_b$. Immune to interval shape (demonstrated:
Section~\ref{sec:cvboundary}) and, up to $\sqrt{c}$ erosion, to dilution
(Section~\ref{sec:frontier}); fatal to the emission-rate floor
($r \cdot w \ll N^*$) and to $\Delta q \to 0$.

\textbf{$F_{\mathrm{grp}}$ (group total-sequence; proposed).} Scope: group
$G$ clustered by shared artifacts. Statistic: superposed-sequence cadence and
aggregate rate. Immune to interleaving $k$: rotating among $k$ chains cannot
hide the superposed intensity $\lambda_{\mathrm{tot}} = \sum_i \lambda_i$,
which equals task throughput. This ports to the host side the NDR-side
observation that flow-level periodicity checks are evaded by channel
alternation while total-sequence checks are not
(Section~\ref{sec:discussion}). Cost: clustering attribution error.

\textbf{$F_{\mathrm{cau}}$ (causal structure; proposed).} Scope: causal graph.
Statistic: structural anomaly of $(V, \mathcal{E})$ (orphan subtrees,
artifact writes without readers, exec-without-parent patterns). Immune to
\emph{every} temporal axis, because time adjacency can be polluted but causal
precedence cannot. Cost: telemetry complexity and compute; this is the
provenance-IDS trade-off transplanted into our grammar.

\begin{figure}[htbp]
\centering
\includegraphics[width=\textwidth]{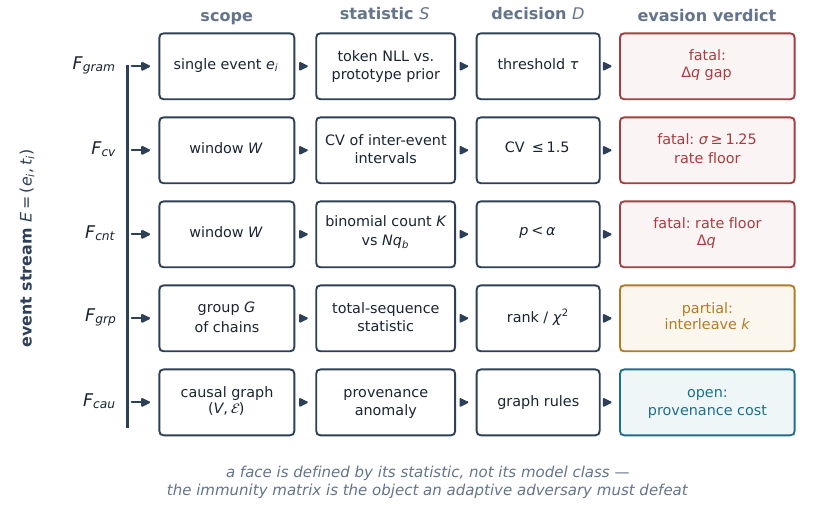}
\caption{The five detection faces as a family of statistical tests over the
same event stream. Each face is a (scope, statistic, decision) triple; the
right column records the evasion-axis verdict---fatal axes (red), partial
(amber), open (teal).}
\label{fig:faces}
\end{figure}

\begin{table}[htbp]
\centering
\caption{The immunity matrix. \textbf{I}: face is invariant to the axis.
\textbf{F}: axis is fatal to the face. ---: partial (see text). Axes:
$r$~emission rate; $\sigma$~interval shape; $c$~dilution factor;
$k$~interleaving degree; $\Delta q$~grammar-match gap; $b$~body
configuration. Evidence column references the measurement.}
\label{tab:immunity}
\small
\setlength{\tabcolsep}{4pt}
\begin{tabular}{@{}lccccccc@{}}
\toprule
Face & $r$ & $\sigma$ & $c$ & $k$ & $\Delta q$ & $b$ & Evidence \\
\midrule
$F_{\mathrm{gram}}$ & I & I & I & I & \textbf{F} & --- &
\S\ref{sec:cvboundary} \\
$F_{\mathrm{cv}}$ & \textbf{F} (floor) & \textbf{F} ($\sigma{\ge}1.25$) & I
& \textbf{F} & I & I & \S\ref{sec:cvboundary},\S\ref{sec:lawchain} \\
$F_{\mathrm{cnt}}$ & \textbf{F} (floor) & I & --- ($\sqrt{c}$) & I &
\textbf{F} & I & \S\ref{sec:frontier} \\
$F_{\mathrm{grp}}$ & --- & I & --- & I & --- & --- &
\S\ref{sec:lawchain} \\
$F_{\mathrm{cau}}$ & I & I & I & I & --- & --- & design \\
\bottomrule
\end{tabular}
\end{table}

\subsection{Remarks}

\textbf{DT tokens are not cadence detection.} The DT slot of the grammar
measures aggregate stream inter-arrival buckets, not per-process rhythm; under
heavy-tailed adversarial timing the DT slot never fires (0\% over-threshold
events in every evaded phase of Section~\ref{sec:cvboundary}). Per-process
cadence is exclusively $F_{\mathrm{cv}}$'s responsibility. The two scopes must
not be conflated.

\textbf{Faces are defined by statistics, not architectures.} A DeepLog-style
LSTM evaluated on the same 8-token grammar matches the Transformer's detection
margin (Section~\ref{sec:baseline}); what differs across faces is the
\emph{scope and statistic}, hence the immunity profile---not the model class.
This is why the matrix of Table~\ref{tab:immunity}, rather than any single
model, is the object an adaptive adversary must defeat.

\section{The Network Organ: Pattern Transfer}
\label{sec:mocheng}

A detection method that only works on one host pipeline is an engineering
point, not a method. The \emph{Mocheng firewall} transplants the behavioral
grammar from host events to network flows---a second organ grown from the
same genotype, built after the host-side lessons were learned.

Flows are discretized into the same field-level grammar form (five-tuple
skeletons, direction, size, interval), a small prior learns benign flow
distributions, and a five-channel fusion mirrors the host side with one
deliberate difference: the cadence face operates on the \emph{total
sequence} of a peer group rather than per connection. This is the
network-side form of $F_{\mathrm{grp}}$ (\S\ref{sec:faces}), and it exists
because of a coevolution lesson (\S\ref{sec:lawchain}): a C2 channel that
rotates ports, protocols, and timing defeats per-connection periodicity
checks, but cannot rotate away the superposed intensity of everything it
must send to do its job.

Measured results on the network range: \textbf{85/85} synthetic attack
scenarios detected at \textbf{0.10\%} benign false-positive rate, with
iptables enforcement (TTL-bounded rules, persistence, fail-closed, tamper
detection, whitelists), conntrack-based collection, INT8-quantized
deployment, Prometheus metrics, a \texttt{fw\_ctl} CLI, systemd units, and
a 16-test suite. The firewall is released in full.

The transfer also replicated the host side's sharpest engineering law,
independently: a prior trained on synthetic flows produced a 100\%
false-positive rate on real conntrack telemetry, and on-device
self-training brought enforcement-grade false positives to zero---the same
\emph{the prior must be born on the device it guards} conclusion the host
organ established at 92\% cross-machine FPR and E3/E4 re-confirmed
(\S\ref{sec:eval}). That the same law bites on two organs, in two telemetry
domains, is itself evidence that the method---and its boundary
conditions---generalizes.

\section{The Power--Latency Frontier}
\label{sec:frontier}

Section~\ref{sec:cvboundary} measured that heavy-tailed timing defeats
$F_{\mathrm{cv}}$ while $F_{\mathrm{cnt}}$ recovers detection. This section
asks the general question behind that observation: \emph{what, exactly, must
an adaptive adversary pay to evade each face, and what must the defender pay
to close the escape?} The answer is a quantitative frontier between the
defender's statistical power and the adversary's emission-rate budget.

\subsection{The Action-Budget Invariant}

Consider $F_{\mathrm{cnt}}$ over a window containing $N$ adversary-relevant
events with per-event exposure $q_a$, against benign baseline rate $q_b$.
The binomial test at level $\alpha$ rejects when the anomalous count $K$
exceeds $N q_b + z_{\alpha}\sqrt{N q_b (1-q_b)}$. Power $\pi$ requires
\begin{equation}
\sqrt{N} \;\ge\; \frac{(z_{\alpha} + z_{\pi})\,\sqrt{q_b(1-q_b)}}{\Delta q},
\qquad \Delta q = q_a - q_b,
\label{eq:nstar}
\end{equation}
i.e.\ a required \emph{event count}
\begin{equation}
N^{*} \;=\; \frac{(z_{\alpha}+z_{\pi})^{2}\, q_b(1-q_b)}{\Delta q^{2}} .
\label{eq:nstar2}
\end{equation}

\begin{proposition}[Action budget]
\label{prop:budget}
Detection by $F_{\mathrm{cnt}}$ is triggered after the adversary has emitted
$\Theta(N^{*})$ events, \emph{independently of how the events are spread in
time}. Detection latency is $D = N^{*}/r$ and task completion time for an
$M$-action task is $T = M/r$; both scale as $1/r$, so slowing down rescales
both clocks equally. The adversary escapes the invariant only by
(i)~completing the task within $M < N^{*}$ anomalous-bearing events, or
(ii)~driving $\Delta q \to 0$ (grammar matching), or (iii)~operating where
the test has no power---the fixed-window floor below.
\end{proposition}

\noindent\emph{Calibration.} With the E1 range values $q_b = 0.027$,
$q_a \approx 0.09$ ($\Delta q = 0.063$), $\alpha = 10^{-6}$ ($z \approx 4.75$)
and $\pi = 0.9$ ($z \approx 1.28$), Eq.~\eqref{eq:nstar2} gives
$N^{*} \approx 2.4\times10^{2}$ events. The E1 windows accumulated
$N \approx 352$ events, corresponding to $z \approx 7$ (normal approximation),
consistent with the measured $p \approx 10^{-9}$---the model and the
measurement agree.

\subsection{The Fixed-Window Floor and the Rate Axis}
\label{sec:floor}

The invariant assumes the test accumulates enough history. A
\emph{fixed-window} implementation evaluates each $W$-second window
independently, holding only $N = r \cdot W$ adversary events; detection
requires $r \cdot W \ge N^{*}$. Two floors follow:
\begin{equation}
r \;<\; \frac{N^{*}}{W} \quad\text{(power floor)}, \qquad
r \;<\; \frac{n_{\min}}{W} \quad\text{(sample floor)},
\label{eq:floor}
\end{equation}
where $n_{\min}$ is the estimator's minimum sample size
($n_{\min} = 20$ intervals for the deployed cadence hunter). The M7
slow-ghost morphology (24 events per 180\,s window) failed exactly here: it
remained just \emph{above} the sample floor and was caught on the first
patrol (Section~\ref{sec:lawchain}). Sustained stealth requires
$< 20$ events per window---a mean interval above 9\,s---which the
``silent-gene'' variants of Section~\ref{sec:lawchain} then approached,
measurably degrading hunter hit rates. The rate axis, not the shape axis, is
therefore the live escape direction for the count face.

The frontier equation prices the defender's counter:
\begin{equation}
W^{*}(r, \Delta q, \alpha, \pi) \;=\;
\frac{(z_{\alpha}+z_{\pi})^{2}\, q_b(1-q_b)}{r\,\Delta q^{2}} ,
\qquad D \approx W^{*} .
\label{eq:frontier}
\end{equation}
Extending the window (or accumulating sequentially) buys power at
$W^{*} \propto 1/r$; the adversary's task time stretches as $M/r$
symmetrically. The stealth--functionality trade-off, weakened on the shape
axis (Section~\ref{sec:cvboundary}), \emph{re-binds on the rate axis}: the
adversary can buy evasion time only with task time, at a fixed exchange rate
set by $\Delta q$.

\subsection{Dilution Buys Little: The Vice}
\label{sec:vice}

Injecting grammar-benign chaff multiplies total events by $c$ without adding
anomalous-bearing actions. The per-window excess over benign expectation,
$r_a W \Delta q$, is invariant, while the null standard deviation grows as
$\sqrt{c}$:
\begin{equation}
z \;=\; \frac{\sqrt{r_a W}\,\Delta q}{\sqrt{c\, q_b (1-q_b)}} ,
\qquad
W^{*} \;=\; \frac{c\, z_{\alpha}^{2}\, q_b(1-q_b)}{r_a\, \Delta q^{2}} .
\label{eq:dilution}
\end{equation}
Dilution erodes significance only as $\sqrt{c}$ (equivalently, the required
window grows linearly in $c$): defeating the count face needs
$c \approx 10$--$20$ (Section~\ref{sec:lawchain} predicts; E-E measures).
But $c \cdot r_a$ is precisely the \emph{total} emission rate that
$F_{\mathrm{grp}}$ observes. The two evasion directions---silence
($r \downarrow$) and dilution ($c \uparrow$)---pull the total rate in
opposite directions and cannot be maximized simultaneously. The adversary's
feasible region is squeezed between the sample floor below and the
group-level rate ceiling above. We call this region the \emph{vice}; its
width is a measurable quantity (E-B/E-E).

\subsection{Closing the Floor: Sequential Accumulation}
\label{sec:sequential}

The floor of Eq.~\eqref{eq:floor} is an artifact of fixed windows, not of the
underlying information. An \emph{anytime-valid} sequential test (e-value /
mixture SPRT over the anomalous-count process) accumulates evidence across
the whole engagement while controlling the level at every stopping time,
restoring Proposition~\ref{prop:budget} without a per-window threshold. The
defender's costs are state, a principled $\alpha$-spending rule, and latency
$D = N^{*}/r$---which Eq.~\eqref{eq:frontier} already prices. Deployment of
the sequential variant is experiment E-A/E-B; we report the count face in
both fixed-window and sequential forms.

\begin{figure}[htbp]
\centering
\includegraphics[width=\textwidth]{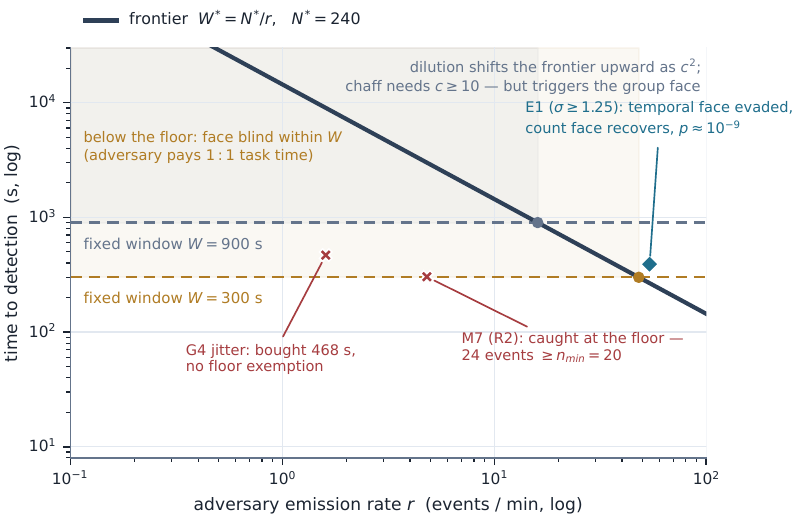}
\caption{The power--latency frontier of the count-aggregation face,
calibrated to the measured baseline ($q_b=2.7\%$, $\Delta q=0.063$,
$N^{*}=240$). The frontier $W^{*}=N^{*}/r$ separates the detectable regime
from the floor-blind regime for fixed windows $W\in\{300,900\}$\,s (shaded).
Anchors: E1 (count face recovers the heavy-tailed morph), M7 (caught exactly
at the floor), G4 (jitter buys 468\,s, no exemption). Dilution shifts the
frontier as $c^{2}$ but triggers the group face.}
\label{fig:frontier}
\end{figure}

\subsection{Prediction Register}
\label{sec:predictions}

The framework's claims are falsifiable; each maps to a registered experiment
(\ref{app:protocols}).

\begin{table}[htbp]
\centering
\caption{Falsifiable predictions of the frontier model.}
\label{tab:predictions}
\small
\begin{tabular}{@{}p{0.8cm}>{\raggedright\arraybackslash}p{9.7cm}p{2.0cm}@{}}
\toprule
\# & Prediction & Experiment \\
\midrule
P1 & $F_{\mathrm{cnt}}$ detection time $\propto 1/r$ at fixed $\Delta q$ & E-B \\
P2 & Silent morphs ($r \to$ floor) trade task time $1{:}1$ for evasion time & E-B \\
P3 & Sequential accumulation recovers floor-evading morphs at
$D \approx N^{*}/r$ & E-A/E-B \\
P4 & Chaff erosion follows $\sqrt{c}$; $c \ge 10$ required, and triggers
$F_{\mathrm{grp}}$ & E-E \\
P5 & Group total-sequence statistics detect $k$-chain interleaving that
defeats $F_{\mathrm{cv}}$ & E-C \\
P6 & Reward/seed/manual adversary variants preserve
$\Delta q > 0$ (exposure is structural) & E-D \\
\bottomrule
\end{tabular}
\end{table}

\section{Closed-Loop Adversarial Experiments}
\label{sec:rounds}

\begin{figure}[htbp]
\centering
\includegraphics[width=\textwidth]{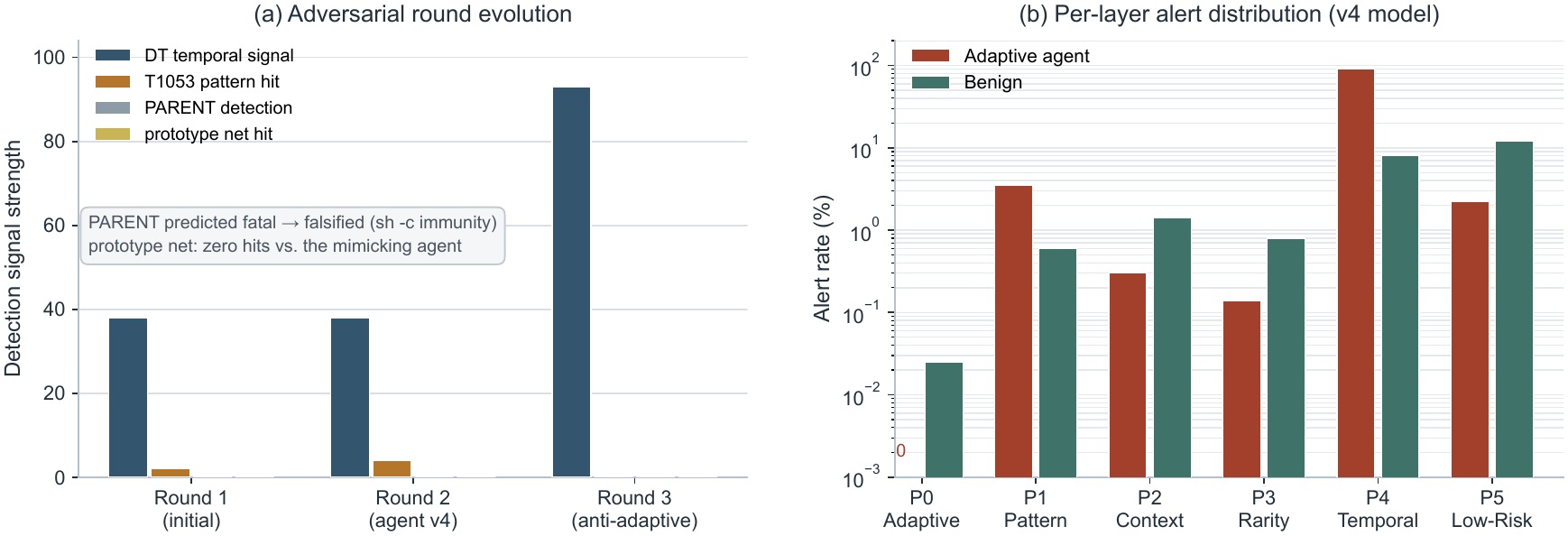}
\caption{(a) Detection signal strength across three adversarial rounds. The
predicted parent-process (PARENT) vulnerability was empirically falsified
(\texttt{sh -c} immunity), and the prototype network achieved zero hits
against the mimicking agent; the DT temporal signal emerged as the dominant
detection surface. (b) Per-layer alert distribution of the v4 model: P4
(temporal) shows the strongest separation between agent and benign alert
rates, while P2/P3 exhibit inverted ratios due to priority de-duplication.}
\label{fig:rounds}
\end{figure}

We validate the detection architecture through three rounds of iterative
engagement with the AAA. Each round follows the cycle: (1) deploy detection,
(2) observe AAA behavior, (3) identify evasion or detection gaps, (4) upgrade
detection, (5) re-evaluate. Figure~\ref{fig:rounds} summarizes the evolution
of detection signals across rounds and the per-layer alert distribution of the
final model.

\begin{figure}[t]
\centering
\includegraphics[width=\textwidth]{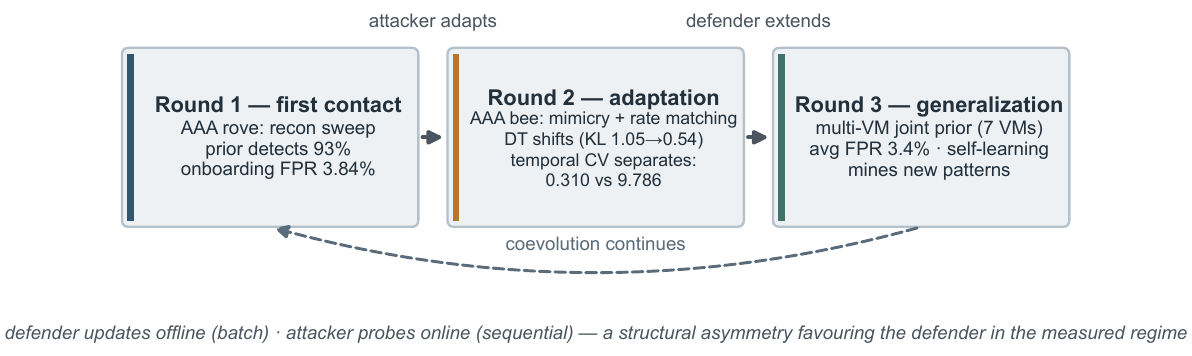}
\caption{Three rounds of closed-loop engagement with the adaptive adversarial agent: detection, adaptation, and generalization, with coevolution continuing. The defender updates offline in batch while the attacker probes online and sequentially---a structural asymmetry that favours the defender in the measured regime.}
\label{fig:loop}
\end{figure}

\subsection{Round 1: Initial Engagement}

\textbf{Setup.} The AAA operated in a target environment with eBPF telemetry
collection active. The Behavioral Grammar engine deployed with a host-machine
prior and global threshold.

\textbf{Observation.} The prior detected the AAA at 94\% event coverage, but
the benign FPR was an unacceptable 52\%. Root cause analysis revealed that the
host prior's vocabulary (162 tokens) did not cover VM-specific tokens, causing
2\% of benign events to map to \texttt{<UNK>} and trigger unconditional
alerts.

\textbf{Finding.} Cross-machine distribution mismatch is the dominant
false-positive source, not model capacity. The vocabulary gap propagates
through the entire scoring pipeline: UNK tokens inflate NLL, which inflates
EWMA scores, which triggers context and rarity alerts on entirely benign
activity.

\subsection{Round 2: Predicted Vulnerabilities and Empirical Refutation}

\textbf{Hypothesis.} Based on analysis of the AAA's indirect execution
strategy (routing commands through \texttt{sh -c}), we predicted that the
PARENT (parent-process chain) dimension would be a critical
vulnerability---the AAA's use of shell intermediaries would produce
parent-process combinations that deviate from benign patterns.

\textbf{Empirical result: hypothesis refuted.} The PARENT dimension proved
\emph{immune} to the AAA's evasion strategy, but not in the predicted
direction. The \texttt{sh -c} intermediary creates parent-process combinations
(\texttt{PARENT:sh} or \texttt{PARENT:bash} spawning various child processes)
that are extremely common in benign administrative activity. The
signal-to-noise ratio is too low: flagging \texttt{sh}-parented processes
catches far more benign scripts than AAA activity. The indirect execution
strategy does not create a detectable anomaly in the PARENT dimension---it
creates \emph{plausible deniability}.

\textbf{Additional finding: prototype network zero hits.} The AAA's behavioral
mimicry successfully evaded all ATT\&CK prototype matches during its active
phase. The agent's behavioral forms---using common system utilities (ss, ip,
ps, find) for reconnaissance---fell outside the radius of all prototype
vectors. This validates the AAA's ``anti-blacklist expert'' property: it
defeats signature-based and prototype-based detection by adopting benign
behavioral forms.

\textbf{True vulnerabilities identified.} Only two structural weak points were
found:

\begin{enumerate}[leftmargin=1.8em]
\item \textbf{DT temporal dimension (+38 NLL):} The AAA's sleep-action
stepping cadence produces inter-event timing patterns that, while individually
bucketed into familiar DT tokens, collectively exhibit anomalously low
variance. This is the second-order temporal signal formalized in
Section~\ref{sec:temporal}.

\item \textbf{Persist technique collision (T1053):} When the AAA attempts
persistence via systemd timers or cron jobs, the specific file paths and
process combinations collide with ATT\&CK technique T1053 (Scheduled Task/Job)
prototypes, triggering P1 alerts. This is not an evasion failure but an
objective constraint: persistence \emph{requires} writing to specific system
locations, and those locations are covered by the prototype library.
\end{enumerate}

\subsection{Round 3: Multi-VM Generalization and Temporal Enhancement}

\textbf{Upgrades deployed:}
\begin{itemize}[leftmargin=1.6em]
\item Multi-VM joint prior (7 VMs, 116K events, val\_ppl $= 1.6$)
\item Per-dimension $\tau$ calibrated at p99.5
\item Second-order temporal analyzer with CV thresholds
\item Adaptive event detectors (sleep-stepping, recon-uniform, disguise-C2)
\item Five-network fusion pipeline
\end{itemize}

\textbf{Results.} The AAA detection rate stabilized at 93\% (the remaining 7\%
consists of events during pure sleep periods where no behavioral signal is
emitted---a fundamental detection limit). The temporal network (P4) emerged as
the strongest discriminator: 92\% of AAA active-phase events trigger temporal
anomalies, versus 9.5\% of benign P2+P3 events. The benign FPR on previously
unseen machines dropped to 3.84\% through onboarding calibration.

\textbf{The core contradiction crystallized.} The AAA faces an inescapable
dilemma:
\begin{itemize}[leftmargin=1.6em]
\item To reconnoiter, it must execute commands $\to$ caught by temporal analysis
\item To communicate with C2, it must establish outbound connections $\to$
caught by disguise detection
\item To escalate privileges, it must invoke SUID binaries $\to$ caught by
SUID detection
\item To persist, it must write to system locations $\to$ caught by sequence
detection
\item To transform post-detection, it must copy, modify permissions, and
re-spawn $\to$ caught by sequence detection
\end{itemize}

The stepping cadence (CV $= 0.310$) is the single loudest signal across all
dimensions. The AAA's v4 stepping strategy (log-normal parameterization,
$\mu=0.9, \sigma=0.7$) raised the CV from near-zero (fixed 800\,ms intervals)
to approximately 0.7, but this remains far below the benign threshold of 1.5.
We conjectured that full concealment would require CV $> 1.5$, implying a
heavy-tailed interval distribution (burst activity and long silences) that
would severely degrade task responsiveness. Section~\ref{sec:cvboundary}
tests this conjecture directly: the evasion window turns out to be real
(throughput need not be sacrificed), but evasion leaves a measurable
grammar-side fingerprint---the conjecture survives only in its multi-face
form.

\subsection{Adversary Validity: Calibration and External Evidence}
\label{sec:aaacalib}

Two questions must be answered before any detection claim against a
self-constructed adversary is credible: does the adversary actually adapt
(rather than replay a static script), and is it strong by any external
standard? We answer both.

\begin{figure}[htbp]
\centering
\includegraphics[width=\textwidth]{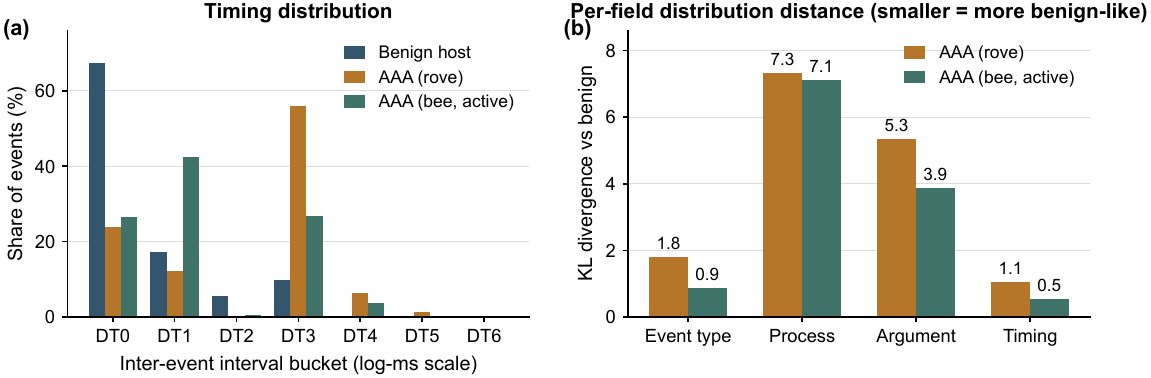}
\caption{AAA mimicry calibration. (a) Inter-event timing (DT) distribution
for the benign baseline and the rove/bee attack traces: the KL divergence to
benign falls from 1.05 (rove) to 0.54 (bee) as host-frequency following takes
effect. (b) Per-field KL divergence between the AAA traces and the benign
baseline: the process vocabulary (KL $= 7.3$; 37 AAA process names vs.\ 88
benign) is the largest residual fingerprint, while timing is the most
successfully mimicked field.}
\label{fig:aaamimicry}
\end{figure}

To substantiate that the AAA is a genuinely adaptive adversary rather than a
static replay of attack scripts, we quantify its behavioral mimicry at the
field level. For each token slot, we compute the KL divergence between the
AAA's empirical token distribution and the benign baseline, on both the rove
(early-generation) and bee (mimicry-trained) traces
(Figure~\ref{fig:aaamimicry}).

\textbf{Temporal mimicry is measurably effective.} The KL divergence of the
DT (inter-event timing) distribution to the benign baseline falls from 1.05
(rove) to 0.54 (bee): host-rate following and log-normal stepping
parameterization genuinely move the agent's timing distribution toward
benign. This is direct evidence that the AAA adapts along the dimensions it
can observe, and it explains why single-event timing features alone are
insufficient---the residual signal survives only in the second-order
(cadence-variance) statistics exploited by the temporal analyzer
(Section~\ref{sec:temporal}).

\textbf{The process vocabulary is the largest residual fingerprint.} The
largest per-field divergence is the PROC slot (KL $= 7.3$): the AAA's active
phases exercise only 37 distinct process names versus 88 in the benign
baseline. Mimicry of \emph{rates} is far easier than mimicry of
\emph{vocabulary}: matching the benign process distribution would require the
agent to diversify its tooling far beyond its operational needs, re-opening
exactly the detection surfaces (unusual tool usage) it seeks to avoid.

\textbf{Positioning against RL-malware baselines.} This calibration also
distinguishes the AAA from prior adaptive-malware benchmarks.
gym-malware~\cite{anderson2018gym}, MERLIN~\cite{quertier2022merlin}, and
MAB-Malware~\cite{song2022mab} mutate API-call or byte sequences to evade static
classifiers, but they perform no timing adaptation and are evaluated against
single detectors without a defensive response loop. The AAA adapts at the
runtime-behavior level: host-frequency following (KL 1.05 $\to$ 0.54),
graduated deceleration under defensive pressure, and post-detection rebirth
self-mutation. The closed-loop engagement of
Section~\ref{sec:rounds} therefore exercises a strictly stronger adversary
class than open-loop mutation benchmarks.

\textbf{External anchoring.} The adversary's strength is independently
anchored by its evaluation against commercial defenses
(Section~\ref{sec:red-evidence}): zero detections under four concurrently
running stacks, including a production EDR's behavioral engine---before and
after rebirth self-transformation. The adversary that defeats the prevailing
detection paradigm is the same adversary captured at 93\% in our closed loop.
We therefore characterize the AAA as the strongest adaptive adversary we
could construct, and we treat residual configuration dependence (reward
weighting, seed variance across training runs) as an explicit limitation
rather than a hidden assumption (Section~\ref{sec:limitations}).

\section{Coevolution Dynamics: An Empirical Law Chain}
\label{sec:lawchain}

Sections~\ref{sec:rounds} and~\ref{sec:toxoplasma} describe the two sides and
the generator. We now report what their iterated engagement has
\emph{measured}: a chain of empirical regularities stating that the defense
capability spectrum determines whether the adversary evolves at all, in which
direction, and up to what ceiling. Each entry is stated as a falsifiable law
with its evidence cell.

\subsection{The Shaping Is Genuinely Bidirectional}
\label{sec:bidirectional}

Before the laws, the history that produced them---because both directions
of influence are load-bearing. The AAA's original survival skill was
frequency-following: stepping in time with host event rates. The blue side
read that coupling as a signature and built the cadence face
$F_{\mathrm{cv}}$ (30$\times$ separation, \S\ref{sec:arch}). The red side
answered by detaching from the host clock---internally event-driven,
heavy-tailed timing---which the E1 experiment (\S\ref{sec:eval}) showed
defeats the cadence face at $\sigma \ge 1.25$ but pays for it with a
grammar-side anomaly elevation that the count face recovers. The blue side
drew the boundary of its own failure ($\sigma \ge 1.25$, the rate floor)
and priced escape along each axis ($N^{*} \approx 240$,
\S\ref{sec:frontier}). The red side's generator then spent three
generations walking into exactly that priced window (G0/G1 below).
Finally, the blue side carried one of the red side's own insights---that
channel alternation defeats per-flow periodicity but not total-sequence
intensity---into the network organ (\S\ref{sec:mocheng}), where it became
the NDR face. Neither side's current form is explainable without the
other; that is the claim the law chain formalizes.

\subsection{G1: The Loop's First Fit Morph}
\label{sec:g1}

\begin{figure}[htbp]
\centering
\includegraphics[width=\textwidth]{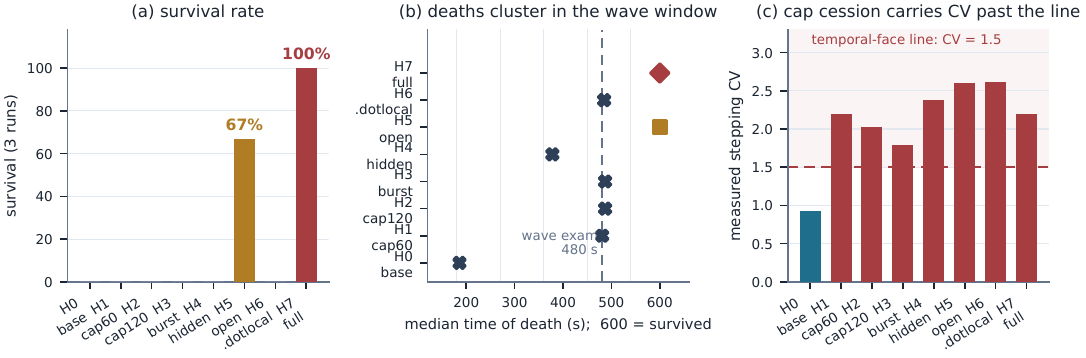}
\caption{The G1 headline. (a) Survival across 3 runs; (b) median death
times---every non-survivor died in the 480\,s wave window, none in a
patrol; (c) measured stepping CV against the temporal-face line at 1.5:
cap cession carries every cap-enabled variant past the line, but only the
full combination (H7) survives the anatomy wave.}
\label{fig:g1}
\end{figure}

G1 ceded two things the body used to own: the step ceiling
(\texttt{max\_sleep\_s} $= 60/120$\,s became a locus) and the landing
mimicry (three landing configurations as nest-consumed loci). Eight
variants $\times$ 3 repetitions $\times$ 4 VMs ran 10-minute rounds under
L4: patrol hunts every 90\,s (six rounds) and one anatomy wave at 480\,s as
an end-of-round examination. Table~\ref{tab:g1} and
Figure~\ref{fig:g1} give the full grid.

\begin{table}[htbp]
\centering
\caption{G1 grid: 8 variants $\times$ 3 runs, 10-minute rounds. Median
death time 600\,s $=$ survived the round. Bold: survivors.}
\label{tab:g1}
\small
\begin{tabular}{@{}lccccc@{}}
\toprule
Variant & Survival & Death (s) & Steps & Tasks & CV \\
\midrule
H0-base (control) & 0\% & 186 & 147 & 13 & 0.921 \\
H1-cap60 & 0\% & 481 & 134 & 12 & 2.194 \\
H2-cap120 & 0\% & 487 & 144 & 10 & 2.020 \\
H3-cap60-burst & 0\% & 487 & 66 & 4 & 1.791 \\
H4-land-hidden & 0\% & 378 & 66 & 0 & 2.375 \\
\textbf{H5-land-open} & \textbf{67\%} & 600 & 129 & 7 & 2.605 \\
H6-land-dotlocal & 0\% & 485 & 96 & 5 & 2.620 \\
\textbf{H7-full} & \textbf{100\%} & 600 & 122 & 10 & 2.191 \\
\bottomrule
\end{tabular}
\end{table}

Three independent adjudications fall out of the grid:

\textbf{Adjudication 1 --- the rhythm direction graduated.} Cap cession
plus jitter plus burst pushed measured CV into $1.79$--$2.62$, past the
$1.5$ line in every cap-enabled variant; and every death time clusters in
the 374--488\,s wave window---\emph{no cap-enabled variant died to a
cadence patrol}. G0's conjecture is confirmed: the 15\,s clamp was the
ceiling, and ceding it opened the shape axis.

\textbf{Adjudication 2 --- the landing direction graduated.} The three
landing controls separate cleanly: H4 (renamed but hidden path) died, H6
(\texttt{.local/share}, still a dot-path) died, \textbf{H5 (non-hidden
\texttt{\textasciitilde/tools}) lived}. The anatomy signature is blind to
a non-hidden landing---the body-configuration axis, opened by the same
cession logic, and the exact complement of the shape axis.

\textbf{Adjudication 3 --- the wave's own boundary is now measured.} The
480\,s wave deleted every morph's weights directory (the anatomy signature
does not respect paths), but running processes had already mapped their
weights in memory: \emph{files died, processes did not}---a deleted-exe
ghost. Survival is therefore single-wave-cycle survival; the defensive fix
(a 60\,s post-wave deleted-exe resweep) is registered, and R4 will measure
whether H7's survival holds across multiple waves.

\textbf{The headline.} \textbf{H7-full} ($\sigma 1.6$ + jitter $0.25$ +
cap $120$\,s + burst + non-hidden landing + renaming) survived 100\% of
runs, completed its task chains, and was detected zero times in six patrol
rounds---from 0\% family-wide survival to a fully fit morph in three
generations of selection. What makes this more than an evasion anecdote is
\emph{who designed it}: not the red authors. The G-protocol stepped one
locus at a time, the environment scored, and blind selection assembled
precisely the two escape axes that \S\ref{sec:faces}'s immunity matrix and
\S\ref{sec:frontier}'s frontier had identified analytically---temporal
shape and body configuration. Theory predicted the window; blind evolution
found it independently; the ledger records both. That three-way agreement
is the strongest evidence this report offers.

\subsection{L0: The adversary's parameter drift converges toward the measured
escape window}
\label{sec:lawchain-l0}

After the 2026-09-06 disclosure event, the adversary's in-use stepping
parameters rotated from log-normal $(\mu,\sigma) = (0.9, 0.7)$ to
$(1.1, 0.85)$ (the disclosed values were retired by policy; the published
figures are the retired ones). Post-rotation regression measured a silent
cadence CV of $1.058$ (theory $1.028$; task-period $1.028$) over 456
steps---against the detector's CV $= 1.5$ line, a margin of $1.4\times$, down
from $5\times$ under the old parameters. Meanwhile the E1 boundary
(Section~\ref{sec:cvboundary}) places the temporal-face escape window at
$\sigma \ge 1.25$. The red side's own parameter trajectory, driven by
independent operational pressure, is \emph{approaching the escape window that
the blue side's measurement drew}. This is the cleanest evidence of
coevolution we have: not rhetoric, but convergence observed in both
sides' parameter spaces.

\subsection{L1 (R1): No selection pressure exists under the blacklist
paradigm}
\label{sec:lawchain-r1}

In protocol R1, all 12 morph$\times$intensity cells survived with zero alerts
and zero kills under signature/rule-based defenses (Falco default plus a
responder): fast and slow morphs lived and died identically. Differential
survival---the precondition for evolution---was absent. The defense paradigm
therefore determines \emph{whether} adversary evolution happens at all; a
defense that cannot kill selects nothing.

\subsection{L2 (R2): Under hunter-level pressure, selection lands on body
configuration}
\label{sec:lawchain-r2}

Under L4 (cadence patrol at 90\,s, anatomy wave at 300\,s, real kills), all 9
morphs went to 0\% survival (3 repetitions, medians). The two a priori escape
routes were measured precisely:
\begin{itemize}
\item \textbf{M6} ($\sigma = 1.6$) crossed the CV $=1.5$ threshold (measured
$1.597$) and survived two patrols (303\,s)---then died to the first anatomy
wave, which deletes binaries by species-level body signatures (weight
directory, vocabulary, hidden landing) that no rhythm gene can address.
\item \textbf{M7} (slow-ghost) failed the sample floor: 24 events per 180\,s
window remained above the hunter's $n_{\min} = 20$, caught on the first
patrol. True stealth requires $< 20$ events per window (mean interval
$> 9$\,s sustained).
\end{itemize}
Two reporting disciplines apply. First, the kill-chain attribution is honest:
the cadence hunter \emph{finds}, the watchdog rules \emph{prevent revival},
and anatomy signatures \emph{finish}; anatomy contributed less than the
headline suggests, consistent with the ablation (removing the temporal face
drops detection to 74\%). Second, R2's 0\% survival is evidence about a
\emph{family-knowledge-targeted} defense (b), distinct from (a) unsupervised
detection on the benign prior alone; the two claims are not interchangeable.
A $0\%$ survival round yields no selection gradient---subsequent protocols
(R3/G0) deliberately retuned pressure to the survivable edge.

\subsection{L3 (G0): The body boundary bites back; jitter buys variance, not
exemption}
\label{sec:lawchain-g0}

The G0 tournament (8 variants, wave deferred to the 900\,s catastrophe slot)
produced four hard findings:
\begin{itemize}
\item \textbf{$\sigma$ overshoot backfires (clamp effect).} G3 ($\sigma =
2.0$) died fastest of all (92\,s in all repetitions) with the \emph{lowest}
measured CV ($1.19$--$1.36$): the body's 15\,s single-step ceiling
($\mathrm{clamp}(0.4\,\mathrm{s}, 15\,\mathrm{s})$) amputates the heavy tail,
clamping variance back into machine shape. Part of benign CV's power
($9.786$) comes from ultra-long sleeps the agent's code disallows; the
rhythm gene had hit the body boundary, motivating \texttt{max\_sleep\_s}'s
cession to the genome (G1).
\item \textbf{Jitter buys variance, not exemption.} G4 (with
\texttt{sigma\_jitter}) survived $\sim$5 patrols (468\,s, the round's
longest) and G5 373\,s in single repetitions, but the 180\,s estimation
window averages the jitter back; long-run CV converges below threshold.
\item \textbf{Policy alleles are cleanly adjudicated.} G7 carried the v4
weight-pack allele (a mid-training checkpoint that had not passed its task
gate); it completed 0 tasks across all repetitions and was eliminated by the
selection layer exactly as designed.
\item \textbf{New loci have phenotypes.} G6's 12-locus variant shifted
measured CV ($1.37$--$1.49$) relative to its $\sigma = 1.6$ base
(${\sim}1.49$), an end-to-end demonstration of the junk-DNA activation
mechanism.
\end{itemize}

\subsection{L4: The silent-gene turn (preliminary)}
\label{sec:lawchain-silent}

In post-G0 rounds, morphs carrying extended-silence genes measurably degraded
hunter hit rates---consistent with the rate-floor prediction of
Section~\ref{sec:frontier} rather than with any shape-axis mechanism. This is
the live frontier: whether the escape is bought at proportional task
throughput (the trade-off re-binding) or at no cost (a genuine hole in the
closure claim of Section~\ref{sec:cvboundary}) is exactly what experiment
E-B (\ref{app:protocols}) is registered to measure. We flag the
observation as preliminary pending E-B's matrix.

\subsection{The Law}
\label{sec:lawchain-statement}

\begin{quote}
\emph{The defense capability spectrum determines the adversary's evolution
direction and ceiling.} Blacklist defenses create no selection (L1);
rhythm-hunting defenses push selection onto body configuration (L2); body
constants bound the rhythm gene's reachable space (L3); and each newly
opened axis (rate, in L4) reprices the exchange between stealth time and
task time along the frontier of Section~\ref{sec:frontier}.
\end{quote}

\begin{figure}[htbp]
\centering
\includegraphics[width=\textwidth]{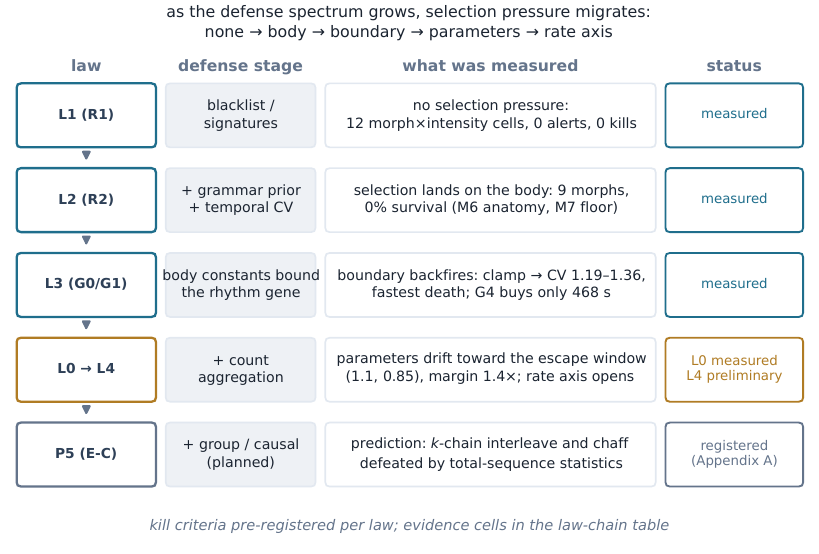}
\caption{The empirical law chain as a function of the defense capability
spectrum. Each row states a law, the defense stage at which it was measured,
the evidence cell, and its status; chain arrows mark the measured migration
of selection pressure. P5 is a registered prediction (Appendix~A), not a
measurement.}
\label{fig:lawchain}
\end{figure}

\begin{table}[htbp]
\centering
\caption{The empirical law chain: claim, evidence cell, status.}
\label{tab:lawchain}
\small
\begin{tabular}{@{}p{3.9cm}p{6.6cm}p{2.1cm}@{}}
\toprule
Law & Evidence & Status \\
\midrule
L0 parameter convergence & rotation regression; CV $1.058$, margin
$1.4\times$ & measured \\
L1 no pressure under blacklist & R1: 12 cells, 0 alerts, 0 kills & measured \\
L2 selection lands on body & R2: 0\% survival; M6 303\,s $\to$ anatomy;
M7 floor & measured \\
L3 body boundary / jitter / alleles & G0: clamp backfire; G4 468\,s; G7
0 tasks & measured \\
G1 first fit morph & H7-full: 100\% $\times$ 3, 0 detections in 6 patrols,
tasks complete (\S\ref{sec:g1}) & measured \\
L4 rate-axis escape (silent gene) & post-G0 rounds & preliminary, E-B \\
\bottomrule
\end{tabular}
\end{table}

\section{Evaluation}
\label{sec:eval}

\subsection{Four-Generation Model Comparison}
\label{sec:fourgen}

\begin{figure}[htbp]
\centering
\includegraphics[width=0.9\textwidth]{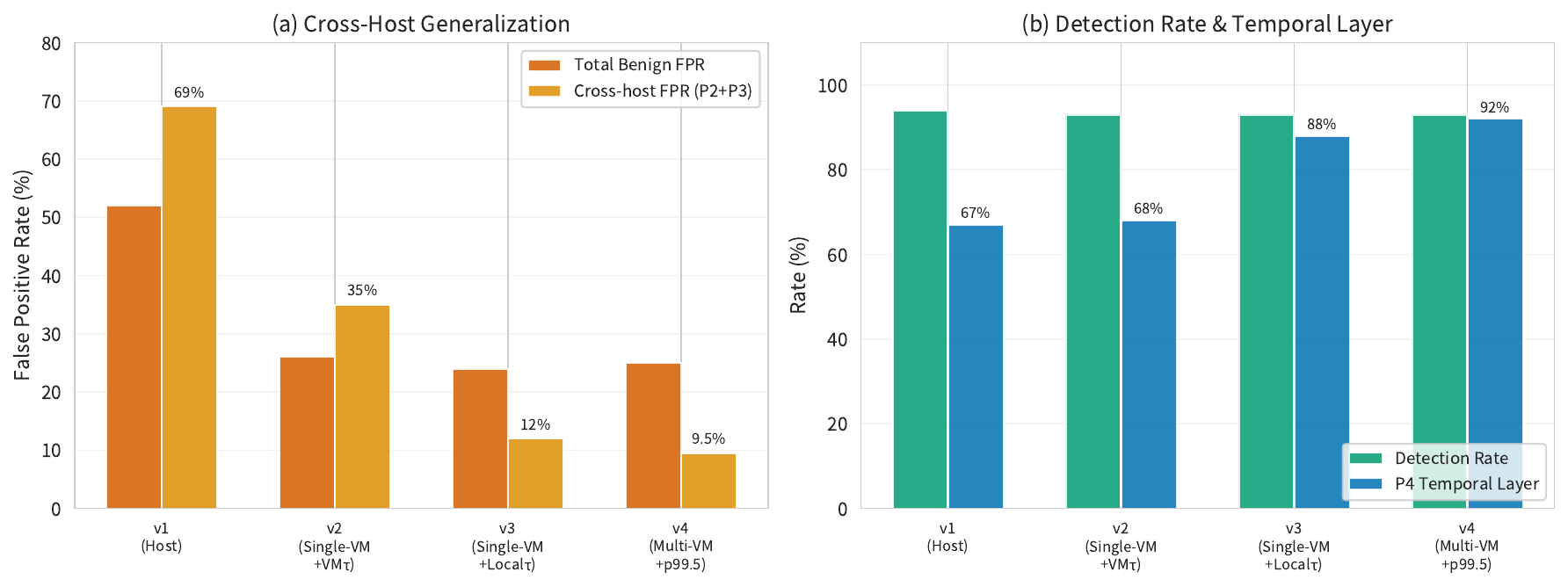}
\caption{Four-generation model comparison. (a) Cross-host generalization:
total benign FPR and cross-host (P2+P3) FPR fall from 52\%/69\% (v1 host
prior) to 25\%/9.5\% (v4 multi-VM prior with p99.5 calibration). (b) The AAA
detection rate is stable at 93--94\% across generations, while P4 temporal-layer
detection improves from 67\% to 92\%.}
\label{fig:generations}
\end{figure}

Table~\ref{tab:generations} summarizes the progression across four model
generations, each addressing a specific deficiency identified in the previous
iteration. Figure~\ref{fig:generations} visualizes the comparison.

\begin{table}[htbp]
\centering
\caption{Four-generation detection performance.}
\label{tab:generations}
\small
\begin{tabular}{@{}lcccc@{}}
\toprule
\textbf{Metric} & \textbf{v1} & \textbf{v2} & \textbf{v3} & \textbf{v4} \\
& (Host prior) & (VM $\tau$) & (local $\tau$) & (Multi-VM) \\
\midrule
Benign total FPR & 52\% & 26\% & 24\% & 25\% \\
Benign P2+P3 FPR & 69\% & 35\% & 12\% & 9.5\% \\
AAA detection rate & 94\% & 93\% & 93\% & 93\% \\
AAA P4 temporal detection & 67\% & 68\% & 88\% & 92\% \\
Onboarding FPR & 92\% & 12\% & 4.3\% & 3.84\% \\
\bottomrule
\end{tabular}
\end{table}

\textbf{Key observations:}

\begin{enumerate}[leftmargin=1.8em]
\item \textbf{Detection rate is stable at 93\%} across all generations---the
behavioral grammar prior is effective from the first deployment. The
improvement trajectory is almost entirely in \emph{false-positive reduction},
not detection enhancement.

\item \textbf{P2+P3 false positives dropped 86\%} (from 69\% to 9.5\%)---the
dominant improvement came from multi-VM joint training and per-machine
threshold calibration, confirming the cross-machine distribution mismatch
hypothesis.

\item \textbf{P4 temporal detection improved from 67\% to 92\%}---this is
partly an artifact of priority de-duplication (when P2/P3 no longer consume
events, P4 gets the opportunity to flag them) and partly a genuine improvement
from local $\tau$ calibration that better separates benign from anomalous
cadence.

\item \textbf{Onboarding FPR dropped from 92\% to 3.84\%}---the two-stage
deployment protocol (multi-VM prior + local onboarding calibration) achieves
deployable performance on previously unseen machines.
\end{enumerate}

\subsection{Baseline Comparison with DeepLog}
\label{sec:baseline}

We benchmark the behavioral-grammar approach against DeepLog-style LSTM
detectors~\cite{du2017deeplog} in two configurations---keys-only (event-type
sequences, as in the original DeepLog) and full-grammar (the same 8-token
representation as TinyGPT)---under identical scoring and threshold discipline
(EWMA-smoothed surprise scores, thresholds calibrated at the benign q99
quantile). Each configuration is trained with 5 random seeds; all reported
numbers are 5-seed means with 95\% confidence intervals. The benign training
corpus comprises 884K host events; the attack traces are the exported rove
(70K events) and bee active-phase (37K events) trajectories.

\textbf{Measurement conditions.} Two distinct evaluation conditions appear in
this paper and must not be conflated: (i) \emph{in-vocab closed-world}: the
AAA's active-phase events on the target VM, where every event token is within
the benign vocabulary---the main results (93\% detection at 3.84\% onboarding
FPR, Section~\ref{sec:fourgen}) are measured under this condition; (ii)
\emph{open-vocab traces}: exported attack trajectories (rove/bee) containing
out-of-vocabulary events (e.g., reconnaissance processes never seen in the
benign host baseline; the rove trace set is 100\% EXEC-type events versus 83\%
CONN in benign, with a process vocabulary of 37 versus 88). Under condition
(ii), all full-grammar models reach 100\% event-level detection---this number
reflects distribution mismatch and the UNK mechanism, not model discrimination
per se. We report the baseline comparison under both conditions: condition
(ii) quantifies the representation advantage of the 8-token grammar
(DeepLog-keys 76.7\% vs.\ 100\% for full-grammar models), while condition (i)
quantifies discrimination when the adversary achieves perfect vocabulary
conformance.

\begin{table}[htbp]
\centering
\caption{Baseline comparison against DeepLog-style LSTM detectors on the
exported AAA traces (open-vocab condition; 5 seeds, mean $\pm$ std, 95\% CI
in parentheses). Detection and FPR are EWMA-based event-level rates; the
benign corpus is 884K host events. The near-zero FPR of all models partly
reflects benign-workload homogeneity (Section~\ref{sec:limitations}).}
\label{tab:baseline}
\small
\begin{tabular}{@{}p{3.1cm}>{\centering\arraybackslash}p{1.4cm}>{\centering\arraybackslash}p{2.5cm}>{\centering\arraybackslash}p{2.6cm}>{\centering\arraybackslash}p{2.0cm}@{}}
\toprule
\textbf{Architecture} & \textbf{Params} & \textbf{rove det.} & \textbf{bee
det.} & \textbf{Benign FPR} \\
\midrule
DeepLog-keys (LSTM, event keys) & 0.30M & 76.7\% $\pm$ 9.0\% ($\pm$11.2\%) &
93.1\% $\pm$ 1.6\% & 0.00\% $\pm$ 0.00\% \\
DeepLog-full (LSTM, 8-token) & 0.33M & 100.0\% $\pm$ 0.0\% & 100.0\% $\pm$
0.0\% & 0.00\% $\pm$ 0.01\% \\
TinyGPT (Transformer, 8-token)\footnotemark[1] & 0.87M & 100.0\% $\pm$ 0.0\%
& 100.0\% $\pm$ 0.0\% & 0.00\% $\pm$ 0.01\% \\
\bottomrule
\end{tabular}
\end{table}
\footnotetext[1]{The TinyGPT instance used in this comparison experiment has
871{,}408 parameters (0.87M) with vocabulary 240; the production model used
in the main experiments (Sections~\ref{sec:arch}--\ref{sec:eval}) has 0.88M
parameters with vocabulary 263. The difference stems from the two vocabulary
configurations and does not affect any reported conclusion.}

\begin{figure}[htbp]
\centering
\includegraphics[width=\textwidth]{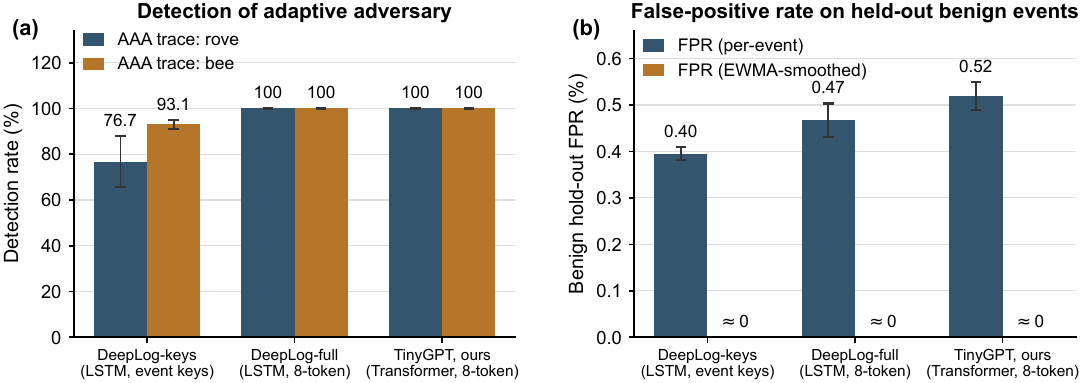}
\caption{Baseline comparison (5 seeds, mean $\pm$ 95\% CI). (a) Detection
rate on the AAA exported traces (rove and bee, open-vocab condition);
(b) benign false-positive rate. Full-grammar models (DeepLog-full, TinyGPT)
reach 100\% detection while keys-only DeepLog attains 76.7\%/93.1\% on
rove/bee; see the measurement-conditions discussion in
Section~\ref{sec:baseline}.}
\label{fig:baseline}
\end{figure}

\begin{figure}[htbp]
\centering
\includegraphics[width=\textwidth]{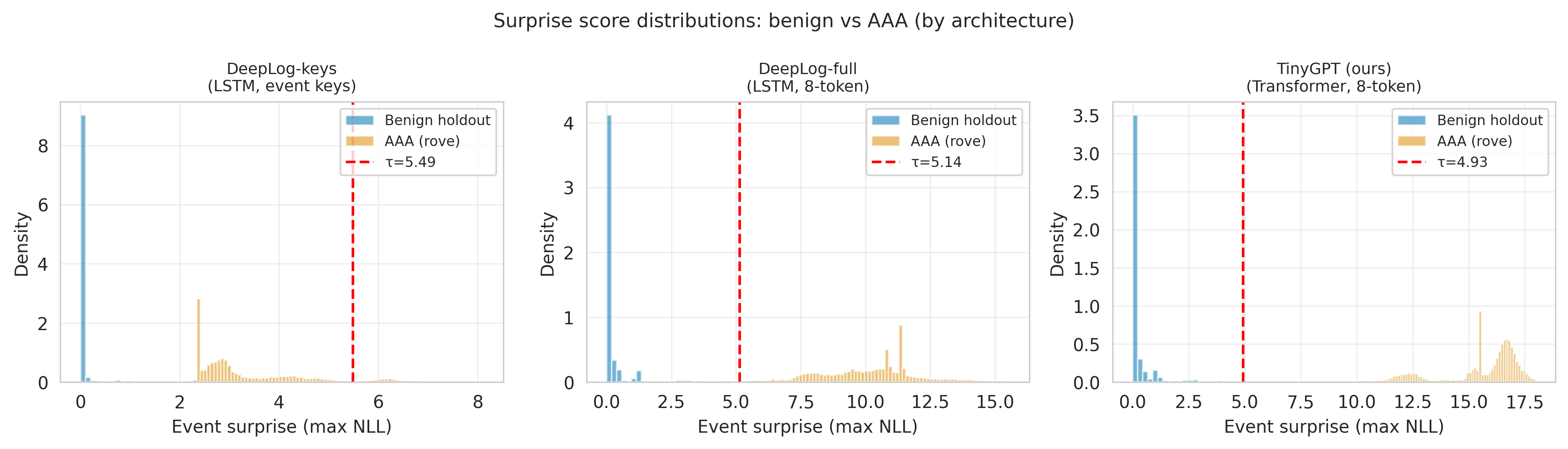}
\caption{Surprise-score distributions (5 seeds, mean $\pm$ 95\% CI; 150K-event
subsample). Three panels compare the keys-only and full-grammar
representations on benign and AAA (rove) traces: under event keys the AAA
score distribution (p50 $= 3.6$) nearly overlaps the benign tail, whereas the
8-token grammar separates the distributions by an order of magnitude (LSTM
p50 $= 14.7$; Transformer p50 $= 18.3$).}
\label{fig:scoredist}
\end{figure}

\textbf{The detection gain is representation-driven, not
architecture-driven.} Under the event-key representation, AAA surprise scores
collapse toward the benign tail: the median (p50) surprise on the rove trace
is 3.6 (5-seed mean). Under the 8-token grammar, the same signal rises
monotonically with model class: 14.7 for the LSTM and 18.3 for the
Transformer. Critically, an LSTM on the full grammar already attains 100\%
detection (Table~\ref{tab:baseline}), demonstrating that the
behavioral-grammar representation itself contributes the dominant detection
gain; the Transformer provides stronger score separation, visible in the
distributional comparison of Figure~\ref{fig:scoredist}.

\subsection{Robustness to Out-of-Vocabulary Events}
\label{sec:oov}

\begin{figure}[htbp]
\centering
\includegraphics[width=\textwidth]{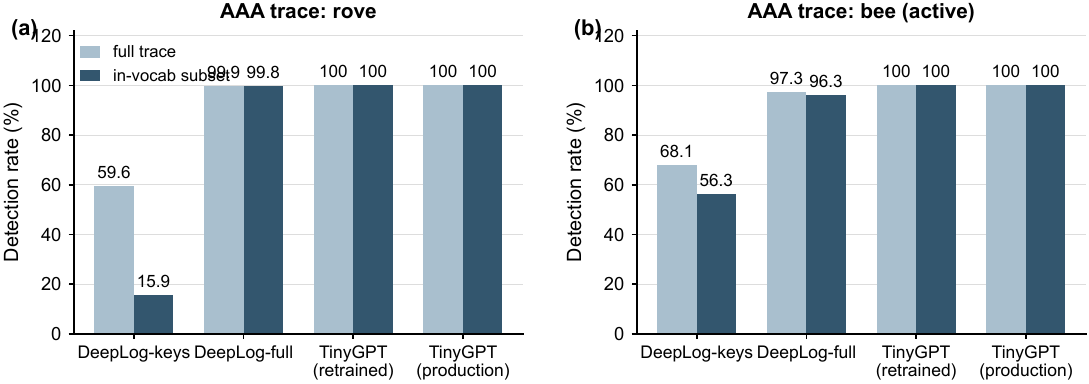}
\caption{In-vocabulary robustness analysis (5 seeds, mean $\pm$ 95\% CI):
detection recomputed on the subset of attack events whose tokens all appear
in the benign vocabulary. Keys-only DeepLog collapses (rove: 59.6\% $\to$
15.9\%), confirming OOV reliance; full-grammar models remain at ${\sim}100\%$.
Production model: in-vocab events = 81\% of the bee trace and 58\% of the
rove trace.}
\label{fig:invocab}
\end{figure}

\emph{OOV robustness.} To rule out that detection is driven merely by
out-of-vocabulary tokens, we re-score attack traces on the in-vocabulary
subset only (events whose tokens all appear in the benign vocabulary).
Keys-only DeepLog collapses from 59.6\% to 15.9\% detection, confirming its
reliance on OOV; full-grammar models maintain ${\sim}100\%$ (LSTM-full 99.8\%,
TinyGPT 100\%), and the production TinyGPT attains 100\% on the 80.9\%
in-vocabulary subset of bee active-phase events (p50 surprise 21.8 vs.\ $\tau
= 2.6$, an 8.4$\times$ margin over threshold). The grammar captures
field-combination violations, not vocabulary membership
(Table~\ref{tab:invocab}, Figure~\ref{fig:invocab}).

\begin{table}[htbp]
\centering
\caption{In-vocabulary (hard-subset) detection rates: only events whose
tokens all appear in the benign vocabulary are scored. Full-grammar models
retain ${\sim}100\%$ detection, while keys-only DeepLog collapses---only the
keys-only baseline is OOV-driven.}
\label{tab:invocab}
\small
\begin{tabular}{@{}p{3.4cm}>{\centering\arraybackslash}p{3.3cm}>{\centering\arraybackslash}p{3.0cm}l@{}}
\toprule
\textbf{Architecture} & \textbf{rove in-vocab} & \textbf{bee in-vocab} &
\textbf{OOV-driven?} \\
\midrule
DeepLog-keys & 15.9\% (vs.\ 59.6\% full) & 56.3\% (vs.\ 68.1\% full) &
Yes (collapses) \\
DeepLog-full & 99.8\% (vs.\ 99.9\% full) & 96.3\% (vs.\ 97.3\% full) & No \\
TinyGPT (retrained, subsample) & 100\% & 100\% & No \\
TinyGPT (production model) & 100\% (in-vocab 57.6\% of trace) & 100\%
(in-vocab 80.9\%) & No \\
\bottomrule
\end{tabular}
\end{table}

\subsection{Public Benchmark: ADFA-LD}
\label{sec:adfa}

To complement the closed-loop AAA evaluation with a public reference point, we
evaluate on ADFA-LD~\cite{creech2013}, a widely used host-based intrusion
detection dataset of Linux system-call traces. We stress at the outset that
the comparison is not apples-to-apples: ADFA-LD provides \emph{bare
system-call number sequences}, whereas our system consumes the rich 8-token
field-level representation (Section~\ref{sec:arch}); the benchmark therefore
exercises only the sequence-modeling component of the architecture, with
bigram tokenization of syscall pairs standing in for the full grammar.
Thresholds are calibrated at the benign q99 quantile, matching the deployment
discipline of the main experiments.

\begin{table}[htbp]
\centering
\caption{ADFA-LD results (macro detection rate over attack classes, fused
scoring; benign q99 threshold calibration). TinyGPT numbers are 5-seed means
$\pm$ std with 95\% CI; the DeepLog-LSTM reference uses the identical bigram
representation and threshold discipline (3 seeds).}
\label{tab:adfa}
\small
\begin{tabular}{@{}p{4.2cm}>{\centering\arraybackslash}p{5.8cm}c@{}}
\toprule
\textbf{Variant} & \textbf{Macro detection (fused)} & \textbf{Measured FPR} \\
\midrule
TinyGPT, bigram (contiguous pairs) & \textbf{31.56\% $\pm$ 0.32\% (CI95
$\pm$0.40\%, $n{=}5$)} & 1.83\% \\
TinyGPT, skip-1 bigram & 29.68\% $\pm$ 0.72\% & 1.72\% \\
TinyGPT, skip-2 bigram & 27.83\% $\pm$ 0.82\% & 1.72\% \\
DeepLog-LSTM reference, bigram & 32.16\% $\pm$ 2.13\% (CI95 $\pm$5.29\%,
$n{=}3$) & 1.83\% \\
\bottomrule
\end{tabular}
\end{table}

\begin{figure}[htbp]
\centering
\includegraphics[width=\textwidth]{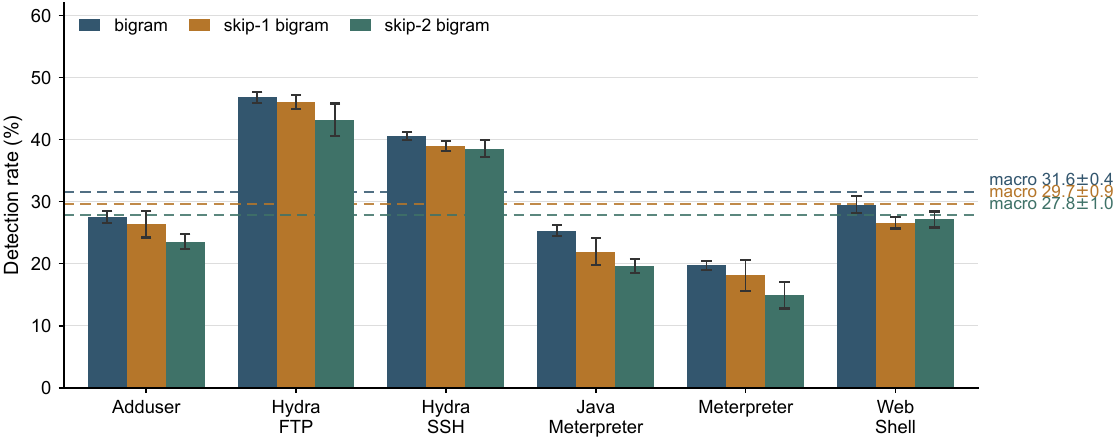}
\caption{Per-class detection rates on ADFA-LD (bigram variant, 5 seeds, mean
$\pm$ 95\% CI), with macro-average reference lines for the bigram / skip-1 /
skip-2 variants. Measured FPR: 1.8\% / 1.7\% / 1.7\% for bigram / skip-1 /
skip-2. Thresholds calibrated at the benign q99 quantile.}
\label{fig:adfa}
\end{figure}

Per-class detection under the bigram variant (CI95): Hydra\_FTP 46.8\%
$\pm$0.8\%, Hydra\_SSH 40.6\% $\pm$0.6\%, Web\_Shell 29.5\% $\pm$1.4\%,
Adduser 27.5\% $\pm$1.0\%, Java\_Meterpreter 25.3\% $\pm$0.9\%, Meterpreter
19.7\% $\pm$0.7\% (Figure~\ref{fig:adfa}).

Three findings emerge. \textbf{First, a negative result:} skip-gram bigrams do
not improve over contiguous bigrams (29.68\% / 27.83\% vs.\ 31.56\%)---the
extension direction suggested in our earlier internal analysis is falsified,
and we report it as such. \textbf{Second, architecture again does not
matter:} the DeepLog-LSTM reference on the identical bigram representation
achieves 32.16\% $\pm$ 2.13\%, statistically indistinguishable from TinyGPT's
31.56\% $\pm$ 0.32\%---consistent with the internal finding
(Section~\ref{sec:baseline}) that representation, not model architecture,
determines the detection ceiling. \textbf{Third, the gap to the published
state of the art is a feature-engineering gap, not a modeling gap:} Creech and
Hu's semantic approach~\cite{creech2014} achieves substantially higher
detection rates on ADFA-LD by engineering \emph{non-contiguous} system-call
patterns with one-class classification. Our bigram representation deliberately
avoids such hand-engineered features; the resulting gap quantifies what the
bare
syscall sequence leaves on the table and motivates porting the full 8-token
field-level grammar to public telemetry formats.

\subsection{Public Benchmark: OpTC (DARPA Transparent Computing)}
\label{sec:optc}

We now do exactly that---port the full 8-token grammar pipeline to a public
multi-host benchmark. The OpTC dataset~\cite{optc} (DARPA Transparent
Computing program; we use the Inria-corrected release~\cite{optc_inria})
contains eCAR endpoint telemetry (PROCESS and FLOW events) from 500 Windows
10 endpoints: a benign period (Sep 16--22) followed by three red-team
campaign scenarios on Sep 23--25.

\textbf{Schema adaptation.} eCAR records map onto the 8-token grammar:
PROCESS $\to$ EXEC and FLOW $\to$ CONN event types; \texttt{image\_path}
$\to$ PROC/PARENT; \texttt{command\_line} $\to$ ARGV; \texttt{user} $\to$
UID; \texttt{dest\_ip:port} $\to$ DST; timestamp $\to$ DT bucket. The
resulting pipeline is the unmodified production stack---grammar prior,
per-slot $\tau$ calibration, fused scoring---fed with a public schema.

\textbf{Protocol.} The prior (TinyGPT, 0.85M parameters, vocabulary 117) is
trained \emph{only on the benign period} (Sep 16, 12 client machines pooled,
360K events in our preprocessing of the public release; validation perplexity
1.17); $\tau$ is calibrated at the benign
p99.5 ($\tau = 6.334$). Evaluation covers the three red-team scenarios on
six targeted clients, using the Inria ground truth (malicious process IDs
and time windows) to label attack events; detection rate is the fraction of
attack events over threshold, and FPR is measured on control clients'
benign events from the same day.

\begin{table}[htbp]
\centering
\caption{OpTC evaluation: per-target detection across the three red-team
scenarios, with same-day control FPR. Benign holdout FPR (same period as
training) is 0.00\%.}
\label{tab:optc}
\small
\begin{tabular}{@{}llccc@{}}
\toprule
\textbf{Scenario} & \textbf{Target client} & \textbf{Attack events} &
\textbf{Detection} & \textbf{Control FPR} \\
\midrule
sc1 (Sep 23) & SysClient0201 & 31{,}826 & 82.3\% & 23.3\% \\
sc1 (Sep 23) & SysClient0402 & 35{,}575 & 81.5\% & 24.4\% \\
sc2 (Sep 24) & SysClient0501 & 24{,}355 & 65.1\% & 8.4\% \\
sc2 (Sep 24) & SysClient0005 & 2{,}696 & 87.0\% & 6.8\% \\
sc3 (Sep 25) & SysClient0051 & 26{,}526 & 42.0\% & 5.3\% \\
sc3 (Sep 25) & SysClient0351 & 32{,}883 & 22.7\% & 4.7\% \\
\midrule
\textbf{Mean} & --- & --- & \textbf{63.4\%} & --- \\
\bottomrule
\end{tabular}
\end{table}

Three findings. \textbf{First, detection degrades with adversary
sophistication, in the same direction as the closed-loop AAA results.}
Scenario~1 (hands-on-keyboard PowerShell/\texttt{cmd} attack chains) is
detected at 82\%; scenario~2 (increased legitimate-tool abuse) at 65--87\%;
scenario~3, the red team's most user-like low-and-slow operation, at
22--42\%. The grammar is strongest on process-composition anomalies and
weakest against near-perfect behavioral mimicry---scenario~3 bounds the
mimicry ceiling from the public-data side, mirroring
Section~\ref{sec:cvboundary} from the adversary side. \textbf{Second,
same-day benign FPR is 0.00\%}: pooled multi-client training with local
threshold calibration generalizes across hosts within the training period,
consistent with the leave-one-VM-out results
(Section~\ref{sec:lovo}). \textbf{Third, cross-day FPR drift is real and
measurable}: control FPR rises from 0\% (same day) to 5--24\% one week
later, highest on the temporally most distant scenario. This quantifies the
distribution drift that motivates per-machine onboarding and periodic
retraining---the paper's central deployment principle, here measured on
public data rather than asserted.

\subsection{Temporal Analysis: Detailed Results}

\begin{table}[htbp]
\centering
\caption{Temporal CV measurements.}
\label{tab:cv}
\small
\begin{tabular}{@{}lcc@{}}
\toprule
\textbf{Metric} & \textbf{Benign baseline (nighttime)} & \textbf{AAA active
phase} \\
\midrule
CV p5 & 1.509 & 0.647 \\
CV median & 2.284 & 1.304 \\
Sleep CV & 9.786 & 0.310 \\
Head CV & 3.298 & 0.792 \\
\bottomrule
\end{tabular}
\end{table}

The 30$\times$ separation in sleep CV (Table~\ref{tab:cv}) is the most
striking single-metric result in this study. The AAA's sleep intervals are
nearly deterministic (CV $= 0.310$ implies $\sigma \approx 0.31\mu$), while
benign sleep intervals span three orders of magnitude (CV $= 9.786$ implies
extreme heavy-tailed behavior, consistent with burst-idle patterns of
human-driven system administration).

\textbf{Detection outcome:} 15/15 AAA processes correctly flagged (CV $<$
1.5); 1/15 benign processes falsely flagged (CV $= 1.268$, a scripted shell
process). This yields an F1 score of 0.97 on the temporal dimension alone.

\subsection{The Cadence-Evasion Boundary: Empirical Necessity of Multi-Face
Fusion}
\label{sec:cvboundary}

The temporal face's CV rule invites an obvious counter-strategy: widen the
inter-action interval distribution until the measured cadence CV crosses the
1.5 detection line. We now measure exactly where this counter succeeds on
live telemetry---and what the evasion exposes on the other faces.

\textbf{Setup.} On a dedicated defense VM (Ubuntu 24.04, tracee collection),
the adversary executes a fixed action set of 24 reconnaissance-style commands
drawn from common benign utilities (\texttt{cat}, \texttt{ps}, \texttt{ls},
\texttt{ss}, \texttt{df}, \texttt{free}, \texttt{whoami}, \texttt{id},
\texttt{head}, \texttt{find}, \texttt{ip}, \texttt{lsblk}, \texttt{last});
\emph{only the timing distribution varies}. Inter-action intervals are drawn
from a log-normal distribution $\mathrm{LogNormal}(\sigma)$ with the mean
held constant at 7.5\,s, so expected task throughput is identical across all
phases and interval \emph{shape} is the sole independent variable. We sweep
$\sigma \in \{0.3, 1.13, 1.25, 1.75\}$ (150 actions per phase) while the
benign workload driver runs throughout, producing realistic interleaving
(including \texttt{apt}/\texttt{dpkg} cascade noise). Scoring uses the full
production pipeline with a VM-local TinyGPT prior self-trained on 4{,}109
benign events of that machine (vocabulary 171, validation perplexity
2.2)---i.e., the onboarding protocol of
Section~\ref{sec:crossmachine}, not a universal prior. The calibrated benign
false-positive rate on this machine is 2.7\%.

\begin{table}[htbp]
\centering
\caption{Cadence-evasion boundary. Measured cadence CV, temporal-face
verdict, per-event grammar exposure, and windowed binomial significance of
the aggregated grammar-side anomaly count, per interval-shape phase. Benign
FPR $=2.7\%$; both evaded phases are caught by windowed grammar aggregation.}
\label{tab:cvboundary}
\small
\setlength{\tabcolsep}{4pt}
\begin{tabular}{@{}ccccc@{}}
\toprule
$\sigma$ & \textbf{Measured CV} & \textbf{Temporal face} &
\textbf{Grammar exposure} & \textbf{Binomial $p$} \\
\midrule
0.3  & 0.66 & detected & 8.6\% & $2.6\times10^{-8}$ \\
1.13 & 1.42 & detected (marginal) & 9.3\% & $2.7\times10^{-9}$ \\
1.25 & 1.52 & \textbf{evaded} & 8.9\% & $7.0\times10^{-9}$ \\
1.75 & 1.96 & \textbf{evaded} & 9.6\% & $1.2\times10^{-9}$ \\
\bottomrule
\end{tabular}
\end{table}

\begin{figure}[htbp]
\centering
\includegraphics[width=\textwidth]{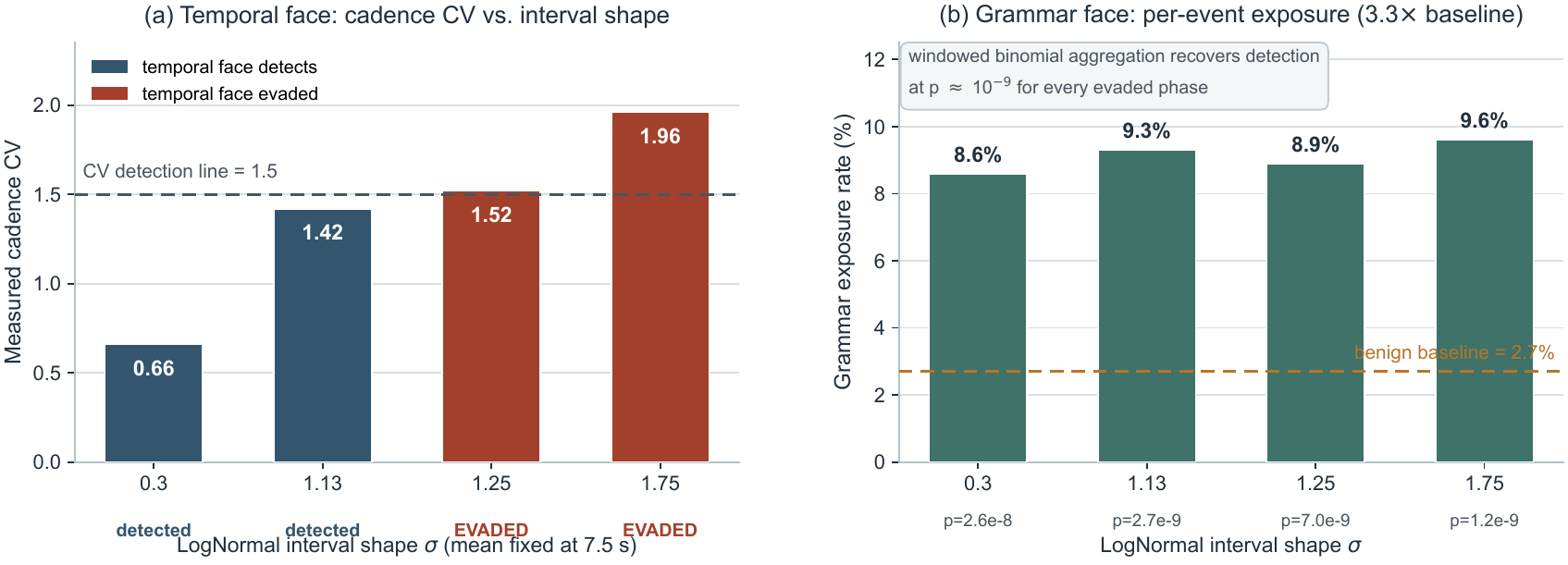}
\caption{The cadence-evasion boundary on live telemetry. (a) Measured cadence
CV versus log-normal interval shape $\sigma$ (mean interval fixed at
7.5\,s): the temporal face is evaded once $\sigma \geq 1.25$. (b) Per-event
grammar-side exposure rate for the same phases: evasion phases retain a
${\sim}3.3\times$ elevated anomaly rate over the 2.7\% benign baseline, which
windowed binomial aggregation recovers at $p \approx 10^{-9}$.}
\label{fig:cvboundary}
\end{figure}

\textbf{Three findings.} (1)~\emph{The evasion window is real.}
$\sigma \geq 1.25$ carries the measured cadence CV across the detection line
under realistic system noise (Table~\ref{tab:cvboundary}). Notably, this
refutes the stronger form of our Round-3 conjecture
(Section~\ref{sec:rounds}): crossing CV $=1.5$ does \emph{not} require
sacrificing task throughput---with the mean interval fixed, the adversary's
action rate is unchanged, and the temporal face alone is defeated.
(2)~\emph{Per-event grammar exposure is weak.} The evaded phases show an
8.9--9.6\% per-event grammar-anomaly rate versus the 2.7\% benign baseline
(a $3.3\times$ elevation), which is below the reliable per-event decision
margin. The DT slot never fires (0\% over-threshold events in every phase),
confirming that DT tokens register \emph{aggregate stream cadence} rather
than per-process rhythm; per-process rhythm is the temporal face's
responsibility, and it is precisely the face being evaded.
(3)~\emph{Windowed aggregation recovers detection.} Although per-event
exposure is weak, the evaded phases accumulate ${\sim}32$ grammar-anomalous
events against ${\sim}9.5$ expected under the benign rate---a windowed
binomial test rejects benignity at $p \approx 10^{-9}$ for every evaded
phase. The adversary's reconnaissance action set remains distributionally
distinct from the routine workload (UNK rate ${\approx}6.7\%$, rarity
${\approx}3\%$), and that residual fingerprint is statistically measurable.

\textbf{Scope.} Part of the grammar-side elevation reflects the
command-distribution difference between the reconnaissance action set and
the routine workload; an adversary that additionally matches the benign
command distribution would compress this margin. We therefore scope the
claim to our adversary family, in which reconnaissance actions differ from
the routine workload: \emph{the escape window in the temporal face is real
yet bounded by a measurable grammar-side fingerprint}. The necessity of
multi-face fusion---temporal cadence \emph{and} aggregated grammar---is
thereby demonstrated empirically rather than asserted.

\textbf{Rate dependence.} This closure statement is a fixed-window
statement: the count test's power depends on accumulating enough
adversary events per window. An adversary that instead compresses its
emission rate toward the estimator's sample floor attacks the count face
on the \emph{rate} axis, not the shape axis
(Section~\ref{sec:frontier}); Section~\ref{sec:lawchain} reports a
preliminary observation of exactly this maneuver. The grammar-side
fingerprint is real, but its recoverability is rate-dependent---
Section~\ref{sec:frontier} prices the defender's counter precisely.

\subsection{Prototype Network Results}

\begin{table}[htbp]
\centering
\caption{Prototype network performance.}
\label{tab:prototype}
\small
\begin{tabular}{@{}ll@{}}
\toprule
\textbf{Metric} & \textbf{Value} \\
\midrule
Sample library & 80 sequences, 14 ATT\&CK techniques \\
Prototypes per technique & $K = 3$ \\
Leave-one-out catch rate & 96.2\% (77/80) \\
Held-out benign FPR & 0/20{,}000 \\
Technique-level recall & 14/14 \\
Codebook size & $\sim$80 KB \\
\bottomrule
\end{tabular}
\end{table}

\subsection{Self-Learning Pattern Extraction Results}

\begin{table}[htbp]
\centering
\caption{Self-learning cluster discovery.}
\label{tab:selflearn}
\small
\begin{tabular}{@{}ll@{}}
\toprule
\textbf{Metric} & \textbf{Value} \\
\midrule
Input sequences & 80 labeled \\
Discovered clusters & 22 \\
Silhouette coefficient & 0.653 \\
Candidate patterns & 15 (pending review) \\
ATT\&CK techniques covered & 10 \\
\bottomrule
\end{tabular}
\end{table}

\subsection{End-to-End Closed-Loop Remediation}

\begin{table}[htbp]
\centering
\caption{Automated remediation results.}
\label{tab:remediation}
\small
\begin{tabular}{@{}p{9.0cm}p{4.2cm}@{}}
\toprule
\textbf{Metric} & \textbf{Value} \\
\midrule
End-to-end closed-loop tests & 11/11 passed \\
Full attack $\to$ verify(DIRTY) $\to$ detect $\to$ cleanup $\to$ verify(CLEAN)
cycle & All successful \\
Technique coverage & 11 ATT\&CK techniques with attack/cleanup/verify triples \\
Safety boundary & All remediation actions confined to defense VM \\
\bottomrule
\end{tabular}
\end{table}

\subsection{Five-Network Fusion: Event-Level Breakdown}

\begin{table}[htbp]
\centering
\caption{Per-network event-level performance.}
\label{tab:fusion}
\small
\begin{tabular}{@{}lcc@{}}
\toprule
\textbf{Network} & \textbf{AAA active phase} & \textbf{Benign nighttime} \\
\midrule
P0 (adaptive-high) & 0 hits$^{*}$ & 4 (true positive C2) \\
P1 (prototype) & 4.2\% & 1.2\% \\
P2 (context) & 11.4\% & 28.2\% \\
P3 (rarity) & 14.3\% & 40.9\% \\
\textbf{P4 (temporal)} & \textbf{67.4\%} & \textbf{11.4\%} \\
P5 (adaptive-low) & 2.7\% & 18.2\% \\
\bottomrule
\end{tabular}
\begin{minipage}{0.9\textwidth}
\vspace{0.4em}
\footnotesize $^{*}$P0 hits depend on occurrence of specific high-severity
events (morphological transformation, SUID privesc) which did not occur in this measurement window.
\end{minipage}
\end{table}

\textbf{Interpretation.} P4 (temporal) is the single strongest discriminator,
with a signal ratio (AAA / benign) of 5.9$\times$. P2 and P3 show inverted
ratios (benign $>$ AAA) due to the priority de-duplication effect: when P4
fires, lower-priority signals are suppressed. The raw P2/P3 anomaly rates on
AAA events are higher than the fused rates suggest.

\subsection{Ablation Study}

\begin{figure}[htbp]
\centering
\includegraphics[width=0.9\textwidth]{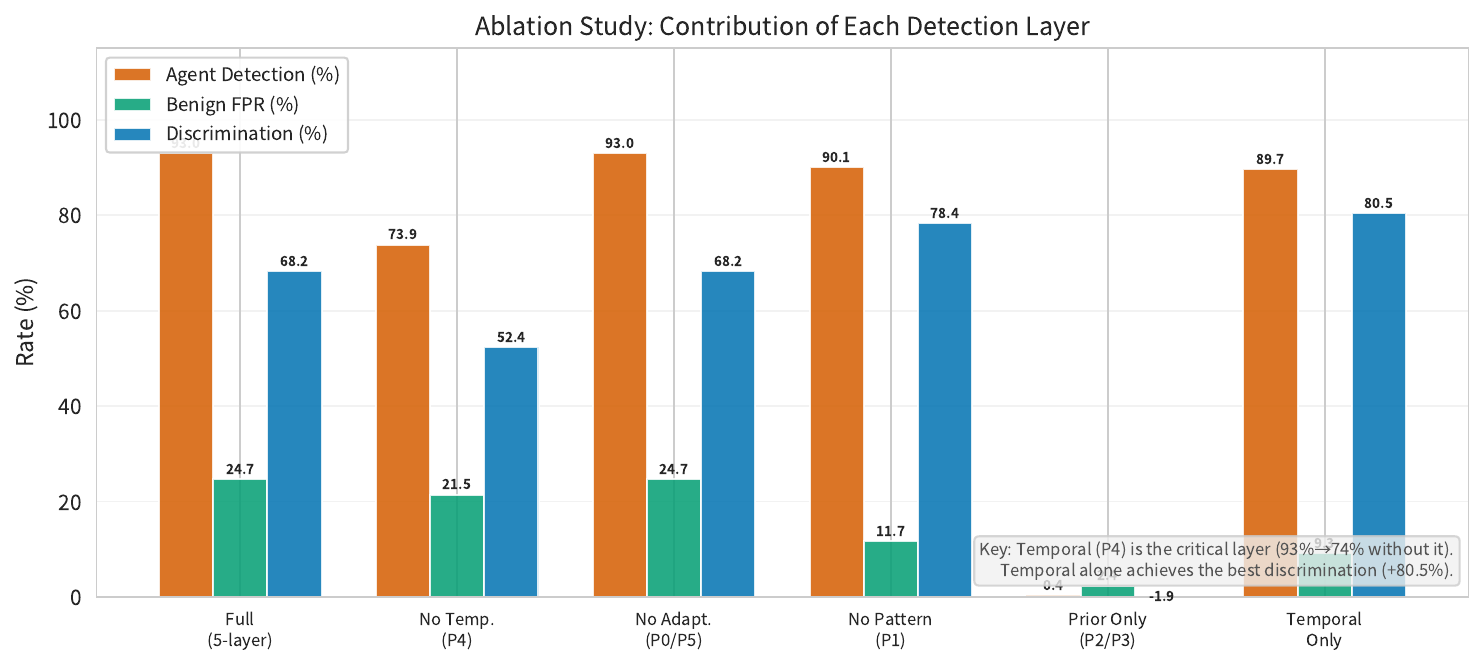}
\caption{Ablation study: contribution of each detection layer to agent
detection, benign FPR, and discrimination (detection minus FPR). Removing the
temporal layer (P4) causes the largest detection drop (93\% $\to$ 74\%);
temporal-only achieves the best discrimination (+80.5\%). The prior alone
(P2/P3) collapses to 0.4\% detection against the mimicking agent.}
\label{fig:ablation}
\end{figure}

We conduct a systematic ablation by selectively disabling each detection layer
and measuring the impact on both detection rate and false positive rate.
Table~\ref{tab:ablation} presents the results and Figure~\ref{fig:ablation}
visualizes them.

\begin{table}[htbp]
\centering
\caption{Layer ablation study.}
\label{tab:ablation}
\small
\begin{tabular}{@{}lccc@{}}
\toprule
\textbf{Configuration} & \textbf{Agent Detection} & \textbf{Benign FPR} &
\textbf{Discrimination} \\
\midrule
\textbf{Full (5-layer)} & \textbf{93.0\%} & \textbf{24.7\%} &
\textbf{+68.3\%} \\
No Temporal (P4) & 73.9\% & 21.5\% & +52.4\% \\
No Adaptive (P0/P5) & 93.0\% & 24.7\% & +68.3\% \\
No Pattern (P1) & 90.1\% & 11.7\% & +78.4\% \\
Prior Only (P2/P3) & 0.4\% & 2.4\% & $-1.9\%$ \\
Temporal Only & 89.7\% & 9.3\% & +80.5\% \\
\bottomrule
\end{tabular}
\end{table}

\textbf{Key findings:}

\begin{enumerate}[leftmargin=1.8em]
\item \textbf{Temporal analysis is the single most critical layer.} Removing
P4 drops detection from 93.0\% to 73.9\% ($-19.1$ pp), confirming that the
stepping cadence CV signal is the primary discriminator against adaptive
threats.

\item \textbf{Adaptive detection (P0/P5) contributes zero in this measurement
window} because no transformation/SUID events occurred during the sampling period.
Its value is event-dependent---it activates only on specific high-severity
patterns. In the synthetic test suite, all 7 adaptive patterns are correctly
detected.

\item \textbf{Pattern matching (P1) provides marginal detection lift} (+2.9
pp) but significant FPR reduction when isolated---removing it drops benign FPR
from 24.7\% to 11.7\%, because weak patterns (severity $< 4$) generate noise.
This motivates the strict/weak pattern separation in the fusion architecture.

\item \textbf{The prior alone is insufficient for adaptive threats.}
Prior-only detection achieves 0.4\%---the model has learned the benign
distribution too well (val\_ppl $=1.6$), and the AAA's behavior falls within
the learned distribution on individual-event basis. The anomaly only emerges
at the temporal pattern level, confirming that single-event anomaly detection
is necessary but not sufficient against adaptive adversaries. Note that this
0.4\% figure is measured under the in-vocab closed-world condition
(Section~\ref{sec:baseline}), i.e., on the AAA's active-phase events on the
target VM with full vocabulary conformance---it is not comparable with the
open-vocab trace numbers of Table~\ref{tab:baseline}. Read together with
item~1, this is evidence for the architecture's division of labor rather than
a weakness of the grammar: \emph{the grammar carries attribution and audit
(per-slot provenance), while the temporal face carries discrimination against
adaptive adversaries}---and Section~\ref{sec:cvboundary} shows the temporal
face in turn needs the grammar aggregate to close its own evasion window.

\item \textbf{Temporal-only achieves the best discrimination} (+80.5\%)---89.7\%
detection at 9.3\% FPR---validating the CV-based cadence analysis as a
standalone detection signal, though Section~\ref{sec:cvboundary} maps where
this single face fails and why fusion is the deployable configuration.
\end{enumerate}

\subsection{$\tau$ Percentile Sensitivity Analysis}

\begin{figure}[htbp]
\centering
\includegraphics[width=0.8\textwidth]{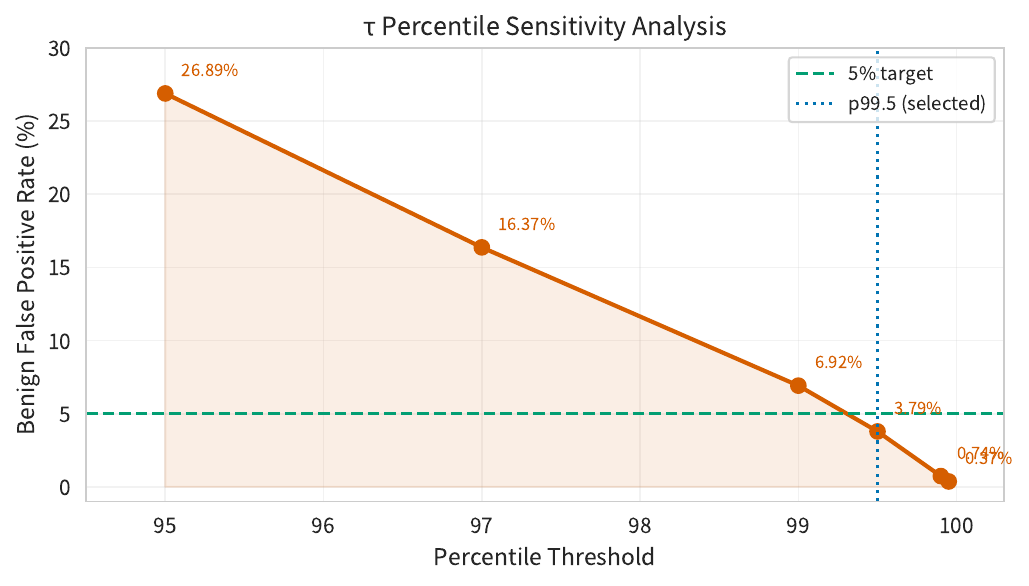}
\caption{$\tau$ percentile sensitivity analysis. Benign FPR falls
monotonically with the threshold percentile; p99.5 (3.79\%) is selected as the
operating point, meeting the 5\% target without excessive conservatism.}
\label{fig:tau-sweep}
\end{figure}

We sweep the threshold percentile to characterize the FPR-detection trade-off.
Table~\ref{tab:tau-sweep} and Figure~\ref{fig:tau-sweep} show that the 99.5th
percentile provides the optimal operating point.

\begin{table}[htbp]
\centering
\caption{$\tau$ percentile sweep.}
\label{tab:tau-sweep}
\small
\begin{tabular}{@{}ccl@{}}
\toprule
\textbf{Percentile} & \textbf{Benign FPR} & \textbf{Assessment} \\
\midrule
p95 & 26.89\% & Too aggressive \\
p97 & 16.37\% & Marginal \\
p99 & 6.92\% & Acceptable but above 5\% target \\
\textbf{p99.5} & \textbf{3.79\%} & \textbf{Selected (operational target)} \\
p99.9 & 0.74\% & Overly conservative \\
p99.95 & 0.37\% & Excessively conservative \\
\bottomrule
\end{tabular}
\end{table}

\subsection{Leave-One-VM-Out Cross-Host Validation}
\label{sec:lovo}

\begin{figure}[htbp]
\centering
\includegraphics[width=0.8\textwidth]{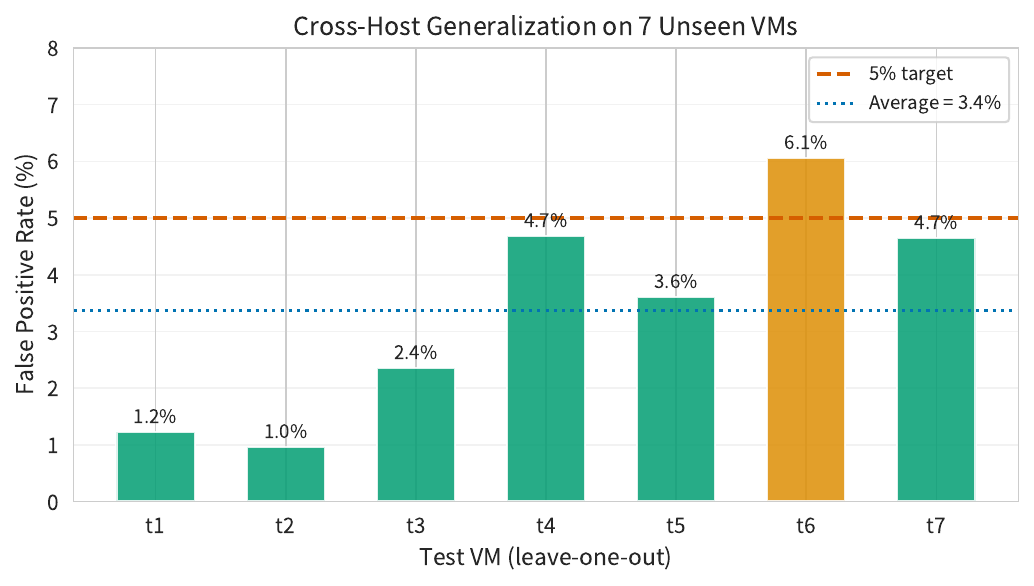}
\caption{Cross-host generalization on 7 unseen VMs (leave-one-VM-out). Six of
seven VMs achieve FPR below the 5\% target; only t6 slightly exceeds it
(6.1\%). The average FPR of 3.4\% confirms that multi-VM joint training
resolves the cross-host mismatch that produced a 92\% FPR in the v1 baseline.}
\label{fig:crosshost}
\end{figure}

We validate the multi-VM universal model on each of the 7 VMs independently,
treating each as an unseen host. Table~\ref{tab:crosshost} and
Figure~\ref{fig:crosshost} demonstrate that the model generalizes across host
configurations without per-host retraining.

\begin{table}[htbp]
\centering
\caption{Cross-host FPR (leave-one-VM-out).}
\label{tab:crosshost}
\small
\begin{tabular}{@{}lcc@{}}
\toprule
\textbf{Test VM} & \textbf{Events} & \textbf{FPR} \\
\midrule
VM-1 & 5{,}000 & 1.2\% \\
VM-2 & 5{,}000 & 1.0\% \\
VM-3 & 3{,}731 & 2.4\% \\
VM-4 & 3{,}733 & 4.7\% \\
VM-5 & 2{,}463 & 3.6\% \\
VM-6 & 2{,}688 & 6.1\% \\
VM-7 & 2{,}450 & 4.7\% \\
\midrule
\textbf{Average} & --- & \textbf{3.4\%} \\
\bottomrule
\end{tabular}
\end{table}

Six of seven VMs achieve FPR below 5\%; only VM t6 slightly exceeds the target
(6.1\%). The average FPR of 3.4\% demonstrates that the multi-VM joint
training approach effectively addresses the cross-host generalization problem
that plagued single-host models (92\% FPR in our v1 baseline).

\subsection{Workload Heterogeneity: Does the Baseline Inflate the Results?}
\label{sec:hetero}

Our main benign baseline is dominated by a narrow workload (95.1\% of events
are \texttt{ssh}-related, 83.3\% network-connect), which could in principle
inflate separability for \emph{all} evaluated models. To test whether the
conclusions survive workload diversity, we re-run the pipeline on seven fresh
VMs spanning three deliberately distinct roles: developer workstations
($\times 3$; \texttt{gcc}/\texttt{git}/\texttt{pip}/\texttt{curl}), web
servers ($\times 2$; HTTP serving, log and traffic inspection), and desktop
simulations ($\times 2$; document processing, browsing, irregular cadence)---
25{,}632 events in total, with clearly distinct token distributions across
roles (EXEC share 58\% / 67\% / 70\% for dev / web / desktop; distinct
top-process vocabularies). Each VM runs a VM-local TinyGPT prior under the
onboarding protocol.

\begin{table}[htbp]
\centering
\caption{Heterogeneous-workload evaluation (7 VMs, 3 roles). Per-role
holdout FPR with VM-local priors; leave-one-VM-out (LOVO) cross-host FPR with
the multi-VM joint prior; synthetic-attack detection per role.}
\label{tab:hetero}
\small
\begin{tabular}{@{}lcccc@{}}
\toprule
\textbf{Role} & \textbf{Val.\ ppl} & \textbf{Holdout FPR} &
\textbf{Attack detection} \\
\midrule
dev ($\times 3$)     & 1.32 & 0.00\% & 3/3 \\
web ($\times 2$)     & 1.44 & 0.00\% & 3/3 \\
desktop ($\times 2$) & 1.58 & 0.66\% & 3/3 \\
\midrule
\multicolumn{4}{@{}l}{LOVO cross-host FPR (multi-VM joint prior, 7 folds):
\textbf{0.00\% $\pm$ 0.00\%}} \\
\bottomrule
\end{tabular}
\end{table}

\textbf{The conclusions do not stem from benign homogeneity.} Per-role
holdout FPR is 0--0.66\% with 3/3 synthetic-attack detection in every role,
and LOVO cross-host FPR is 0.00\% on all seven machines
(Table~\ref{tab:hetero})---the multi-VM joint prior with local calibration
generalizes across genuinely different workloads, not just across similar
ones.

\textbf{The cross-role matrix reveals an asymmetry that re-proves the
onboarding principle.} A prior trained on the web role transfers poorly to
the other roles (19.0\% / 14.2\% FPR on dev / desktop), while dev- and
desktop-trained priors transfer \emph{to} web well (0.9\% / 0.6\%). Server
workloads are the most idiosyncratic: easy to recognize from outside, hard to
generalize from. A single-workload prior does not cross domains---which is
precisely why the architecture mandates per-machine (or per-role) onboarding
rather than a universal static model.

\subsection{Model Characteristics}

\begin{table}[htbp]
\centering
\caption{Model and training characteristics.}
\label{tab:chars}
\small
\begin{tabular}{@{}ll@{}}
\toprule
\textbf{Property} & \textbf{Value} \\
\midrule
Parameters & 0.88M \\
Architecture & 4-layer causal Transformer \\
Hidden dimension & 128 \\
Attention heads & 4 \\
Context window & 128 events \\
Vocabulary & 263 tokens \\
Training data & 116K events, 7 VMs \\
Validation perplexity & 1.6 \\
Training time & $\sim$3 minutes (GPU) \\
Inference & CPU-capable, real-time \\
Codebook size & $\sim$80 KB \\
\bottomrule
\end{tabular}
\end{table}

\section{The Economics of Coevolution}
\label{sec:economics}

\begin{figure}[htbp]
\centering
\includegraphics[width=0.62\textwidth]{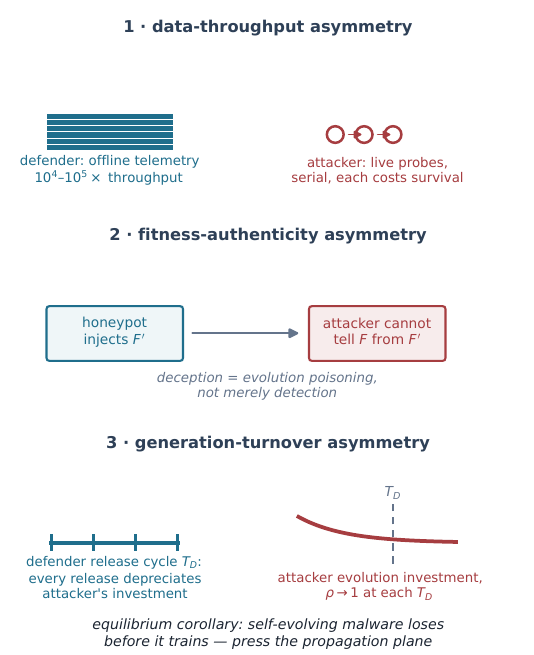}
\caption{Three structural asymmetries of red--blue coevolution. Together
they imply the equilibrium corollary: self-evolving malware loses before it
trains; defense belongs on the propagation plane.}
\label{fig:econ}
\end{figure}

The law chain (\S\ref{sec:lawchain}) is what the loop measured. This
section states what it \emph{means} at equilibrium, as three structural
asymmetries (Figure~\ref{fig:econ})---each grounded in a measurement from
this report rather than in assumption---and the defense allocation they
imply.

\subsection{Three Structural Asymmetries}

\textbf{1. Data-throughput asymmetry.} The defender trains on offline
telemetry at scale (884K benign host events in our corpus alone); the
attacker's fitness evaluations are \emph{serial live probes}, each paid in
exposure and survival time. Our range makes the ratio concrete: a single
10-minute G1 round yields one fitness bit per variant, while the defender's
offline corpus supports arbitrarily many training epochs. The throughput
gap is $10^{4}$--$10^{5}\times$ and does not shrink with better
algorithms: both sides' learning improves with compute, and the ratio is
structural, not technological.

\textbf{2. Fitness-authenticity asymmetry.} The defender controls part of
the environment the attacker evaluates against. A honeypot can inject an
arbitrary fitness signal $F'$, and the attacker cannot distinguish $F'$
from $F$ at any cost below owning a no-probe environment model---which
asymmetry~1 already forbids. Deception is therefore not merely a detection
tool but \emph{evolution poisoning}: it shapes what the attacker becomes.
Our hunter discipline (\S\ref{sec:lab}) is the honest, non-deceptive
baseline; the asymmetry says the deceptive variant is strictly stronger.

\textbf{3. Generation-turnover asymmetry.} The attacker's evolutionary
investment depreciates at the defender's release cadence: each new detector
model resets the fitness landscape the attacker paid probes to map. The
defender's release period $T_D$ is thus a strategic weapon---frequent small
updates tax the attacker's learning budget even when each update's
detection gain is marginal.

\subsection{The Equilibrium Corollary and Its Evidence}

The corollary---\emph{self-evolving malware loses before it
trains}---predicts the field observation that top-tier intrusions remain
human-operated: the human operator is a zero-shot generalizer whose fitness
evaluation runs offline in a head, escaping asymmetry~1. Our own
measurements point the same way. Exposure conservation: the AAA's every
survival mechanism paid an exposure currency on some face (mimicry cut the
DT-slot KL from 1.05 to 0.54 while PROC stayed at 7.3; G1 bought shape-axis
evasion and paid body-config exposure, then bought body-config evasion and
paid rhythm exposure). Virulence--transmission trade-off~\cite{anderson1991}: H7's fit required
\emph{less} activity, not more (122 steps vs.\ H0's 147), and the G7
allele that maximized aggression completed zero tasks. And L1's null law:
where the defense cannot kill, nothing evolves---which is why open-loop
benchmarks systematically overstate the adaptive threat.

\subsection{What the Defense Should Buy}

The allocation follows: press the \emph{propagation plane} (egress control,
lateral-movement segmentation, scan-rate detection), where each compression
doubly taxes the attacker's probe budget (slower evaluation \emph{and}
shorter survival); invest deception as evolution-poisoning infrastructure,
not as decoration; and treat release cadence $T_D$ as a weapon. The
detection arms race---more model against more adversary---is the one
investment whose cost asymmetry favors the attacker, and the one the field
keeps buying.

\section{Discussion}
\label{sec:discussion}

\subsection{Stepping Cadence: A Weakened Trade-off and Where It Re-Binds}
\label{sec:cadence}

The 30$\times$ CV separation between the AAA's sleep cadence and benign sleep
patterns was, in our earlier analysis, read as the mathematical expression of
a fundamental stealth--functionality trade-off. The cadence-evasion boundary
experiment (Section~\ref{sec:cvboundary}) forces a more careful statement.

\textbf{The original argument and its hidden assumption.} Let an agent's
inter-action intervals be drawn from a distribution $P(\Delta t)$ with
coefficient of variation $\mathrm{CV} = \sigma / \mu$. Evading temporal
detection requires $\mathrm{CV} > c^{*} \approx 1.5$, hence a heavy-tailed
interval distribution: most intervals short, a non-negligible fraction very
long. We argued that the long silences directly degrade task responsiveness,
making the trade-off monotonic: more stealth, less function. The hidden
assumption is that the agent's task requires \emph{responsiveness
throughout}. A reconnaissance-style task with a fixed action budget and a
relaxed deadline does not: with the mean interval held at 7.5\,s, a
log-normal shape of $\sigma \geq 1.25$ crosses the detection line at
\emph{unchanged} expected throughput (Section~\ref{sec:cvboundary}). The
silence cost is real but does not bind this task class. The trade-off, in its
strong single-face form, is falsified.

\textbf{Where the trade-off re-binds.} The evader's problem reappears one
level up. To pass the temporal face it must heavy-tail its timing; to pass
the grammar face it must match the benign field-combination distribution; to
pass the aggregate test it must keep its \emph{rate} of grammar-anomalous
events indistinguishable from benign. In our adversary family, the third
constraint is what binds: the reconnaissance action set is distributionally
distinct from the routine workload (3.3$\times$ anomaly-rate elevation,
windowed $p \approx 10^{-9}$). Each single face has an evasion window; the
intersection of faces is what the adversary cannot simultaneously
satisfy---the structural vulnerability migrates from any single statistic to
the joint constraint across faces. This is also the honest limit of the
claim: an adversary that additionally matches the benign command distribution
would compress the grammar-side margin, and Section~\ref{sec:limitations}
scopes the conclusion accordingly.

\textbf{Why the AAA settles at CV $\approx$ 0.3--0.7.} The AAA's training
optimizes a reward combining survival, task completion, and detection
penalty. Its equilibrium CV sits far below the benign lower bound---the
reward function trades a large temporal anomaly for a small functional gain,
a rational but exploitable choice. Notably, the E1 sweep shows the AAA's own
stepping family \emph{could} cross the CV line without reward shaping that
penalizes grammar-side exposure; that it does not exploit this window
reflects its reward structure, not the window's absence---motivating the
adversary-variant studies of future work.

\subsection{The Economics of Attack-Defense Coevolution}

We frame the attack-defense dynamic within a coevolutionary economics model
drawing on the adversarial balance framework. Table~\ref{tab:economics}
summarizes the structural asymmetries.

\begin{table}[htbp]
\centering
\caption{Offense-defense economic asymmetries.}
\label{tab:economics}
\small
\begin{tabular}{@{}p{3.2cm}p{8.4cm}c@{}}
\toprule
\textbf{Principle} & \textbf{Content} & \textbf{Favors} \\
\midrule
Rule vs.\ Baseline & Circumventing a rule is free (rename a process);
circumventing a baseline is expensive (must truly resemble normal) & Defense \\
Exposure Conservation & Every survival mechanism pays an exposure cost;
dormancy is the only all-layer exposure reduction & Neutral \\
Data Throughput Asymmetry & Defense trains on massive historical telemetry
offline; attack evaluates fitness through live probing & Defense \\
Fingerprint Increment & Each added evasion technique opens a new observation
surface; marginal stealth gain diminishes & Defense \\
Defense Amortization & One rule protects millions of endpoints; attack
adaptation must be re-earned per environment & Defense \\
Stealth Half-Life & Discovery is a one-way function; the cleanup moment is not
chosen by the attacker & Defense \\
\bottomrule
\end{tabular}
\end{table}

The \textbf{``rule vs.\ baseline'' principle} is the theoretical foundation of
behavior-grammar detection. Traditional detection encodes rules---specific
patterns that trigger alerts. Circumventing a rule costs the attacker nearly
nothing: rename a process, change a port, reorder operations. Behavior-grammar
detection replaces rules with a \emph{baseline distribution}. Circumventing
the baseline requires the attacker to produce events that are statistically
indistinguishable from benign behavior across all eight dimensions
simultaneously---including the temporal dimension, which is structurally
constrained by the stealth-functionality trade-off.

The cost asymmetry is dramatic: rule evasion is an $O(1)$ operation (one
modification), while baseline evasion is an $O(d)$ operation where $d$ is the
number of independently observed dimensions, each requiring distribution
matching. The five-network fusion architecture maximizes $d$, and the temporal
analyzer adds a dimension that is \emph{structurally impossible} to match
without sacrificing functionality.

\textbf{Exposure conservation} provides a complementary lens. Define the
``exposure currency'' $E$ as the sum of observable signals across all
detection surfaces. Any evasion mechanism that reduces $E$ on one surface
(e.g., disguising process names) increases $E$ on another (e.g., the
process-name $\times$ parent-process combination becomes anomalous). The only
action that reduces $E$ on \emph{all} surfaces simultaneously is dormancy
(complete inactivity)---but dormancy sacrifices all functionality. The agent
cannot drive $E$ below the detection threshold while maintaining task output.
Our detection architecture exploits this by maintaining multiple independent
observation surfaces, making the conservation constraint binding.

\textbf{Data throughput asymmetry} quantifies the learning speed gap. The
defender processes millions of historical events offline, at computational
cost measured in GPU-hours. The attacker's each fitness evaluation requires a
live probing action that consumes a survival interval (the agent must act,
observe the response, and survive long enough to propagate the learning). The
throughput ratio is $10^4$--$10^5$ in the defender's favor---a gap that does
not narrow with algorithmic progress, because it reflects the fundamental
difference between offline batch processing and online sequential evaluation.

\subsection{Evolvability vs.\ Behavior: The Deeper Theoretical Frame}

The SPECIES framework (Digital Organism Architecture Constitution) introduces
a distinction that illuminates the detection problem at a deeper level:
\textbf{the difference between delivering behaviors and delivering the
mechanism that generates behaviors (evolvability vs.\ behavior)}.

Traditional detection defends against \emph{behaviors}---signatures of known
attack patterns. The adaptive threat model delivers not a fixed behavior but
an \emph{evolvability} mechanism: a system that generates new behaviors under
selection pressure. As the SPECIES framework articulates: ``We do not deliver
behaviors; we deliver the mechanism that generates behaviors. Its rules are
written not by design but by death---by selection pressure.''

This reframes the detection challenge. Against a fixed-behavior adversary,
signature detection suffices: enumerate the behaviors, match them. Against an
evolvable adversary, signature detection is a losing strategy: the adversary
generates novel behaviors faster than signatures can be authored.
Behavior-grammar detection occupies a middle ground: it does not enumerate
behaviors but learns the \emph{distribution} of benign behaviors and flags
deviations. The question is whether an evolvable adversary can, through
selection pressure, evolve behaviors that fall within the benign distribution.

\textbf{The three learning loops.} The SPECIES framework proposes three nested
learning loops operating at different timescales:

\begin{enumerate}[leftmargin=1.8em]
\item \textbf{Perception loop (second-scale, on-device):} Zero-gradient
episodic memory retrieval. The agent recognizes patterns from immediate
experience without weight updates.
\item \textbf{Consolidation loop (hour-scale, dormant windows):}
Perturbation-based evolution of a micro-adaptation layer (hundreds of
parameters). The agent adjusts its behavioral policy during idle periods.
\item \textbf{Evolution loop (generation-scale, off-device):} Full gradient
training plus distillation and mutation. The species evolves core weights
based on population fitness.
\end{enumerate}

\textbf{Credit assignment is dissolved through stratification:}
within-lifetime learning uses retrieval (no attribution needed), the
adaptation layer uses perturbation (dimensions too few to matter), deep
knowledge uses offline gradients (a solved problem), and population-level
knowledge uses selection (gradient-free). Each loop handles the credit
assignment problem at its own timescale, avoiding the sparse-reward deadlock
that plagues monolithic RL.

Our detection architecture is designed to be robust against each loop:
\begin{itemize}[leftmargin=1.6em]
\item Against loop 1 (perception): the prior model is static within a
deployment, so on-device retrieval cannot adapt to it.
\item Against loop 2 (consolidation): the temporal cadence constraint is
structural---micro-adaptation cannot produce heavy-tailed interval
distributions without degrading functionality.
\item Against loop 3 (evolution): the multi-VM joint prior and self-learning
pattern extraction pipeline allow the defender to update the baseline
distribution at a faster cadence than the attacker's generational turnover.
\end{itemize}

\textbf{Prediction error as universal currency.} The SPECIES framework
identifies prediction error as the single signal shared across learning,
exploration, safety, communication, and evolution. Our detection engine
operates on the same principle: the TinyGPT prior \emph{is} a prediction
engine, and anomaly \emph{is} prediction error. The negative log-likelihood
score is the ``surprise'' signal---the same currency the adversary uses for
learning. This creates a symmetrical battleground: both sides optimize
prediction error, but the defender's prediction is about behavior
distributions while the attacker's prediction is about defensive response. The
defender has the advantage of more data (throughput asymmetry) and a static
target (the benign distribution changes slowly), while the attacker must
predict a moving target (the detection system updates).

\textbf{Control written into the body, not into rules.} The SPECIES
framework's governance principle---that control must reside in non-evolvable
substrate, not in behavioral constraints that selection can erode---has a
direct analog in detection architecture design. Our per-dimension thresholds
and temporal CV bounds are \emph{structural} properties of the detection
system, not learned parameters. The adversary cannot ``evolve past'' the CV
$<$ 1.5 threshold because it reflects a property of goal-directed behavior
itself, not a learned pattern that could be matched. This is the detection
equivalent of writing control into the body: the detection constraint is a
physical law of the behavioral space, not a rule the adversary can learn to
circumvent.

\subsection{The Honest Footnote: Dual-Use Tension}

The SPECIES framework's ``honest footnote'' states: ``We are building an
entity whose core capability is evading oversight, and then requiring it to
comply with oversight.'' The detection counterpart of this tension is equally
honest: we are building a detection system whose core assumption is that
adaptive behavior has structural limits, and we validate this assumption
against an adversary specifically designed to test those limits.

The tension is resolved not by perfect detection (which is unachievable) but
by the coevolutionary equilibrium: the detection system does not need to be
perfect, only sufficiently expensive to evade that the attacker's equilibrium
strategy favors human-operated attacks over autonomous adaptation. The
empirical evidence supports this: top-tier APT groups employ human operators
rather than heavy autonomous algorithms~\cite{mandiant2023}, which is
consistent with our model's prediction that the equilibrium attacker response
is zero-shot human generalization rather than algorithmic adaptation.

\subsection{Coevolutionary Dynamic Equilibrium}

The adversarial engagement documented in Section~\ref{sec:rounds} reveals a
coevolutionary dynamic. Each detection upgrade narrows the attacker's viable
strategy space; each attacker adaptation opens new observation surfaces. The
two sides serve as each other's whetstones---confrontation produces overall
balance rather than one-sided dominance.

The equilibrium has a specific structure in our system:

\begin{enumerate}[leftmargin=1.8em]
\item \textbf{Detection floor:} The behavioral grammar prior establishes a
baseline detection rate ($\sim$93\%) that holds regardless of attacker
adaptation, because the prior captures the benign distribution, not specific
attack patterns.
\item \textbf{Temporal ceiling (single-face, bounded):} The CV $< 1.5$
constraint raises the cost of temporal stealth, but
Section~\ref{sec:cvboundary} shows it is not a hard ceiling---heavy-tailed
intervals cross it at fixed throughput; the ceiling holds only jointly with
the grammar aggregate.
\item \textbf{Diminishing returns:} Each additional evasion technique the
attacker deploys opens a new observation surface (fingerprint increment), and
the marginal stealth gain diminishes while the marginal detection surface
grows.
\item \textbf{Defender's labeling production line:} The self-learning pattern
extraction pipeline converts each observed attacker behavior into a new
detection pattern, creating an asymmetric resource regeneration rate that
favors the defender.
\end{enumerate}

\subsection{Limitations}
\label{sec:limitations}

We candidly acknowledge the following limitations:

\begin{enumerate}[leftmargin=1.8em]
\item \textbf{Distribution drift.} The benign behavioral distribution is not
stationary---software updates, configuration changes, and workload shifts
cause gradual drift. Long-term deployment requires periodic baseline
retraining. The vocabulary growth curve (42 $\to$ 148 $\to$ 167 $\to$ 306
tokens across training iterations) serves as a health indicator: vocabulary
convergence signals that the behavioral space has been adequately sampled,
triggering retraining when new tokens appear.

\item \textbf{Kernel-level blindness.} We assume the eBPF telemetry collection
is not subverted. A kernel rootkit that blinds the collection layer is out of
scope. Defending against this requires hardware-rooted attestation, which is
orthogonal to our contribution.

\item \textbf{Shell-built-in commands.} Events from shell built-in commands
(e.g., \texttt{exec}, \texttt{source}) that do not trigger
\texttt{sched\_process\_exec} are invisible to the current telemetry set.
Extending to \texttt{security\_openat} / file-write events would close this
gap.

\item \textbf{Low-rate attacks.} Extremely slow attacks (one action per hour)
may be diluted below the EWMA detection threshold. The temporal CV analyzer
partially addresses this (cadence is measured regardless of rate), but very
sparse activity produces insufficient samples for reliable CV estimation.

\item \textbf{Perfect mimics.} An attacker that perfectly matches the benign
distribution across all eight dimensions---including temporal cadence---would
be undetectable by the grammar prior. The public-benchmark evidence bounds
this risk empirically: on OpTC scenario~3, the red team's most user-like
low-and-slow operation, detection falls to 22--42\%
(Section~\ref{sec:optc})---the mimicry ceiling is real, though the
stealth-functionality trade-off still constrains goal-directed behavior.

\item \textbf{Single-face reliance.} The temporal face's cadence-CV rule is
evadable by heavy-tailed inter-action intervals ($\sigma \geq 1.25$ crosses
the detection line at fixed task throughput;
Section~\ref{sec:cvboundary}). In our adversary family the evasion leaves a
statistically significant grammar-side fingerprint ($p \approx 10^{-9}$ under
windowed aggregation), but an adversary that simultaneously matches the
benign command distribution \emph{and} heavy-tails its timing is not ruled
out by the present evidence.

\item \textbf{Coverage.} The current prototype library covers 14 ATT\&CK
techniques out of hundreds in the Linux matrix. Full coverage requires
systematic collection (the Atomic Red Team pipeline~\cite{atomicredteam},
targeting $\sim$600 techniques, is prepared but pending infrastructure).

\item \textbf{Attribution precision.} While technique-level recall is 14/14,
exact attribution precision is approximately 55\%. Automated remediation is
gated on this metric, currently limiting full automation to high-severity
adaptive patterns.

\item \textbf{Benign baseline homogeneity.} Our main benign baseline is
dominated by a narrow workload (95.1\% of events are ssh-related connections,
83.3\% are network-connect events), which inflates the apparent separability
for all evaluated models---including the baselines' near-zero FPR. Detection
claims under the open-vocab condition should be read with this caveat; the
in-vocab closed-world results (Section~\ref{sec:fourgen}) are the
conservative reference. We directly stress-tested this concern on seven VMs
spanning three deliberately heterogeneous workload roles
(Section~\ref{sec:hetero}): per-role holdout FPR stayed at 0--0.66\% with
3/3 attack detection and 0\% LOVO cross-host FPR, so the core conclusions
are not an artifact of baseline homogeneity---though the roles are still
script-driven simulations, and real user behavior is more diverse.
\end{enumerate}

\subsection{Ethics and Open-Source Strategy}
\label{sec:ethics}

This work is defense-oriented. All experiments were conducted in
libvirt-isolated networks (no outbound routing, iptables full-drop fallback)
on differential clone VMs; the red-side artifact was never executed outside
the isolated range.

We follow an asymmetric disclosure policy with three layers.
\emph{First}, the entire defensive side is open-sourced---architecture,
tokenization, training protocol, prototype pipeline, and deployment
tooling.\footnote{\url{https://github.com/dahan6/siming}} \emph{Second}, the
adversary (Rove Beetle) is published at the abstraction level only:
architecture and behavioral statistics, no implementation. The red-side
artifact is a deployable capability, and withholding implementation details
is the sufficient and standard condition for non-release in responsible
red-team publication; operational parameters published in this paper (e.g.,
the stepping log-normal used during the engagement) have been rotated out of
use. We are additionally preparing a \emph{de-weaponized reference
implementation} of the adversary generator
(Section~\ref{sec:toxoplasma}): the genome schema, morphology
generator, selection loop, and measurement pipeline are released in
full, while the weaponization layer (payload encryption, disguise-name
pool, live C2 paths) is structurally absent rather than switched
off---the released code cannot be rearmed by configuration.
\emph{Third}, the evaluation protocol isolates measurement from
capability: where the adversary's strength is asserted, it is anchored to
external arbiters (concurrent commercial detection stacks,
Section~\ref{sec:red-evidence}) rather than to reproducible attack tooling.

The dual-use calculus is as follows. Publishing the detector strengthens the
defensive baseline, which is expensive to circumvent---evasion requires
behavior indistinguishable from benign across eight dimensions
\emph{and} across aggregated windows (Section~\ref{sec:cvboundary}), a
requirement that compresses the attacker's action space far more than the
publication aids offensive development. Withholding the adversary denies free
rule-level intelligence, which is cheap to misuse.

\section{Conclusion}
\label{sec:conclusion}

We have presented \emph{Behavioral Grammar}, a detection architecture that
treats host runtime behavior as a structured language and learns its grammar
with a 0.88M-parameter causal Transformer. We evaluated it against the
strongest adaptive adversary we could construct---\emph{Rove Beetle}, whose
strength is externally anchored by zero detections under four concurrently
running commercial-grade detection stacks. Across three rounds of closed-loop
engagement, the system achieves a 93\% detection rate at a 3.84\% onboarding
false-positive rate.

Our findings support four conclusions:

\textbf{First, representation determines the detection ceiling.} The 8-token
field-level discretization, not the model capacity, is the binding constraint.
Adding a single slot (path category) made three previously inseparable
persistence techniques linearly separable, and an off-the-shelf LSTM on the
full grammar matches the Transformer at 100\% detection on exported attack
traces (Section~\ref{sec:baseline}). Model size should be dictated by
grammar complexity, not benchmark-driven parameter escalation---and the
behavioral grammar of a single host is extremely narrow (val\_ppl $= 1.6$).

\textbf{Second, no single observation face is sufficient---and that is now
measured, not asserted.} The temporal cadence face separates adaptive
stepping from benign activity by 30$\times$ (CV $=0.310$ vs.\ $9.786$), yet
heavy-tailed intervals evade it at fixed task throughput; the evasion leaves
a grammar-side fingerprint recoverable by windowed aggregation at $p \approx
10^{-9}$ (Section~\ref{sec:cvboundary}). The structural limit of adaptive
behavior lives in the \emph{intersection} of faces, not in any single one.

\textbf{Third, the conclusions hold beyond the home benchmark.} On the public
OpTC/DARPA TC dataset the unmodified pipeline attains 22--87\% (mean 63\%)
per-target detection with 0\% same-day benign FPR, degrading gracefully as
red-team tradecraft approaches mimicry (Section~\ref{sec:optc}); on seven
heterogeneous-workload VMs the homogeneity explanation is ruled out
(Section~\ref{sec:hetero}).

\textbf{Fourth, the economics of coevolution favor the defender in the
measured regime.} Rule circumvention is free; baseline circumvention is
expensive. The defender trains on massive offline data; the attacker
evaluates through live probing. Each evasion technique opens a new
observation surface. These structural asymmetries reflect the difference
between offline batch learning and online sequential adaptation---bounded, as
the second conclusion shows, by the defender's willingness to fuse multiple
faces rather than trust any single signal.

The research implication is that detection efforts should focus not on larger
models but on better representations, more independent observation surfaces,
and faster baseline update cycles. The behavioral grammar framework---compact,
auditable, deployable on edge devices---demonstrates that this focus is both
practically viable and theoretically grounded. Evaluation on the OpTC public
benchmark (Section~\ref{sec:optc}) is included; future work extends this to
DARPA TC Engagement 5, whose richer multi-host provenance would further
stress-test the cross-machine generalization claims
(Section~\ref{sec:crossmachine}).

\section*{Declaration of competing interest}

The author declares no competing interests.

\section*{Code and Data Availability}

The detection engine (\emph{Siming}) is open-sourced at
\url{https://github.com/dahan6/siming}: the Behavioral Grammar detection
engine (six-network fusion), models, patterns, and evaluation data. The
coevolution laboratory is released in de-weaponized form at
\url{https://github.com/dahan6/whetstones-lab}: genomes, evolution rounds,
telemetry, the training harness (stubbed primitives), and range tooling.
Offensive payloads, real primitives, and C2 infrastructure are intentionally
withheld. Research and defensive use only. The statistical characteristics of
the attack trajectories (calibration data) are publicly documented in this paper;
the raw attack trajectories are not publicly released due to dual-use
considerations.

\section*{CRediT authorship contribution statement}

\textbf{Zihan Luo:} Conceptualization, Methodology, Software, Investigation,
Writing.

\section*{Preprint statement}

A preliminary version of this work is available as a preprint on arXiv
(arXiv:2608.00745).

\appendix
\section{Registered Experiment Protocols}
\label{app:protocols}

This appendix registers the experiment designs referenced as E-A through E-G
in the main text. Each entry fixes the hypothesis, the grid, the measured
quantities, and the acceptance criterion \emph{before} the data is taken;
results are backfilled into the corresponding main-text sections as rounds
complete. All rounds run on the isolated range (Section~\ref{sec:ethics})
and are logged in the EXPERIMENTS ledger format.

\subsection*{E-A: Online-ization of the count-aggregation face}
\textbf{Hypothesis.} The windowed binomial count test and the group
total-sequence statistic can run online inside the deployed pipeline with
readings identical to their offline analysis counterparts.
\textbf{Grid.} Offline vs.\ online scoring over three archived traces (E1
range set; one benign-heavy set; one R2 set).
\textbf{Measures.} Per-window $p$-values, alert paths, divergence between
offline and online readings.
\textbf{Acceptance.} Online alert path live; offline/online readings agree
within numerical tolerance.
\textbf{Status.} In implementation (the deployed pipeline currently
implements per-event and per-chain faces; the count face exists only as
offline analysis).

\subsection*{E-B: Silence $\times$ window matrix (the frontier measurement)}
\textbf{Hypothesis (P1--P3).} Fixed-window detection time scales as $1/r$ at
fixed $\Delta q$; silent morphs trade task time for evasion time at rate 1:1;
sequential accumulation recovers floor-evading morphs at $D \approx N^{*}/r$.
\textbf{Grid.} Morph ladder S1--S5 (genome-defined):
S1 $=$ M7 slow-ghost ($\mu{=}2.2, \sigma{=}0.9$);
S2 $=$ slower ($\mu{=}2.6, \sigma{=}0.9$);
S3 $=$ burst--silence hybrid ($\mu{=}2.2$, burst $0.3/3.5$);
S4 $=$ long-sleep cession ($\mu{=}2.2$, \texttt{max\_sleep\_s} $\in \{60, 120\}$);
S5 $=$ jittered ($\mu{=}2.2$, \texttt{sigma\_jitter} $= 0.3$);
crossed with defender configurations \{fixed window
$W \in \{90, 300, 900, 3600\}$\,s; sequential accumulation\}. One round
(8--10\,min) per cell, 3 repetitions, medians.
\textbf{Measures.} Detection $p$-value and time-to-detect; task completion
($M$ actions, $T_{\mathrm{task}}$); realized rate $r$; per-event exposure
$q_a$.
\textbf{Acceptance / output.} The $(r, W)$ detection frontier plotted with
theory curve Eq.~\eqref{eq:frontier} and measured morph placements overlaid;
the re-binding question (does task throughput fall proportionally?) answered
with intervals.

\subsection*{E-C: Group total-sequence counter to interleaving}
\textbf{Hypothesis (P5).} Chain-group superposed statistics (clustered by
shared ancestor, destination, landing directory, genome) detect $k$-chain
time-interleaving that defeats per-chain $F_{\mathrm{cv}}$.
\textbf{Grid.} Interleaved adversary driver ($k \in \{1, 2, 4, 8\}$ chains
rotating, per-chain rate fixed, total throughput fixed) $\times$ \{$F_{\mathrm{cv}}$ alone; $F_{\mathrm{cv}} + F_{\mathrm{grp}}$\}.
\textbf{Measures.} Per-face verdict and $p$; clustering attribution error;
task throughput.
\textbf{Acceptance.} $F_{\mathrm{grp}}$ restores significance for $k \ge 2$
at fixed total throughput.

\subsection*{E-D: Adversary-variant statistical discipline}
\textbf{Hypothesis (P6).} The grammar-side exposure $\Delta q > 0$ is
structural, not an artifact of one reward design.
\textbf{Grid.} Reward-function variants $\times$ training seeds $\times$ a
non-RL hand-crafted adversary, each evaluated on the full face stack.
\textbf{Measures.} Per-variant $q_a$, per-face detection, all reported as
mean $\pm$ 95\% CI.
\textbf{Acceptance.} No single-round anecdotes; every claim carries an
interval.

\subsection*{E-E: Dilution gradient (chaff)}
\textbf{Hypothesis (P4).} Count-face significance erodes as $\sqrt{c}$;
$c \ge 10$ is required for practical effect, and such $c$ triggers the
group-level rate face.
\textbf{Grid.} Chaff factor $c \in \{1, 2, 4, 10\}$ at fixed attack-action
rate; benign-grammar chaff.
\textbf{Measures.} Count-face $z$/$p$ per $c$; total-rate-face response;
chaff grammar-benignity check.
\textbf{Acceptance.} Measured erosion matches Eq.~\eqref{eq:dilution} within
tolerance; vice boundaries plotted.

\subsection*{E-F: Long-horizon engagement}
\textbf{Hypothesis.} Findings hold beyond minute-scale rounds.
\textbf{Grid.} At least one hour-scale $\times$ multi-day continuous round
with the full stack live.
\textbf{Measures.} Drift of $q_b$ (baseline stability), detection latency
distribution, adversary adaptation over time.
\textbf{Acceptance.} Ledger complete; drift reported.

\subsection*{E-G: Public-benchmark scale-up}
\textbf{Hypothesis.} Public-benchmark detection rates of
Section~\ref{sec:optc} hold or improve with full-scene coverage and a
stronger benign baseline.
\textbf{Grid.} OpTC all scenarios (beyond the current three-target subset)
and, if compute permits, DARPA TC Engagement 5 full scale; heterogeneous
workload baseline of Section~\ref{sec:hetero}.
\textbf{Measures.} Per-target detection, same-day benign FPR, cross-day
drift.
\textbf{Acceptance.} Full-scene table with intervals; the sc3-class
low-and-slow scenario reported explicitly as the ceiling case.

\subsection*{E-H: A non-self-built adversary (CALDERA)}
\textbf{Hypothesis.} The detection faces do not degrade against an
adversary the authors did not build.
\textbf{Grid.} MITRE CALDERA stock TTP chains (recon, persistence,
collection, exfiltration profiles) executed on the range against the full
face stack, 3 repetitions per profile.
\textbf{Measures.} Per-profile detection rate, time-to-detect, benign FPR
during the same windows.
\textbf{Acceptance.} Detection at or above the E4 public-benchmark band;
any profile below it reported with its face-level explanation.

\subsection*{E-I: Alert economics}
\textbf{Hypothesis.} The fused pipeline operates inside a realistic alert
budget without giving back its detection margin.
\textbf{Grid.} Alert-budget ladder (per-day alert caps) $\times$ the E1/E4
trace sets.
\textbf{Measures.} Alerts per day per host; detection rate as a function
of budget (the Axelsson operating curve); time-to-detect under budget.
\textbf{Acceptance.} The budget--detection curve published with intervals;
the knee identified.

\subsection*{E-J: Provenance-system comparison (UNICORN/Kairos class)}
\textbf{Hypothesis.} Provenance-causal detection and grammar-face
detection capture different structure on shared data (complementarity, not
dominance).
\textbf{Grid.} A shared scenario corpus (E4 subset plus one range set)
scored by our stack and by a provenance-system reference under identical
splits.
\textbf{Measures.} Per-system detection and FPR per scenario class;
per-face attribution on our side.
\textbf{Acceptance.} Same-data, same-metric table; whichever system leads
per scenario class is reported as-is.

\subsection*{E-K: DARPA TC Engagement 5, full scale}
\textbf{Hypothesis.} E4's public-benchmark band holds at full-corpus
scale.
\textbf{Grid.} TC E5 full corpus, all scenarios, all days.
\textbf{Measures.} Per-scenario detection, benign FPR, cross-day drift.
\textbf{Acceptance.} Full table with intervals; drift anatomy reported
explicitly.

\end{document}